\documentclass[pdflatex,sn-mathphys-num]{sn-jnl}

\usepackage{graphicx}%
\usepackage{multirow}%
\usepackage{amsmath,amssymb,amsfonts}%
\usepackage{amsthm}%
\usepackage{mathrsfs}%
\usepackage[title]{appendix}%
\usepackage{xcolor}%
\usepackage{textcomp}%
\usepackage{manyfoot}%
\usepackage{booktabs}%
\usepackage{algorithm}%
\usepackage{algorithmicx}%
\usepackage{algpseudocode}%
\usepackage{listings}%
\usepackage{amssymb}
\usepackage{amsmath}
\usepackage{siunitx}
\usepackage{multirow}
\usepackage{adjustbox}
\usepackage{graphicx}
\usepackage{natbib}
\usepackage{longtable}
\usepackage{lscape} 
\usepackage[table,xcdraw]{xcolor}
\usepackage{changepage}
\usepackage{tabularray}
\DeclareSIUnit\pound{lb}
\DeclareSIUnit\kilogram{kg}
\DeclareSIUnit\Watts{W}
\DeclareSIUnit\kilowatthour{kWh}
\DeclareSIUnit\lux{lx}

\theoremstyle{thmstyleone}%
\theoremstyle{thmstyletwo}%

\theoremstyle{thmstylethree}%

\begin{document}

\title[Article Title]{Defining an Evaluation Method for External Human-Machine Interfaces}


\author[1,2]{\fnm{Jose} \sur{Gonzalez-Belmonte}}\email{josegonz@umich.com}

\author*[1]{\fnm{Jaerock} \sur{Kwon}}\email{jrkwon@umich.edu}

\affil*[1]{\orgdiv{Electrical and Computer Engineering}, \orgname{University of Michigan-Dearborn}, \orgaddress{\street{4901 Evergreen Rd}, \city{Dearborn}, \postcode{48128}, \state{Michigan}, \country{United States of America}}}

\affil[2]{\orgdiv{Math and Computer Science}, \orgname{Lawrence Technological University}, \orgaddress{\street{21000 W Ten Mile Rd}, \city{Southfield}, \postcode{48076}, \state{Michigan}, \country{United States of America}}}


\abstract{As the number of fatalities involving Autonomous Vehicles increase, the need for a universal method of communicating between vehicles and other agents on the road has also increased. Over the past decade, numerous proposals of external Human-Machine Interfaces (eHMIs) have been brought forward with the purpose of bridging this communication gap, with none yet to be determined as the ideal one. This work proposes a universal evaluation method conformed of 223 questions to objectively evaluate and compare different proposals and arrive at a conclusion. The questionnaire is divided into 7 categories that evaluate different aspects of any given proposal that uses eHMIs: ease of standardization, cost effectiveness, accessibility, ease of understanding, multifacetedness in communication, positioning, and readability. In order to test the method it was used on four existing proposals, plus a baseline using only kinematic motions, in order to both exemplify the application of the evaluation method and offer a baseline score for future comparison. The result of this testing suggests that the ideal method of machine-human communication is a combination of intentionally-designed vehicle kinematics and distributed well-placed text-based displays, but it also reveals knowledge gaps in the readability of eHMIs and the speed at which different observers may learn their meaning. This paper proposes future work related to these uncertainties, along with future testing with the proposed method.}

\keywords{autonomous, eHMI, guidelines, av, review, pedestrian, road communication}



\maketitle

\section{Introduction}\label{Introduction}

In 2012 the Institute of Electrical and Electronics Engineers (IEEE) published an article for their online news release where ``distinguished members of IEEE [...] selected autonomous vehicles (AVs) as the most promising form of intelligent transportation, anticipating that they will account for up to 75 percent of cars on the road by the year 2040" \cite{tardo_ieee_2012}. This article has been referenced multiple times by sites such as  Forbes \cite{motavalli_self-driving_2012}, CNN \cite{newcombwired_you_2012}, and WebProNews \cite{wolford_self-driving_2012}. Slightly more recently, in 2015, a study by Viereckl et al. \cite{viereckl_racing_2015} postulated that AVs will achieve full autonomy by 2030 and represent 30\% of the market.

Perhaps more than anything, these predictions show a common expectation for the ubiquity of AVs in the near future. As we approach these predictions, concerns over the safety of AVs are increasing. Eighteen fatal crashes have been linked to AVs between 2018 and 2022 \cite{salter_how_2023,mcfarland_uber_2018}. While right now this is small compared to the 1.19 million deaths attributed to road accidents in 2021 alone \cite{world_health_organization_global_2023}, it provides us with early evidence of the need for higher safety standards in the field. 

Pedestrians are disproportionately affected by road fatalities. The victim of the first AV fatality in 2018 was a pedestrian \cite{salter_how_2023}, and according to the World Health Organization's (WHO) road safety report from 2023, pedestrians represented 23\% of all vehicle fatalities in 2021, followed only by four-wheel vehicles at 30\% \cite{world_health_organization_global_2023}. If cities are to move towards a driverless future, pedestrians must feel and be safe in that environment. As it’s pointed out by Carmona et al \cite{carmona_ehmi_2021}, while previous studies have revealed pedestrian trust on AVs to be above the median \cite{jayaraman_pedestrian_2019,rodriguez_palmeiro_interaction_2018}, most of these opinions are not representative of real-life experiences. Since AVs are not yet commonplace, participants filling these surveys have to do so based on their experience with virtual reality experiments, videos, images, or imagined scenarios.

To improve road safety, some have proposed the use of external Human-Machine Interface displays (eHMIs) to display important information to drivers and pedestrians based on the vehicle’s decisions \cite{carmona_ehmi_2021}. The term was first coined in 2016 \cite{peng_roles_2016} and since then, many proposals for eHMIs have been made, but none have been yet adapted as the standard or accepted to be the definite form of machine-human communication for AVs. While plenty of studies have been conducted on the effectiveness of different types of eHMIs using simulated scenarios \cite{de_clercq_external_2019, bazilinskyy_survey_2019, carmona_ehmi_2021, eisma_external_2021, loew_go_2022, guo_video-based_2022}, there has not been any work to objectively evaluate them to determine the ideal proposal.

This paper looks at previous research on the effectiveness of eHMIs at reducing ambiguity, their effect on the trust pedestrians and drivers have on AVs, and the difficulties of introducing them and standardizing them in our multi-faceted roads. It then proposes a collection of questionnaires that can be universally applied to evaluate any proposal for AV communication that uses eHMIs, and tests it against common proposals. The end result is a general evaluation method that can be used by manufacturers to design eHMIs better suited for our streets and their users.

\section{Related Work}\label{Related Work}

This paper is heavily based on the work of Carmona et al. \cite{carmona_ehmi_2021} where they analyze current eHMI technologies in order to determine guidelines for the future. The authors determined that eHMIs must be readable, easy to understand, multi-faceted,, positioned such that it is visible from multiple angles, colored appropriately, and capable of adapting its method of display to account for the current environment of the vehicle.

This paper also builds upon the study by de Clercq et al. \cite{de_clercq_external_2019}. In that study, participants were put in a VR simulation with multiple types of eHMI and asked to hold a button whenever they felt safe to cross. Participants were also surveyed on their preference of eHMI display (or lack thereof). The conclusions of the study found that participants almost unanimously preferred having an eHMI over no eHMI, and that text-based designs required no training when compared to other types of eHMIs. It also concluded that standardization and regulation would be important in the future.

A similar work by Guo et al. \cite{guo_video-based_2022} had participants connected to an eye-tracking device and looking at 29 video clips of the same vehicle with different eHMIs in three possible locations, mixed with clips of the same vehicle with no eHMI. Participants were asked to indicate when they felt safe to cross. The study concluded that while vehicle kinematics influenced decision making in approximately 12\% of participants, 90\% of them still rated eHMI presence as “very important” or “important” in factors influencing their road-crossing decisions.

Another important previous work is that of de Winter and Dodou \cite{de_winter_external_2022}, where they evaluated and discussed the arguments in favor and against the use of eHMIs. No final conclusion was drawn.

Finally, Bazilinskyy et al. \cite{bazilinskyy_survey_2019} used a crowdsourcing service to survey two thousand participants on the clarity of twenty-eight eHMI proposals. The results pointed towards textual egocentric eHMIs being regarded as the clearest type, but raised concerns about liability, legibility, and technical feasibility.

\section{Methods}\label{Methodology}

To determine a method of evaluating eHMI proposals, this paper:
\begin{enumerate}
    \item Outlines general categories for what characteristics are generally considered important in an eHMI.
    \item Produces, for each category, a list of requisites to fulfill it.
    \item Provides questions that evaluate the presence of these requisites.
    \item Details how to utilize the questionnaire when evaluating an eHMI proposal, and how to record one's answers.
    \item Presents formulas to calculate the individual score corresponding to each category, based on the results of the questionnaire, as well as a formula to calculate the final qualitative score for the proposal.
    \item Tests the produced evaluation method using the five eHMI proposals seen in \cite{de_clercq_external_2019}, comparing the proposals for a text based display, a smiling display, frontal braking lights, a sweeping light display (``Knight Rider"), and the exclusive usage of kinematics as a baseline method of external AV communication.
\end{enumerate}

\subsection{eHMI Guidelines and Question Design}
Based on the guidelines proposed by Carmona et al. \cite{carmona_ehmi_2021}, as well as the observations on the qualities of non-external Human-Machine Interfaces (HMI) posed by Mahmud et al. \cite{mahmud_interface_2020}, seven large characteristics of an ideal method of external communication can be established:
\begin{enumerate}
    \item Standardization
    \item Cost Effectiveness 
    \item Accessibility
    \item Ease of Understanding 
    \item Constant Communication
    \item Positioning
    \item Readability
\end{enumerate}
    
In Appendix \ref{secA1} is a questionnaire divided into seven categories, each evaluating the fulfillment of one of the aforementioned seven characteristics. 

\subsubsection{Standardization}
An eHMI must be easily implementable in as many types of autonomous vehicles as possible if it is to become the standard. It’s important to maintain full consistency among all implementations to prevent confusion \cite{de_clercq_external_2019}.

Due to the varied nature of eHMIs, it’s difficult to score the amount of elements being standardized for a single proposal in a vacuum. The questions designed for this guideline hence focus on high-level elements that stand out as requiring consistency between manufacturers, such as symbols, text, timing, color, brightness, framing, and sound. Since a proposal may employ multiple eHMIs,  these high level elements are put into independent scores corresponding to each eHMI, then added together into a singular value that serves as the score penalty. These penalties are applied to a baseline number determined later in this paper.

\subsubsection{Cost Effectiveness}
Production must be cost-effective for vehicle manufacturers in order to ease the adoption of the technology \cite{carmona_ehmi_2021}.

The questions regarding cost effectiveness can be divided into five high-level relevant amounts: manufacturing costs, installation costs on a new vehicle, installation cost in an existing vehicle, maintenance costs, and operation costs. Each cost has its own question where the user of the questionnaire must enter the corresponding amount, in United States dollars and adjusting for inflation to 2022. The addition of these costs is subtracted from a baseline amount, divided by that same amount to normalize it, and multiplied by 10. The baseline amount was chosen based on the average cost of a vehicle in 2022 at \$48 301 according to Kelley Blue Book \cite{kelley_blue_book_new-vehicle_nodate}, 

\subsubsection{Accessibility}
The method of visual communication must be readable for all types of road users, regardless of background. AVs will be sharing space with people of different ages, abilities, cultures, technical knowledge, et cetera. Considerations must be made on a symbol’s possible meaning depending on time, place, and context. 

According to Indiana University \cite{indiana_university_types_nodate}, accessibility can be divided into seven categories. Two of these (Mobility and Medical) are not immediately relevant to the topic of external human-machine interfaces. The five remaining types are:
\begin{enumerate}
    \item Vision
    \item Auditory
    \item Neurological
    \item Cognitive
    \item Psychological
\end{enumerate}

Clear requirements must be defined for each type to ensure a true multifaceted eHMI that is effective for as many people as possible.

\paragraph{Vision}
This section focuses on aiding people with limited or affected vision. 

The eHMI should use pictograms and text to avoid possible ambiguity, as well as to prevent creating barriers of literacy and language. Guidelines for vision accessibility were extracted from the signaling guidelines from Blind Low Vision NZ \cite{blind_low_vision_nz_accessible_2013}, as well as the standards set by the U.S. Access Board \cite{us_access_board_americans_2010}. Additional considerations are made for the variation of vehicle types and circumstances.

Due to colorblindness, the proposal should also not depend on the distinction of colors to communicate information. Since there are many types of colorblindness, there is not a single specific combination of colors to avoid, but people diagnosed with the most common type of colorblindness experience difficulty distinguishing red, green, brown, and orange \cite{colour_blind_awareness_types_nodate}.

Additionally, as stablished by \cite{carmona_ehmi_2021}, eHMIs should not employ colors that have pre-existing associations in the road, such as green or red. Since standards for visibility may also change depending on the country, the color of the eHMI should be able to be adjusted to match local regulations.

Lastly, an eHMI should be clearly distinguishable from the rest of the vehicle, regardless of the color of its paint job.

\paragraph{Auditory}
These guidelines evaluate the quality of any sound cues that the proposal may include, including their clearness and distinguishability. Most visual guidelines targeted at hard-of-hearing people are covered under the visual guidelines, so this section focuses on the acoustic guidelines given by the International Federation of Hard-of-Hearing People \cite{international_federation_of_hard-of-hearing_people_hearing_2021}, as well as elevator and machinery standards \cite{us_access_board_americans_2010}. 

\paragraph{Neurological}
This section targets the reduction of seizures triggered by photosensitivity, following the Web Content Accessibility Guidelines \cite{noauthor_web_nodate,noauthor_understanding_nodate}, as well as the ones postulated by WebAxe \cite{noauthor_vestibular_2015}. These include avoiding flashing lights, the color red, and animations with parallax. Guidelines contradicted by other accessibility guidelines, such as avoiding high contrast (otherwise recommended for visual accessibility), were not included.

\paragraph{Cognitive}
These include reading difficulties, as well as cultural differences that may affect how someone understands signage in an eHMI. Some symbols and colors may have different meanings depending on the cultural context of the observer \cite{krys_be_2015,larsen_not_2018} or even their age \cite{bazilinskyy_survey_2019}. This part of the questionnaire, hence, scores higher the proposals that make use of text or symbols with proven universal meanings, as well as accessible design of text such as concise wording and plain language \cite{office_on_trafficking_in_persons_language_nodate,initiative_wai_cognitive_nodate}. 

Dyslexia is also a particular concern for text-based eHMIs, so guidelines were extracted from Blind Low Vision NZ \cite{blind_low_vision_nz_accessible_2013} and Medium \cite{brokering_creating_2023} in order to address it.

\paragraph{Psychological}
This section concerns itself with the possible mental and emotional impact of the content shown, be it through the use of triggering sounds or upsetting imagery. Questions addressing these concerns were extrapolated from Jankowski \cite{jankowski_improving_2022}.

While \cite{noauthor_web_nodate} recommends giving the ability to turn off timers for cognitive accessibility, this is unrealistic in a road environment, so it was not included.

\subsubsection{Ease of Understanding}
The signaling of an AV should unambiguously communicate the intention of the vehicle to observers encountering it for the first time. This can be measured with self-reported participant surveys as well as observed learning speed in experiments.

Using the definition of elderly and minor according to the U.S. Department of Justice \cite{noauthor_justice_2000} and the Social Security Administration \cite{noauthor_social_nodate} respectively, the score for this section is determined by the percentage of pedestrians that understand the meaning for all of its distinct methods of communication and eHMI(s), after nine encounters with it. The number of encounters was based on the data provided by \cite{de_clercq_external_2019}.

Since some roads have separate lanes for cyclists, they are also taken into account for the questions in this section.

\subsubsection{Constant Communication}
The eHMI(s) in this proposal must have the capability to quickly update observers on the AV’s actions and status at all times \cite{carmona_ehmi_2021}. Some examples of this information can be seen in \cite{bazilinskyy_survey_2019} and include: decelerating, accelerating, engaging in driver mode, detecting a pedestrian, and experiencing technical difficulties. The questionnaire lists multiple possible situations where the vehicle should communicate changes to outside agents, and scores higher those proposals that communicate a wider variety of information. 

\subsubsection{Positioning}
Visual eHMIs must be positioned so they are visible to all observers that may interact with the AV in that specific situation. This should be the case regardless of angle, distance, size of the vehicle, or presence of other vehicles on the road. 

Since the nature of the communication between AVs and other agents on the road varies, a method was needed to score the position of each eHMI in a proposal based on its visibility to the intended target. To discover this, we used the software developed for \cite{gonzalez-belmonte_analytical_2026}. This simulation program uses a series of ray-casted lines to determine what points on a selected vehicle type are visible to a camera, across a series of user-selected permutations, then produces a heatmap illustrating it. This software takes the following set of parameters:
\begin{itemize}
    \item Locations where the camera may be.
    \item Direction that the camera may look to.
    \item Locations where the target vehicle may be. This is the vehicle that records what parts of it are being captured by the camera.
    \item Locations where filler vehicles may be. These are of the same model as the target vehicle, and are solely there to block the vision of the camera.
    \item How close does the target vehicle have to be to the camera to be seen.
\end{itemize}

\begin{figure}[h]
    \centering
    \includegraphics[width=\linewidth,clip,keepaspectratio]{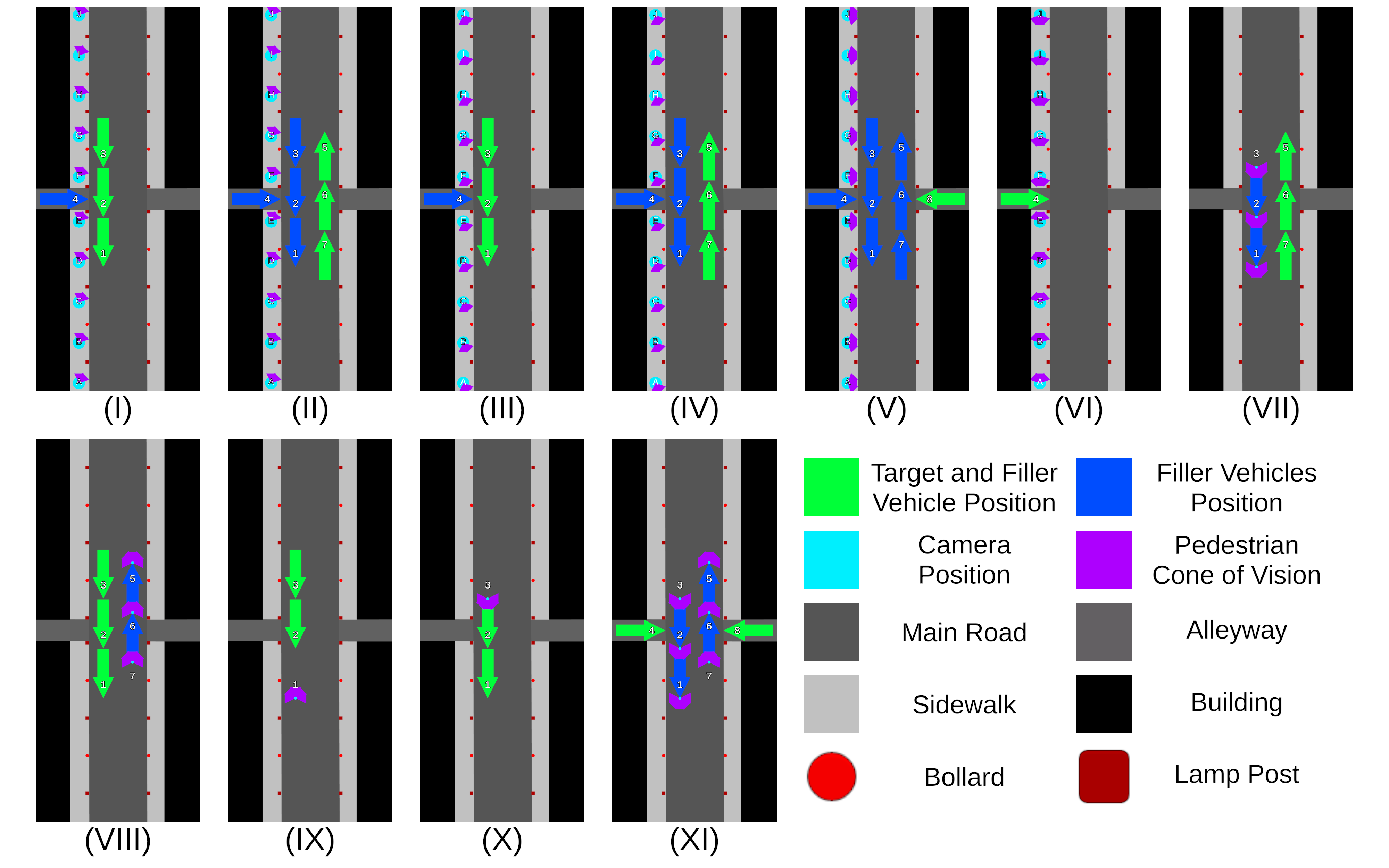}
    \caption{Diagram illustrating the configurations used to determine visibility. }
    \label{fig:positioning_configurations}
\end{figure}

Since we were looking for what parts on the exterior of the vehicle are visible for each observer the situation might be in, we decided there were eight possible overall purposes for an eHMI proposal to have, and simulated scenarios to address them:

\begin{itemize}
    \item To inform pedestrians on the sidewalk of the intent of the AV as an approaching vehicle on the road (Fig. \ref{fig:positioning_configurations}(I, IV)).
    \item To inform pedestrians on the sidewalk of the intent of the AV as it is moving away on the road (Fig. \ref{fig:positioning_configurations}(II, III)).
    \item To inform pedestrians on the sidewalk as the AV comes out of an alleyway or exit that is on the opposite side of the road (Fig. \ref{fig:positioning_configurations}(V)).
    \item To inform pedestrians on the sidewalk of the AV's intent as it exits an alleyway or exit that goes over the sidewalk the pedestrian is on (Fig. \ref{fig:positioning_configurations}(VI)).
    \item To inform drivers driving in the opposite direction of the AV of the intent of the vehicle (Fig. \ref{fig:positioning_configurations}(VII, VIII)).
    \item To inform drivers in front of the AV that are driving in the same direction (Fig. \ref{fig:positioning_configurations}(IX)).
    \item To inform drivers behind the AV that are driving in the same direction (Fig. \ref{fig:positioning_configurations}(X))..
    \item To inform drivers on the road as the AV is about to turn into it from an alleyway or exit (Fig. \ref{fig:positioning_configurations}(XI)).
\end{itemize}

Due to the limited resolution of the software, we also decided to use the SUV 3D model for these simulations, as the largest vehicle in the "medium" classification of the software.  

Similarly to the Standardization section, this section of the questionnaire also requires to analyze each distinct eHMI that makes up the proposal individually, as different signaling may serve different purposes. For each eHMI in the proposal, the user must determine a separate score based on its placement and purpose. 

\begin{figure}[h]
    \centering
    \includegraphics[width={4in},clip,keepaspectratio]{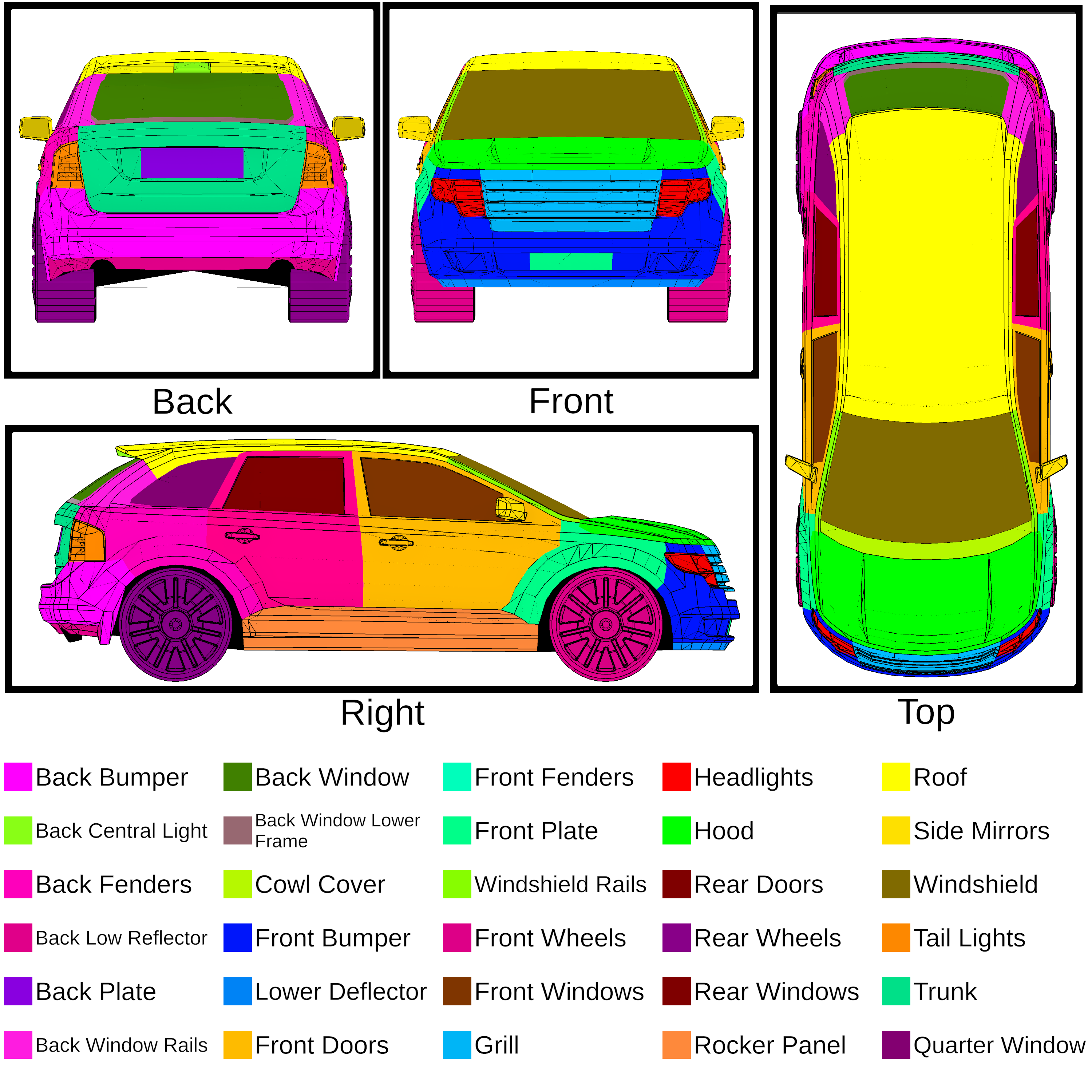}
    \caption{Diagram illustrating the external parts of the SUV from the simulation software, as defined by this section of the questionnaire.}
    \label{fig:vehicle_part_diagram}
\end{figure}

Parts on the exterior of the vehicle that can be largely considered places for eHMI were identified by dividing the SUV model into the following elements, as illustrated in Fig. \ref{fig:vehicle_part_diagram}: back bumper, back central light, back fenders, back low reflector, back plate, back window rails, back window, back window lower frame, back quarter window, cowl cover, front bumper, front low reflector, front doors, front fenders, front plate, windshield side rails, front wheels, front windows, grill, headlights, hood, back doors, rear wheels, rear windows, rocker panels, roof, side mirrors, side mirror's front, windshield, tail lights, and trunk.

\begin{figure}[h]
    \centering
    \includegraphics[width=4in,clip,keepaspectratio]{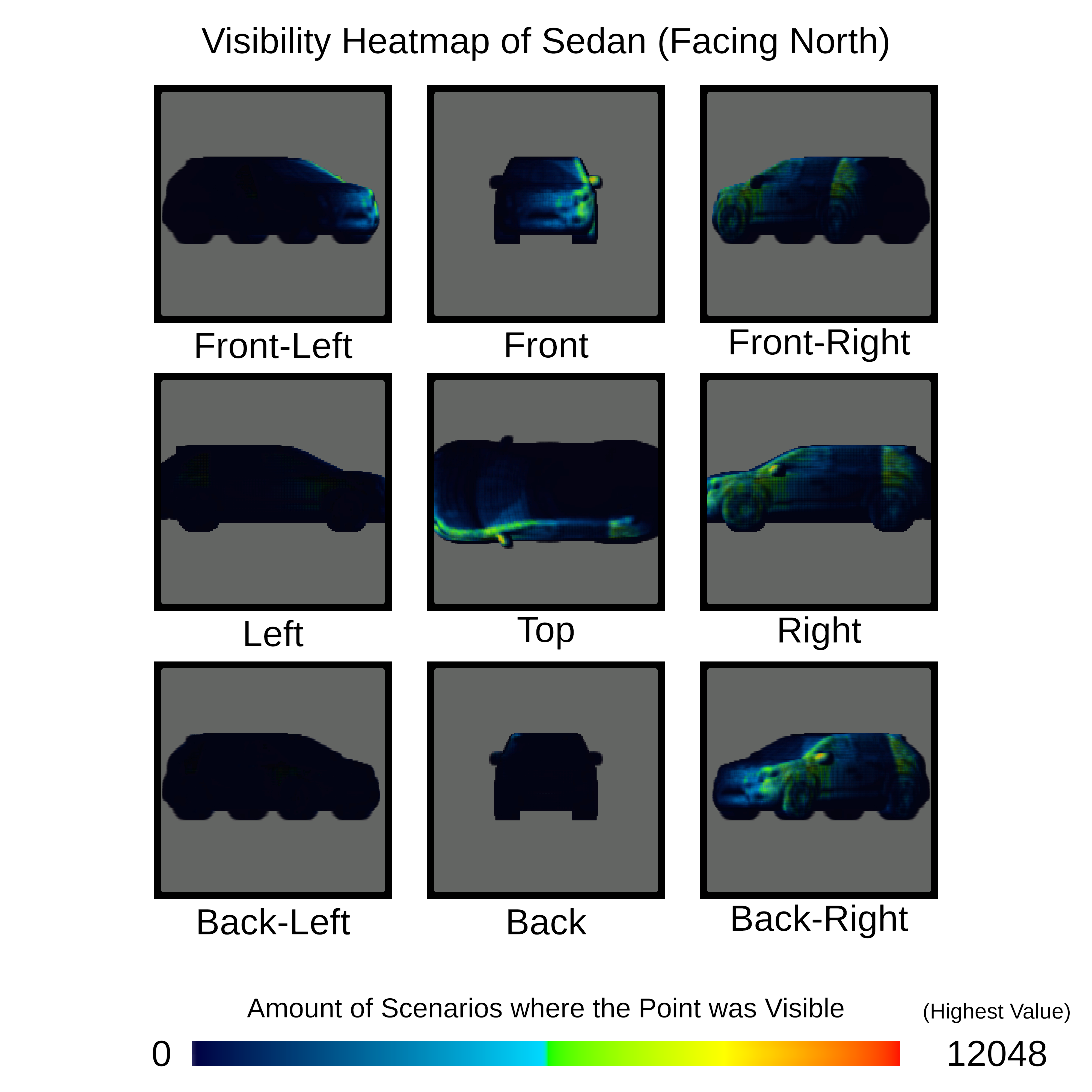}
    \caption{Results showing how many times a Grid Point was recorded, across
all scenarios where the vehicle was an SUV driving North, on the left side of
the virtual environment, and the observer was looking North, towards the front of the vehicle.}
    \label{fig:run_sample}
\end{figure}

To find what elements of the vehicle were better suited for each purpose, the  resulting heat-maps, such as the one in Fig. \ref{fig:run_sample}, were compared against Fig. \ref{fig:vehicle_part_diagram} in order to determine their visibility. Each point has a value indicating many times it was recorded, and its color displays the relationship between that point value and the highest value in the mesh. 
\pagebreak

For each scenario in Fig. \ref{fig:positioning_configurations}, for each exterior element of the vehicle, if at least one point on the element was visible an amount of times equal or higher than 50\% of the highest value recorded for that scenario, it is considered to be visible in that scenario. The results of this process, seen in Table \ref{tab:position_test_results}, were used to determine the formula that the user may employ to calculate how well suited the position of the eHMI is for its purpose.
\pagebreak
\begin{table}[]
\resizebox{\columnwidth}{!}{%
\begin{tabular}{|l|l|}
\hline
\textbf{Vehicle Part} &
  \textbf{\begin{tabular}[c]{@{}l@{}}Configurations where any Grid Point on the vehicle part \\ was recorded in at least 50\% of its permutations\end{tabular}} \\ \hline
Back Bumper.            & II,X                \\ \hline
Back Central Light      & II                  \\ \hline
Back Fenders            & II,IV,VII,VIII      \\ \hline
Back Low Reflector      & X                   \\ \hline
Back Plate              & II,X                \\ \hline
Back Window Rails       & II,IV,VII,VIII      \\ \hline
Back Window             & II,X                \\ \hline
Back Window Lower Frame & II,X                \\ \hline
Back Quarter Window     & IV,VII,VIII         \\ \hline
Cowl Cover              &                     \\ \hline
Front Bumper            & I,IV,V,XI,IX        \\ \hline
Front Low Reflector     & I,V,XI,IX           \\ \hline
Front Doors             & IV,VI,VII,VIII      \\ \hline
Front Fenders           & I,III,IV,V,VI,XI,IX \\ \hline
Front Plate             & V,IX                \\ \hline
Windshield Side Rails   & I,IV,V,VI,VII,VIII  \\ \hline
Front Wheels            & I,IV,II,VI          \\ \hline
Front Windows           & IV                  \\ \hline
Grill                   & I,IV,V,XI,IX        \\ \hline
Headlights              & I,III,IV,V,XI,IX    \\ \hline
Hood                    & I,III,IV,V,XI,IX    \\ \hline
Back Doors              &                     \\ \hline
Rear Wheels             & I,IV                \\ \hline
Rear Windows            & I,IV                \\ \hline
Rocker Panels           & VI                  \\ \hline
Roof                    & II,                 \\ \hline
Side Mirrors            & II,                 \\ \hline
Side Mirror's Front     & I,II,IV,V           \\ \hline
Windshield              & I,V                 \\ \hline
Tail Lights             & II                  \\ \hline
Trunk                   & II,X                \\ \hline
\end{tabular}%
}
\caption{Visibility of different external vehicle elements. Each row indicates the configurations from Fig. \ref{fig:positioning_configurations} in which any Grid Point on the given vehicle part was captured in at least 50\% of its permutations.} 
\label{tab:position_test_results}
\end{table}

Two additional questions were added: one to account for symmetry (since some proposals may not place their eHMIs symmetrically on the vehicle, such as in the place of the driver), and one to account for the eHMI being placed on the roof of the vehicle (e.g. police sirens). 

\subsubsection{Readability}
Other road users must be able to unambiguously and correctly interpret the content displayed by a visual eHMI, regardless of the conditions of the interaction. The questionnaire focuses on four elements: dynamic light adaptation, light conditions, weather conditions, distance, and viewing angle. 

\paragraph{Dynamic Light Adaptation}
Since eHMIs are seen and interacted with at multiple times of the day, the vehicle must be able to adapt their illuminance based on its environment. 

\paragraph{Light Conditions}
This refers to the lighting conditions of the interaction. The questionnaire scores clear visibility under five difference ranges of lighting conditions, based on the illuminance values of ten light sources and conditions, as recorded by Schlyter \cite{schlyter_radiometry_2023} and the American Meteorological Society \cite{american_metereological_society_civil_2024}:
\begin{itemize}
    \item Between a moonless night with overcasted sky (0.0001 \textit{lx}) and moonless night with airglow (0.002 \textit{lx}) 
    \item Between the dark limit of civil twilight (3.5 \textit{lx}) and a public area with dark surroundings (20 \textit{lx})
    \item Between a very dark overcast day (100 lx)  and a normal overcast day (1000 \textit{lx}) 
    \item Between full daylight without direct sun (10000 \textit{lx} to 25000 \textit{lx}) and full Daylight with the sun overhead (130000 \textit{lx})
\end{itemize}

\paragraph{Weather Conditions}
These questions focus on the readability of the eHMI under three different levels of rain (as defined by Skilling \cite{skilling_meteorologists_2018}), three different levels of fog (as defined by the U.S. Department of Commerce \cite{us_department_of_commerce_about_nodate}), and when lightning is present. 

\paragraph{Distance}
These evaluate the minimum and maximum readable distance of the eHMI at ten distance thresholds, based on the work of Priambodo et al. \cite{priambodo_road_2018}, where they determined the readability distance for road signs. The units were converted to feet using the international foot definition (i.e. 1 foot = 0.3048 meter exactly as seen in \cite{national_institute_of_standards_and_technology_us_2019}) and rounded to the nearest whole number. The shortest distance is used as the standard distance to evaluate the other sections.

\paragraph{Viewing Angle}
While positioning is tackled in its own section, this part of the questionnaire evaluates the best possible viewing angles for any given part of all relevant visual eHMIs in the proposal. A point is scored for each 10 degrees where the eHMI can be fully seen and read, starting at 90 degrees (full front view of the eHMI).

\subsection{Score Calculation}
Each section of the questionnaire produces a score between 0 and 10. These results are multiplied by seven adjustable weight values between 0 and 1 (\(S, CE, A, EU, CC, P, R\)), which in turn add up to 7. The final score of an evaluated eHMI proposal is between 0 and 70.

\subsubsection{Standardization Score}
The questions in the standardization section of the questionnaire are repeated for each eHMI present in the overall proposal. The values of all elements are added together and subtracted from a pre-determined baseline value. The result is then multiplied by 10 and clamped between 0 and 10 to prevent negative scores.

Since the only eHMI proposal that would require no standardization is one that makes use of no additional components, the baseline value (\(S_B\)) was determined using 

    \begin{equation} \label{BaselineCalculation:12}
        {S_B} = {S_{P_{no eHMI}}} + S_{count}  
    \end{equation}

where \({S_{P_{no eHMI}}}\) is the total penalty score after evaluating the vehicle kinematics as an eHMI, while \(S_{count}\) corresponds to the total number of questions in Table \ref{tab:QS_table}. 

\subsubsection{Cost Effectiveness Score}
The Cost Effectiveness Score is calculated by adding the values from each question and subtracting them from the baseline score, based on the average cost of a vehicle in 2022 according to \cite{kelley_blue_book_new-vehicle_nodate}. 

According to \cite{pollak_metric_2018}, the ratio between the number of sales of new and used vehicles is approximately 0.75:1. This means that roughly there is 25\% less vehicles on the road that were purchased new, rather than used. Since we're trying to find the cost of implementing this technology in a larger scale, we account for this by multiplying the amount of money required to implement it in a new vehicle by 0.75, before the Cost Effectiveness Score is added together and normalized between 0 and 10, clamped.  

\subsubsection{Ease of Understanding Score}
Calculated by finding the \textit{feel-safe percentage} (\(FSP\)) of the scenario of each question, dividing the cumulative time when the observer reacted as intended by the total time of the interaction, as defined in \ref{fspf:5}, located in the Appendix. The percentages are then added together and normalized between 0 and 10.

\subsubsection{Positioning Score}
Each of the configurations seen in Fig. \ref{fig:positioning_configurations} produced their own distinct heat-map. To calculate a score, each part of the vehicle that an eHMI may be in was given a formula based on the possible purpose of the eHMI and whether that part of the vehicle is visible under the conditions that would facilitate that purpose. It was decided that the formula would make use of the highest value from a set, in order to prevent creating a scoring system that punishes eHMIs for being dedicated to a certain task, as that guideline is evaluated by a different part of the questionnaire.

Since a proposal may have different eHMI elements that communicate distinct information, these are calculated separately before their score is averaged and normalized between 0 and 10 to get a total score for the positioning in the eHMI proposal.

\subsubsection{Accessibility Score, Constant Communication Score, and Readability Score}
Calculated by adding the final cumulative score of the questionnaire, dividing it by the corresponding count of questions for that category, and multiplying the result by 10.
 
\subsection{Tested eHMI Proposals} 
Ahead are descriptions of the five eHMI used in the experiment by de Clercq et al \cite{de_clercq_external_2019}. For the purpose of this evaluation, all vehicles will be treated as Electrical Vehicles with any displays placed on the bumper.

\subsubsection{Vehicle Kinematics as Indication (No eHMI)}
Previous research shows that pedestrians base their decision to cross a street mostly based on a vehicle’s speed, distance, and rolling motion \cite{kadali_proactive_2016}. Moore et al. \cite{moore_case_2019} argues that there is no social void from the removal of the driver that needs to be filled by additional communication. While previous research points to eHMIs shortening the length of crossing events by around 1 second \cite{guo_video-based_2022}, one could argue this time difference is not enough to justify the implementation costs of the technology.

In order to both test and exemplify the use of this evaluation method, as well as to provide a baseline for future results, this paper evaluates the kinematic movement of the vehicle \cite{kadali_proactive_2016} as a stand-in eHMI, and answers the proposed questionnaire questions in relation to it. The kinematic movement described is the motion where a vehicle initiates a deceleration followed by a stop, where the vehicle is slightly moved back, as one would normally see in the average internal combustion engine vehicle. An example of this motion can be seen in \cite{toyota_motor_corporation_pedestrian_2020}.

Any time a question would address the display of the eHMI, this test uses the whole left and right side of the vehicle, as this is where the motion of the vehicle is better observed.

\begin{figure}[h]
    \centering
    \includegraphics[width=0.75\linewidth]{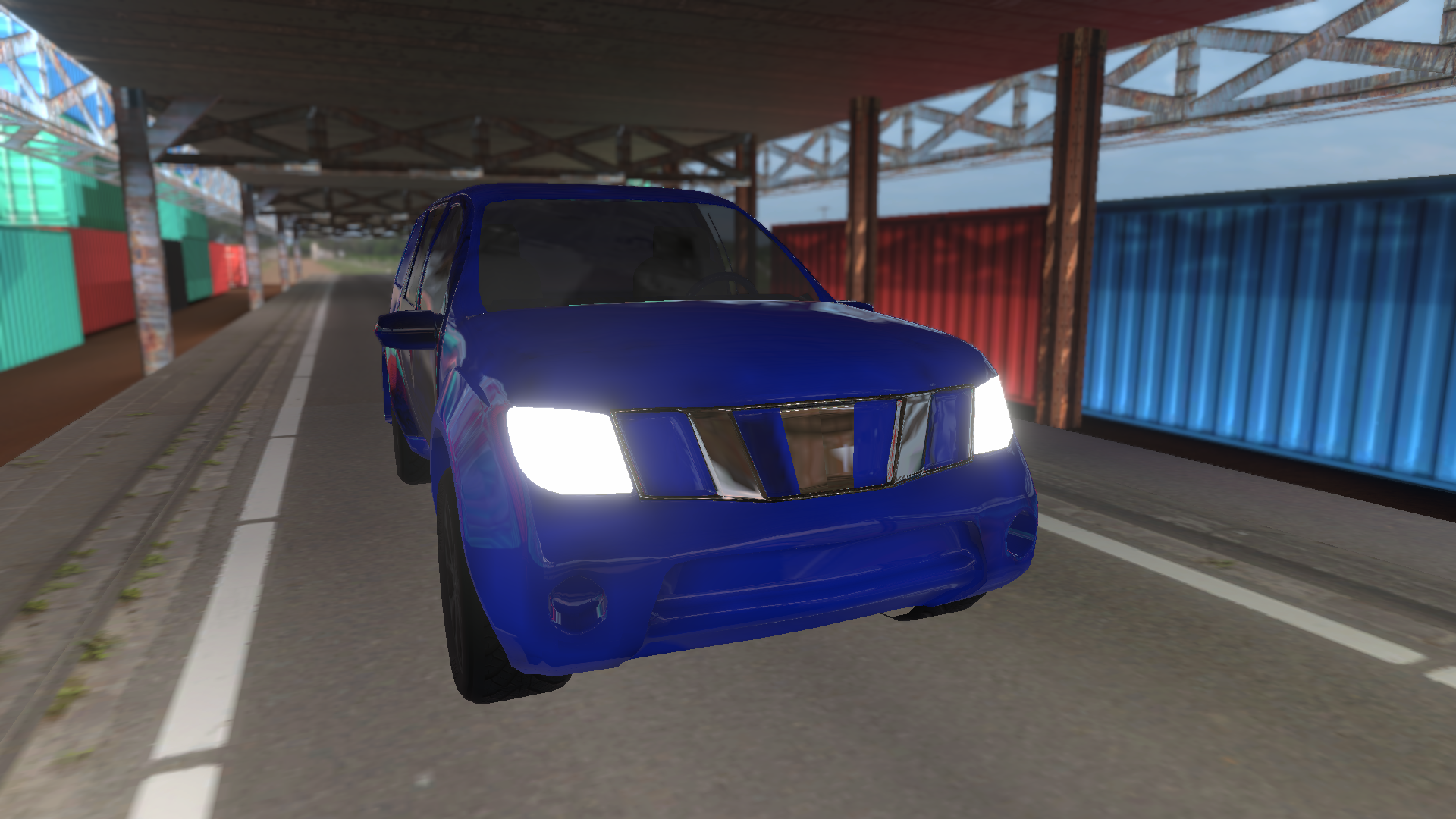}
    \caption{The baseline example with no eHMIs.}
    \label{fig:baseline_example}
\end{figure}

\subsubsection{Frontal Braking Lights (FBL)}
\begin{figure*}[h!]
    \centering
    \includegraphics[width=0.75\linewidth]{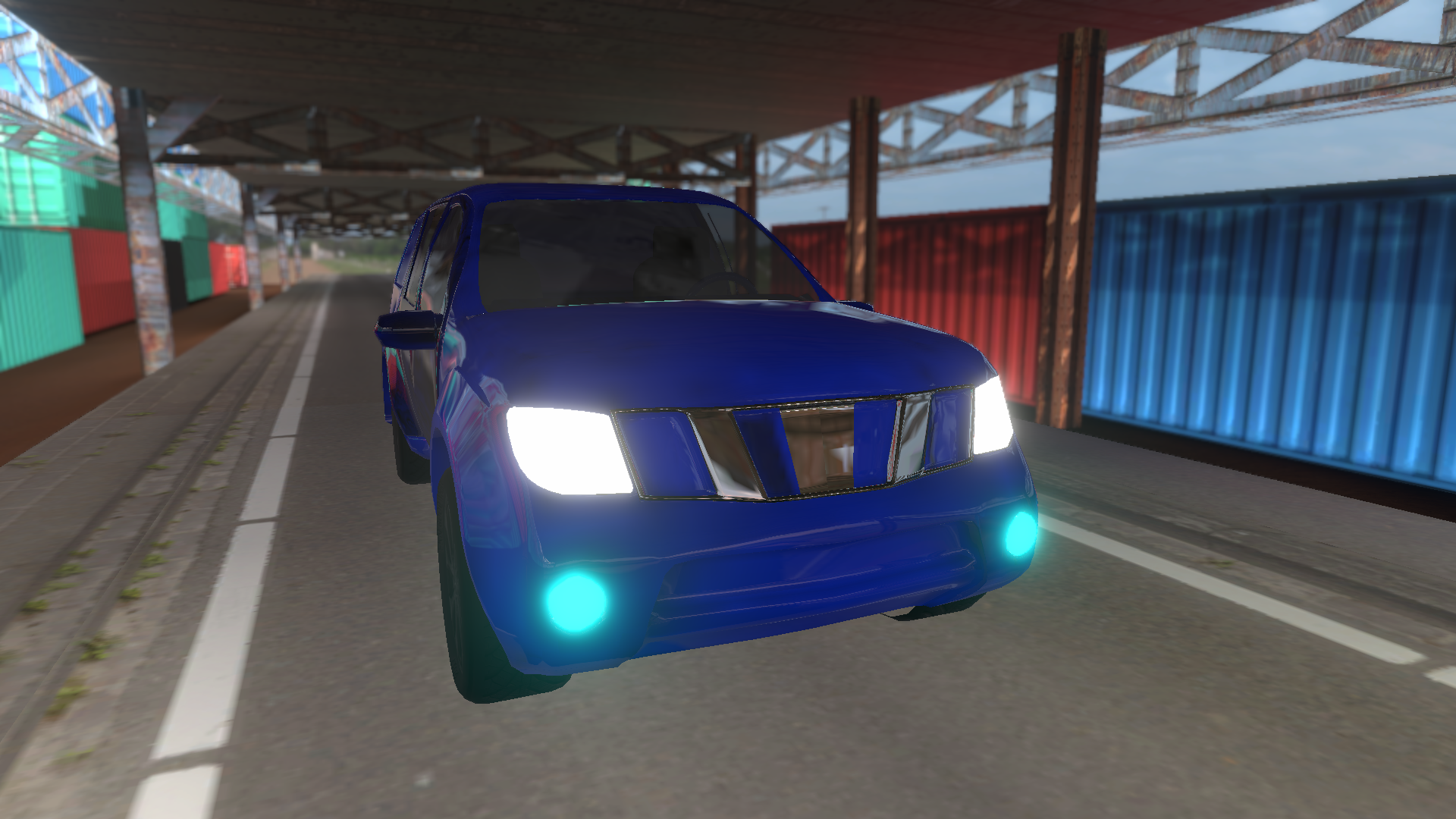}
    \caption{Version of frontal braking lights being evaluated.}
    \label{fig:fbl_example}
\end{figure*}

A technology constantly discussed and reinvented for over a century \cite{de_clercq_external_2019, poliak_evaluation_2023, petzoldt_potential_2018, veach_front_2005, hochel_green_nodate}. This proposal adds an additional pair of lights in the front of the AV, near or under the headlights. These lights remain turned off unless the brakes are engaged, in which case they light up in different intensity levels based on how much the vehicle is slowing down, much like rear brake lights do in current vehicles. An example can be seen in Fig. \ref{fig:fbl_example}.

\subsubsection{“Knight Rider” Display (KRD)}
\begin{figure}
    \centering
    \includegraphics[width=0.75\linewidth]{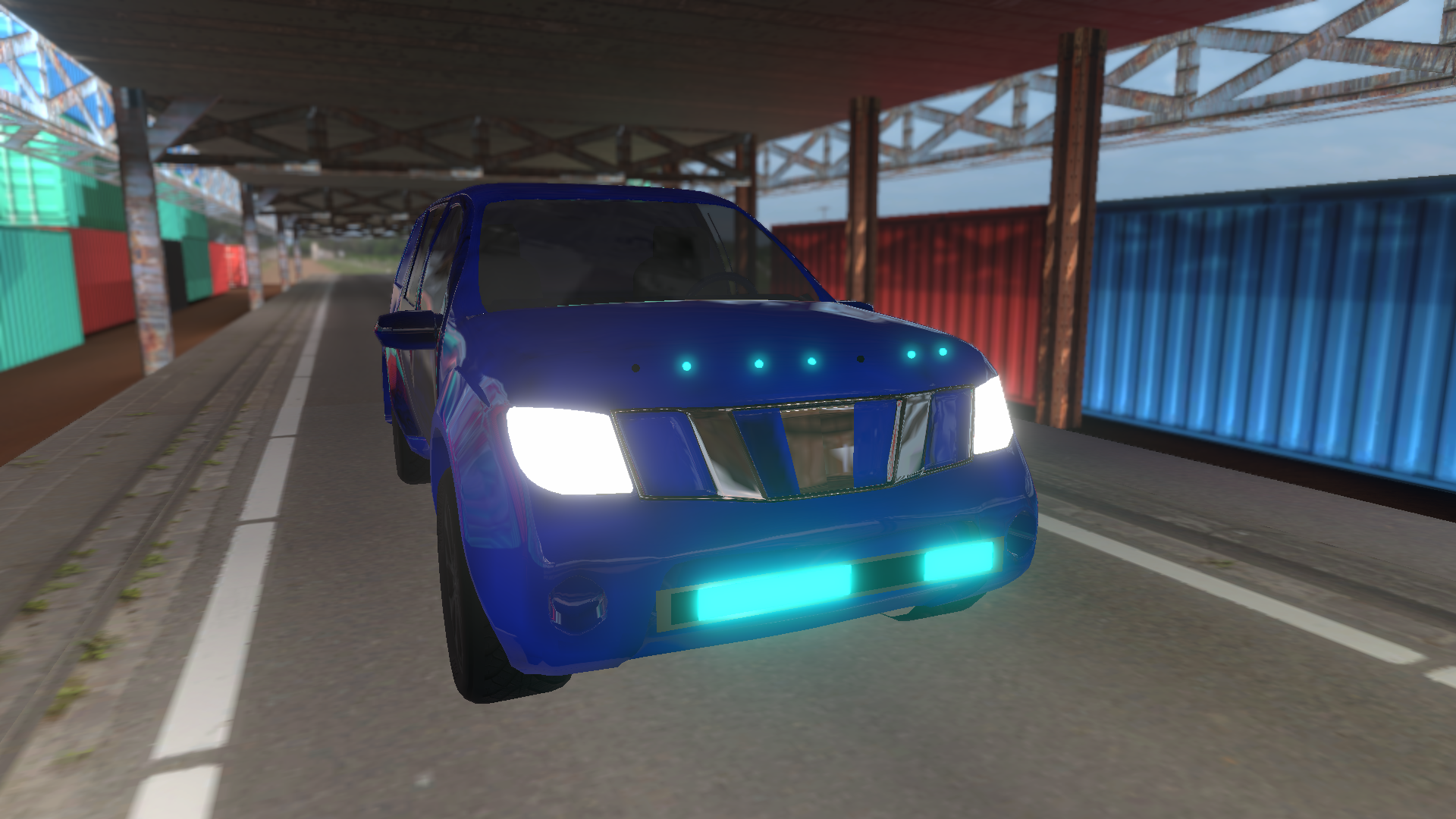}
    \caption{Example of the Knight Rider display mid-animation.}
    \label{fig:kr_example}
\end{figure}

Described in a 2017 article by Ford \cite{noauthor_ford_2017}, then used for the experiments conducted in \cite{de_clercq_external_2019} and \cite{guo_video-based_2022}. This eHMI makes use of a single bar of LED lights that maintains solid luminance during normal driving, and swipes inwards and outwards for 0.5s when it indicates intent to stop. Additionally, in order to both provide a full example of the standardization questionnaire, as well as to remain faithful to the specific version of this proposal seen in \cite{de_clercq_external_2019}, this version includes a series of LEDs that turn on in synchronization with the bumper display, as seen in Fig. \ref{fig:kr_example}. For this analysis, the display initiates the sweeping animation as soon as it starts to slow down, and maintains it until seconds before it starts moving again. Both the display and the lights are turned off when the vehicle is parked, when the engine is off, and when the vehicle is not being driven autonomously.

\subsubsection{Bumper Smile Display (BSD)}
\begin{figure*}[h!]
    \centering
    \includegraphics[width=0.75\linewidth]{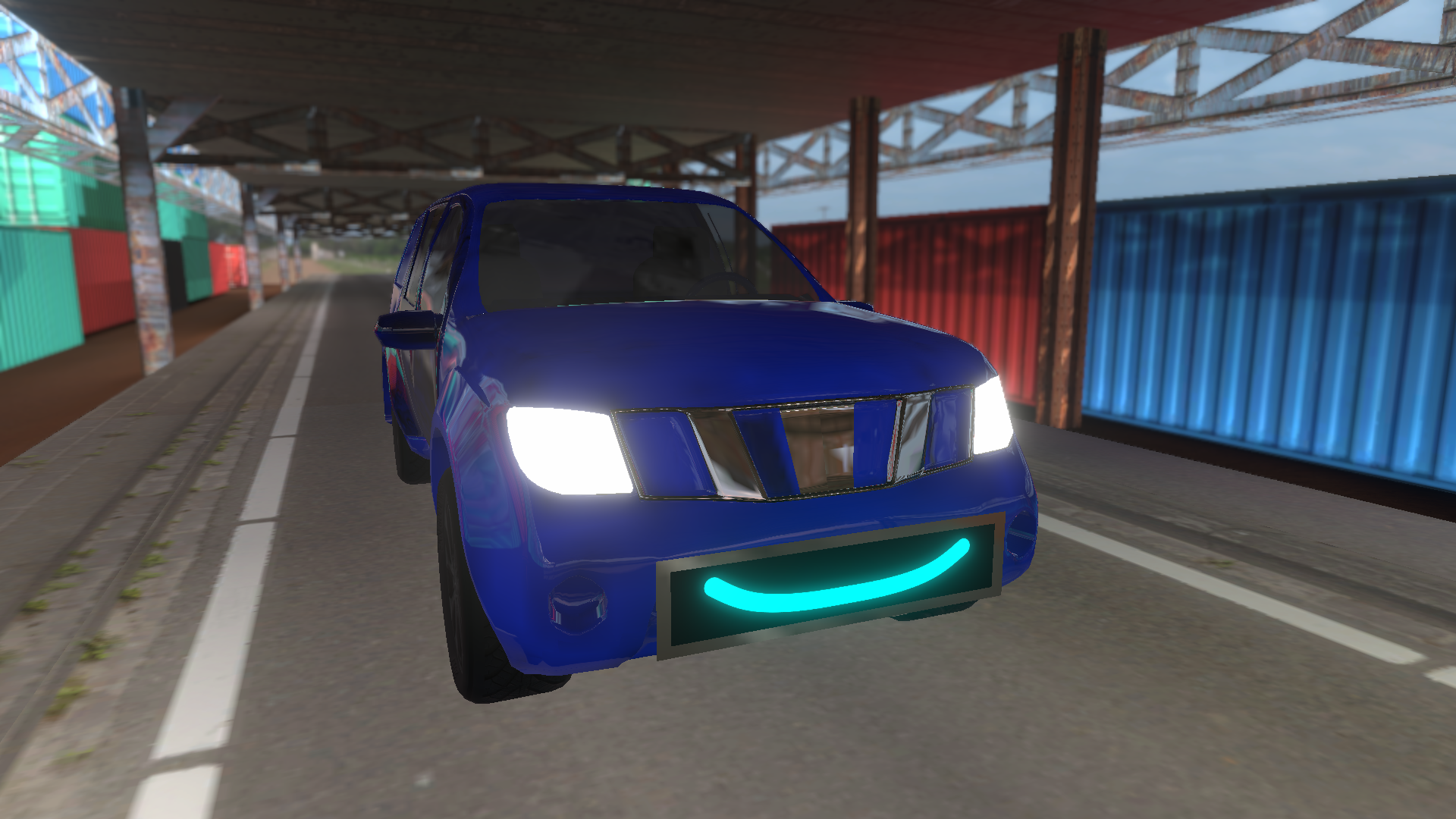}
    \caption{Example of the smiling display, taken from a promotional video of the Smiling Car concept \cite{noauthor_smiling_nodate}.}
    \label{fig:smile_example}
\end{figure*}

A concept produced by Semcon \cite{noauthor_smiling_nodate} to humanize AVs and replace visual contact. As seen on Fig. \ref{fig:smile_example}, the vehicle has an LED display on the front bumper that draws a solid dash-shaped mouth whenever it’s in motion, and a downwardly arched smile when the vehicle has noticed pedestrians, decided to give them right of way, and started to decelerate. The display is turned off when the vehicle is parked, when the engine is off, and when the vehicle is not being driven autonomously. 

\subsubsection{Bumper Text Display (BTD)}
\begin{figure}[h]
    \centering
    \includegraphics[width=0.75\linewidth]{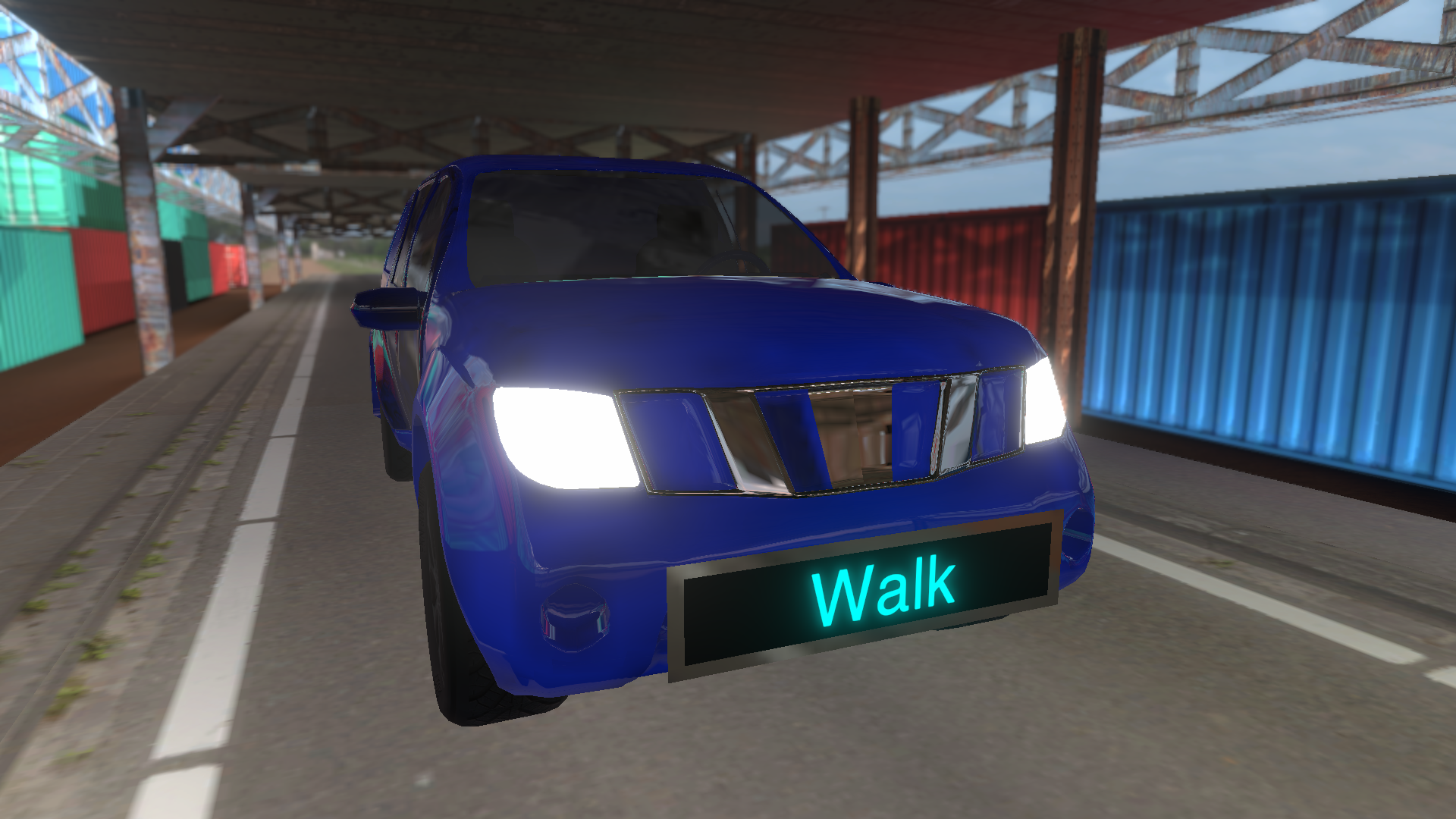}
    \caption{Example of a text display on the front bumper of a vehicle.}
    \label{fig:text_example}
\end{figure}

A common proposal \cite{de_clercq_external_2019, clamann_evaluation_2017, bazilinskyy_survey_2019}, is to communicate vehicle intention (e.g. ``Walk" vs ``Don’t Walk") and status (e.g ``Manual Driving Engaged") using a LED display on the front of the vehicle \cite{clamann_evaluation_2017}, or in the windshield area of the vehicle \cite{noauthor_nissan_2015}. An example can be seen in Fig. \ref{fig:text_example}, making use of the Helvetica font. The display is turned off when the vehicle is parked, when the engine is off, and when the vehicle is not being driven autonomously. 

While this display is capable of communicating additional information through other messages, since the example present in \cite{de_clercq_external_2019} is limited to pedestrian crossing simulation, it only displays two messages for this analysis: ``Don't Walk" when the vehicle is in motion or not yielding, and ``Walk" once the vehicle has noticed pedestrians, decided to give them right of way, and started to decelerate. The text is displayed using the Helvetica font, with a height of 3 inches.

\section{Results and Discussion} \label{Results}

The full series of questionnaires, as well as the method used to calculate each total, can be found in the Appendix. All evaluations are made exclusively assuming that the eHMI proposal mainly indicates to pedestrian when a vehicle is yielding. 

\subsection{Standardization Scores}
Since each proposal may have multiple individual separate eHMIs to communicate information, this section evaluates each proposal independently. None of the proposals make use of sound, and for the purpose of this evaluation all proposals have their position on the vehicle specified.

\subsubsection{Vehicle Kinematics as Indication (No eHMI)} 
This proposal has only one element: The vehicle chassis that shows the kinematic motions of the vehicle as it comes to a stop.
\paragraph{Element 1: Vehicle Chassis} Since the kinematic motion of the vehicle is built into its primary functionality, standardization of it requires no additional work. As seen in Table \ref{tab:result1_table1}, most of the questions were not applicable to this proposal, except for two exceptions:
\begin{itemize}
    \item The proposal relies on kinematic motions to communicate a stopping action, and hence yielding intent. It does not have a kinematic way to indicate when it is not stopping beyond the absence of the stopping behavior, and the acceleration of the vehicle isn't accompanied by other motions to draw attention to it. This means \(Aks = 1\), which results in \(S_{23_{{No eHMI}_1}} = 1\). 
    \item Three  conditions may trigger a change in the stopping motion: Whether the vehicle is stopping or not (counted as one condition), the distance between the vehicle and it's stopping destination, and the deceleration rate of the vehicle. This results in \(S_{27_{{No eHMI}_1}} = 3\). 
\end{itemize}

\begin{table}[h]
\centering
\caption{Standardization Questionnaire Results - Vehicle Kinematics}
\label{tab:result1_table1}
\begin{tabular}{|lllrllllll|}
\hline
\multicolumn{10}{|c|}{\textbf{NO EHMI: ELEMENT 1 (VEHICLE CHASSIS)}} \\ \hline
\multicolumn{1}{|l|}{\textbf{S1}} &
  \multicolumn{1}{l|}{\textbf{S2}} &
  \multicolumn{1}{l|}{\textbf{S3}} &
  \multicolumn{1}{l|}{\textbf{S4}} &
  \multicolumn{1}{l|}{\textbf{S5}} &
  \multicolumn{1}{l|}{\textbf{S6}} &
  \multicolumn{1}{l|}{\textbf{S7}} &
  \multicolumn{1}{l|}{\textbf{S8}} &
  \multicolumn{1}{l|}{\textbf{S9}} &
  \textbf{S10} \\ \hline
\multicolumn{1}{|r|}{0} &
  \multicolumn{1}{r|}{0} &
  \multicolumn{1}{r|}{0} &
  \multicolumn{1}{r|}{0} &
  \multicolumn{1}{r|}{0} &
  \multicolumn{1}{r|}{0} &
  \multicolumn{1}{r|}{0} &
  \multicolumn{1}{r|}{0} &
  \multicolumn{1}{r|}{0} &
  \multicolumn{1}{r|}{0} \\ \hline
\multicolumn{1}{|l|}{\textbf{S11}} &
  \multicolumn{1}{l|}{\textbf{S12}} &
  \multicolumn{1}{l|}{\textbf{S13}} &
  \multicolumn{1}{l|}{\textbf{S14}} &
  \multicolumn{1}{l|}{\textbf{S15}} &
  \multicolumn{1}{l|}{\textbf{S16}} &
  \multicolumn{1}{l|}{\textbf{S17}} &
  \multicolumn{1}{l|}{\textbf{S18}} &
  \multicolumn{1}{l|}{\textbf{S19}} &
  \textbf{S20} \\ \hline
\multicolumn{1}{|r|}{0} &
  \multicolumn{1}{r|}{0} &
  \multicolumn{1}{r|}{0} &
  \multicolumn{1}{r|}{0} &
  \multicolumn{1}{r|}{0} &
  \multicolumn{1}{r|}{0} &
  \multicolumn{1}{r|}{0} &
  \multicolumn{1}{r|}{0} &
  \multicolumn{1}{r|}{0} &
  \multicolumn{1}{r|}{0} \\ \hline
\multicolumn{1}{|l|}{\textbf{S21}} &
  \multicolumn{1}{l|}{\textbf{S22}} &
  \multicolumn{1}{l|}{\textbf{S23}} &
  \multicolumn{1}{l|}{\textbf{S24}} &
  \multicolumn{1}{l|}{\textbf{S25}} &
  \multicolumn{1}{l|}{\textbf{S26}} &
  \multicolumn{1}{l|}{\textbf{S27}} &
  \multicolumn{3}{l|}{\multirow{2}{*}{}} \\ 
\multicolumn{1}{|r|}{0} &
  \multicolumn{1}{r|}{0} &
  \multicolumn{1}{r|}{1} &
  \multicolumn{1}{r|}{0} &
  \multicolumn{1}{r|}{0} &
  \multicolumn{1}{r|}{0} &
  \multicolumn{1}{r|}{3} &
  \multicolumn{3}{l|}{} \\ \hline
\multicolumn{3}{|l|}{\textbf{Penalty Total}} &
  \multicolumn{7}{r|}{4} \\ \hline
\end{tabular}
\end{table}

\subsubsection{Front Braking Lights (FBL)} 
Made up of two lights, identical to each other, so they are counted as one eHMI.
\paragraph{Element 1: Lights} 
For the purposes of this evaluation, the lights were treated as "Pictograms". This proposal does not use frames, background, text or animations. It contains displays that can be turned off (\(S2 = 1\)); can be manufacturer to produce more than one color of light (\(S11 = 1\)); must decide when to turn on based on yielding decision and distance from the pedestrian (\(S12 = 2\)); and has two possible states for yielding and unyielding (\(S15 = 2 - 1\)). 

\begin{table}[h]
\centering
\caption{Standardization Questionnaire Results - Frontal Braking Lights}
\label{tab:result1_table2}
\begin{tabular}{|lllrllllll|}
\hline
\multicolumn{10}{|c|}{\textbf{FRONT BRAKING LIGHTS: ELEMENT 1 (LIGHTS)}} \\ \hline
\multicolumn{1}{|l|}{\textbf{S1}} &
  \multicolumn{1}{l|}{\textbf{S2}} &
  \multicolumn{1}{l|}{\textbf{S3}} &
  \multicolumn{1}{l|}{\textbf{S4}} &
  \multicolumn{1}{l|}{\textbf{S5}} &
  \multicolumn{1}{l|}{\textbf{S6}} &
  \multicolumn{1}{l|}{\textbf{S7}} &
  \multicolumn{1}{l|}{\textbf{S8}} &
  \multicolumn{1}{l|}{\textbf{S9}} &
  \textbf{S10} \\ \hline
\multicolumn{1}{|r|}{0} &
  \multicolumn{1}{r|}{1} &
  \multicolumn{1}{r|}{0} &
  \multicolumn{1}{r|}{0} &
  \multicolumn{1}{r|}{0} &
  \multicolumn{1}{r|}{0} &
  \multicolumn{1}{r|}{0} &
  \multicolumn{1}{r|}{0} &
  \multicolumn{1}{r|}{0} &
  \multicolumn{1}{r|}{0} \\ \hline
\multicolumn{1}{|l|}{\textbf{S11}} &
  \multicolumn{1}{l|}{\textbf{S12}} &
  \multicolumn{1}{l|}{\textbf{S13}} &
  \multicolumn{1}{l|}{\textbf{S14}} &
  \multicolumn{1}{l|}{\textbf{S15}} &
  \multicolumn{1}{l|}{\textbf{S16}} &
  \multicolumn{1}{l|}{\textbf{S17}} &
  \multicolumn{1}{l|}{\textbf{S18}} &
  \multicolumn{1}{l|}{\textbf{S19}} &
  \textbf{S20} \\ \hline
\multicolumn{1}{|r|}{1} &
  \multicolumn{1}{r|}{2} &
  \multicolumn{1}{r|}{0} &
  \multicolumn{1}{r|}{0} &
  \multicolumn{1}{r|}{1} &
  \multicolumn{1}{r|}{0} &
  \multicolumn{1}{r|}{0} &
  \multicolumn{1}{r|}{0} &
  \multicolumn{1}{r|}{0} &
  \multicolumn{1}{r|}{0} \\ \hline
\multicolumn{1}{|l|}{\textbf{S21}} &
  \multicolumn{1}{l|}{\textbf{S22}} &
  \multicolumn{1}{l|}{\textbf{S23}} &
  \multicolumn{1}{l|}{\textbf{S24}} &
  \multicolumn{1}{l|}{\textbf{S25}} &
  \multicolumn{1}{l|}{\textbf{S26}} &
  \multicolumn{1}{l|}{\textbf{S27}} &
  \multicolumn{3}{l|}{\multirow{2}{*}{}} \\ 
\multicolumn{1}{|r|}{0} &
  \multicolumn{1}{r|}{0} &
  \multicolumn{1}{r|}{0} &
  \multicolumn{1}{r|}{0} &
  \multicolumn{1}{r|}{0} &
  \multicolumn{1}{r|}{0} &
  \multicolumn{1}{r|}{0} &
  \multicolumn{3}{l|}{} \\ \hline
\multicolumn{3}{|l|}{\textbf{Penalty Total}} &
  \multicolumn{7}{r|}{5} \\ \hline
\end{tabular}
\end{table}

\subsubsection{Knight Rider Display (KRD)}
Made up of two elements: One display on the front bumper, and seven LED lights on the hood that light up in synchronicity with the bumper display.
\paragraph{Element 1: Display} 
For the purpose of this evaluation, the sweeping animation on the display is treated as a pictogram. It does not make use of text. It contains a display that can be turned off (\(S2 = 1\)); has a frame that can be made in more than one color (\(S4 = 1\)) and thickness (\(S5 = 1\)); has a background to the lights that can be made in more than one color (\(S7 = 1\)); may change its animation based on yielding decision and distance from the stopping position (\(S10 = 2\)); the display itself can be made in a variety of colors (\(S11 = 1\)); and has two possible states for yielding and unyielding (\(S15 = 2 - 1\)). 
Additionally, it features one intermittent animation (\(Arc = 1, Aks = 0\)) that loops every 0.5 second (\(Ams = 500\)) resulting in \(S23 = \frac{500}{1000} + 0 + 1 = 1.5\) \cite{de_clercq_external_2019}.

\begin{table}[h]
\centering
\caption{Standardization Questionnaire Results - Knight Rider Screen Display}
\label{tab:result1_table3}
\begin{tabular}{|lllrllllll|}
\hline
\multicolumn{10}{|c|}{\textbf{KNIGHT RIDER DISPLAY: ELEMENT 1 (DISPLAY)}} \\ \hline
\multicolumn{1}{|l|}{\textbf{S1}} &
  \multicolumn{1}{l|}{\textbf{S2}} &
  \multicolumn{1}{l|}{\textbf{S3}} &
  \multicolumn{1}{l|}{\textbf{S4}} &
  \multicolumn{1}{l|}{\textbf{S5}} &
  \multicolumn{1}{l|}{\textbf{S6}} &
  \multicolumn{1}{l|}{\textbf{S7}} &
  \multicolumn{1}{l|}{\textbf{S8}} &
  \multicolumn{1}{l|}{\textbf{S9}} &
  \textbf{S10} \\ \hline
\multicolumn{1}{|r|}{0} &
  \multicolumn{1}{r|}{1} &
  \multicolumn{1}{r|}{0} &
  \multicolumn{1}{r|}{1} &
  \multicolumn{1}{r|}{1} &
  \multicolumn{1}{r|}{0} &
  \multicolumn{1}{r|}{1} &
  \multicolumn{1}{r|}{0} &
  \multicolumn{1}{r|}{0} &
  \multicolumn{1}{r|}{2} \\ \hline
\multicolumn{1}{|l|}{\textbf{S11}} &
  \multicolumn{1}{l|}{\textbf{S12}} &
  \multicolumn{1}{l|}{\textbf{S13}} &
  \multicolumn{1}{l|}{\textbf{S14}} &
  \multicolumn{1}{l|}{\textbf{S15}} &
  \multicolumn{1}{l|}{\textbf{S16}} &
  \multicolumn{1}{l|}{\textbf{S17}} &
  \multicolumn{1}{l|}{\textbf{S18}} &
  \multicolumn{1}{l|}{\textbf{S19}} &
  \textbf{S20} \\ \hline
\multicolumn{1}{|r|}{1} &
  \multicolumn{1}{r|}{0} &
  \multicolumn{1}{r|}{0} &
  \multicolumn{1}{r|}{0} &
  \multicolumn{1}{r|}{1} &
  \multicolumn{1}{r|}{0} &
  \multicolumn{1}{r|}{0} &
  \multicolumn{1}{r|}{0} &
  \multicolumn{1}{r|}{0} &
  \multicolumn{1}{r|}{0} \\ \hline
\multicolumn{1}{|l|}{\textbf{S21}} &
  \multicolumn{1}{l|}{\textbf{S22}} &
  \multicolumn{1}{l|}{\textbf{S23}} &
  \multicolumn{1}{l|}{\textbf{S24}} &
  \multicolumn{1}{l|}{\textbf{S25}} &
  \multicolumn{1}{l|}{\textbf{S26}} &
  \multicolumn{1}{l|}{\textbf{S27}} &
  \multicolumn{3}{l|}{\multirow{2}{*}{}} \\ 
\multicolumn{1}{|r|}{0} &
  \multicolumn{1}{r|}{0} &
  \multicolumn{1}{r|}{1.5} &
  \multicolumn{1}{r|}{0} &
  \multicolumn{1}{r|}{0} &
  \multicolumn{1}{r|}{0} &
  \multicolumn{1}{r|}{0} &
  \multicolumn{3}{l|}{} \\ \hline
\multicolumn{3}{|l|}{\textbf{Penalty Total}} &
  \multicolumn{7}{r|}{9.5} \\ \hline
\end{tabular}
\end{table}

\paragraph{Element 2: Lights} 
The group of lights on the hood do not make use of frames, text, or a background. They can be turned off (\(S2 = 1\)); can be manufactured in a variety of colors (\(S11 = 1\)); have two possible states for yielding and unyielding (\(S15 = 2 - 1\)); has the same animation timing as the main display (\(Arc = 1, Ams = 500\)). Its determining condition is whether the main display is in an animated or static state (\(S27 = 1\)).

\begin{table}[h]
\centering
\caption{Standardization Questionnaire Results - Knight Rider LED Lights}
\label{tab:result1_table4}
\begin{tabular}{|lllrllllll|}
\hline
\multicolumn{10}{|c|}{\textbf{KNIGHT RIDER DISPLAY: ELEMENT 2 (LIGHTS)}} \\ \hline
\multicolumn{1}{|l|}{\textbf{S1}} &
  \multicolumn{1}{l|}{\textbf{S2}} &
  \multicolumn{1}{l|}{\textbf{S3}} &
  \multicolumn{1}{l|}{\textbf{S4}} &
  \multicolumn{1}{l|}{\textbf{S5}} &
  \multicolumn{1}{l|}{\textbf{S6}} &
  \multicolumn{1}{l|}{\textbf{S7}} &
  \multicolumn{1}{l|}{\textbf{S8}} &
  \multicolumn{1}{l|}{\textbf{S9}} &
  \textbf{S10} \\ \hline
\multicolumn{1}{|r|}{0} &
  \multicolumn{1}{r|}{1} &
  \multicolumn{1}{r|}{0} &
  \multicolumn{1}{r|}{0} &
  \multicolumn{1}{r|}{0} &
  \multicolumn{1}{r|}{0} &
  \multicolumn{1}{r|}{0} &
  \multicolumn{1}{r|}{0} &
  \multicolumn{1}{r|}{0} &
  \multicolumn{1}{r|}{0} \\ \hline
\multicolumn{1}{|l|}{\textbf{S11}} &
  \multicolumn{1}{l|}{\textbf{S12}} &
  \multicolumn{1}{l|}{\textbf{S13}} &
  \multicolumn{1}{l|}{\textbf{S14}} &
  \multicolumn{1}{l|}{\textbf{S15}} &
  \multicolumn{1}{l|}{\textbf{S16}} &
  \multicolumn{1}{l|}{\textbf{S17}} &
  \multicolumn{1}{l|}{\textbf{S18}} &
  \multicolumn{1}{l|}{\textbf{S19}} &
  \textbf{S20} \\ \hline
\multicolumn{1}{|r|}{1} &
  \multicolumn{1}{r|}{0} &
  \multicolumn{1}{r|}{0} &
  \multicolumn{1}{r|}{0} &
  \multicolumn{1}{r|}{1} &
  \multicolumn{1}{r|}{0} &
  \multicolumn{1}{r|}{0} &
  \multicolumn{1}{r|}{0} &
  \multicolumn{1}{r|}{0} &
  \multicolumn{1}{r|}{0} \\ \hline
\multicolumn{1}{|l|}{\textbf{S21}} &
  \multicolumn{1}{l|}{\textbf{S22}} &
  \multicolumn{1}{l|}{\textbf{S23}} &
  \multicolumn{1}{l|}{\textbf{S24}} &
  \multicolumn{1}{l|}{\textbf{S25}} &
  \multicolumn{1}{l|}{\textbf{S26}} &
  \multicolumn{1}{l|}{\textbf{S27}} &
  \multicolumn{3}{l|}{\multirow{2}{*}{}} \\ 
\multicolumn{1}{|r|}{0} &
  \multicolumn{1}{r|}{0} &
  \multicolumn{1}{r|}{1.5} &
  \multicolumn{1}{r|}{0} &
  \multicolumn{1}{r|}{0} &
  \multicolumn{1}{r|}{0} &
  \multicolumn{1}{r|}{1} &
  \multicolumn{3}{l|}{} \\ \hline
\multicolumn{3}{|l|}{\textbf{Penalty Total}} &
  \multicolumn{7}{r|}{5.5} \\ \hline
\end{tabular}
\end{table}

\subsubsection{Bumper Smile Display (BSD)}
Made up of one element: A display on the front bumper that shows a grin when the vehicle is yielding, and maintains a straight line that simulates a neutral expression otherwise.
\paragraph{Element 1: Display}
The smile is a pictogram. The display does not make use of text. It can be turned off (\(S2 = 1\)); has a frame that can be made in more than one color (\(S4 = 1\)) and thickness (\(S5 = 1\)); has a background to the lights that can be made in more than one color (\(S7 = 1\)); may change its pictogram  based on yielding decision and distance from the stopping position (\(S10 = 2\)); the display itself can be made in a variety of colors (\(S11 = 1\)); and has two possible states for yielding and unyielding (\(S15 = 2 - 1\)). 

\begin{table}[h]
\centering
\caption{Standardization Questionnaire Results - Bumper Smile Display}
\label{tab:result1_table5}
\begin{tabular}{|lllrllllll|}
\hline
\multicolumn{10}{|c|}{\textbf{BUMPER SMILE DISPLAY: ELEMENT 1 (DISPLAY)}} \\ \hline
\multicolumn{1}{|l|}{\textbf{S1}} &
  \multicolumn{1}{l|}{\textbf{S2}} &
  \multicolumn{1}{l|}{\textbf{S3}} &
  \multicolumn{1}{l|}{\textbf{S4}} &
  \multicolumn{1}{l|}{\textbf{S5}} &
  \multicolumn{1}{l|}{\textbf{S6}} &
  \multicolumn{1}{l|}{\textbf{S7}} &
  \multicolumn{1}{l|}{\textbf{S8}} &
  \multicolumn{1}{l|}{\textbf{S9}} &
  \textbf{S10} \\ \hline
\multicolumn{1}{|r|}{0} &
  \multicolumn{1}{r|}{1} &
  \multicolumn{1}{r|}{0} &
  \multicolumn{1}{r|}{1} &
  \multicolumn{1}{r|}{1} &
  \multicolumn{1}{r|}{0} &
  \multicolumn{1}{r|}{1} &
  \multicolumn{1}{r|}{0} &
  \multicolumn{1}{r|}{0} &
  \multicolumn{1}{r|}{2} \\ \hline
\multicolumn{1}{|l|}{\textbf{S11}} &
  \multicolumn{1}{l|}{\textbf{S12}} &
  \multicolumn{1}{l|}{\textbf{S13}} &
  \multicolumn{1}{l|}{\textbf{S14}} &
  \multicolumn{1}{l|}{\textbf{S15}} &
  \multicolumn{1}{l|}{\textbf{S16}} &
  \multicolumn{1}{l|}{\textbf{S17}} &
  \multicolumn{1}{l|}{\textbf{S18}} &
  \multicolumn{1}{l|}{\textbf{S19}} &
  \textbf{S20} \\ \hline
\multicolumn{1}{|r|}{1} &
  \multicolumn{1}{r|}{0} &
  \multicolumn{1}{r|}{0} &
  \multicolumn{1}{r|}{0} &
  \multicolumn{1}{r|}{1} &
  \multicolumn{1}{r|}{0} &
  \multicolumn{1}{r|}{0} &
  \multicolumn{1}{r|}{0} &
  \multicolumn{1}{r|}{0} &
  \multicolumn{1}{r|}{0} \\ \hline
\multicolumn{1}{|l|}{\textbf{S21}} &
  \multicolumn{1}{l|}{\textbf{S22}} &
  \multicolumn{1}{l|}{\textbf{S23}} &
  \multicolumn{1}{l|}{\textbf{S24}} &
  \multicolumn{1}{l|}{\textbf{S25}} &
  \multicolumn{1}{l|}{\textbf{S26}} &
  \multicolumn{1}{l|}{\textbf{S27}} &
  \multicolumn{3}{l|}{\multirow{2}{*}{}} \\ 
\multicolumn{1}{|r|}{0} &
  \multicolumn{1}{r|}{0} &
  \multicolumn{1}{r|}{0} &
  \multicolumn{1}{r|}{0} &
  \multicolumn{1}{r|}{0} &
  \multicolumn{1}{r|}{0} &
  \multicolumn{1}{r|}{0} &
  \multicolumn{3}{l|}{} \\ \hline
\multicolumn{3}{|l|}{\textbf{Penalty Total}} &
  \multicolumn{7}{r|}{8} \\ \hline
\end{tabular}
\end{table}

\subsubsection{Bumper Text Display (BTD)}
Made up of one element: A display on the front bumper that can show text indicating the intent of the vehicle.
\paragraph{Element 1: Display}
The display does not make use of pictograms. It can be turned off (\(S2 = 1\)); uses text that needs standardization of font, letter spacing, color, and size (\(S3 = 4\)); has a frame that can be made in more than one color (\(S4 = 1\)) and thickness (\(S5 = 1\)); has a background to the text that can be made in more than one color (\(S7 = 1\)); may change its text content based on yielding decision and distance from the stopping position (\(S16 = 2\)); the display itself can be made in a variety of colors (\(S17 = 1\)); and has two possible states for yielding and unyielding (\(S18 = 2 - 1\)). 

\begin{table}[h]
\centering
\caption{Standardization Questionnaire Results - Bumper Text Display}
\label{tab:result1_table6}
\begin{tabular}{|lllrllllll|}
\hline
\multicolumn{10}{|c|}{\textbf{BUMPER TEXT DISPLAY: ELEMENT 1 (DISPLAY)}} \\ \hline
\multicolumn{1}{|l|}{\textbf{S1}} &
  \multicolumn{1}{l|}{\textbf{S2}} &
  \multicolumn{1}{l|}{\textbf{S3}} &
  \multicolumn{1}{l|}{\textbf{S4}} &
  \multicolumn{1}{l|}{\textbf{S5}} &
  \multicolumn{1}{l|}{\textbf{S6}} &
  \multicolumn{1}{l|}{\textbf{S7}} &
  \multicolumn{1}{l|}{\textbf{S8}} &
  \multicolumn{1}{l|}{\textbf{S9}} &
  \textbf{S10} \\ \hline
\multicolumn{1}{|r|}{0} &
  \multicolumn{1}{r|}{1} &
  \multicolumn{1}{r|}{4} &
  \multicolumn{1}{r|}{1} &
  \multicolumn{1}{r|}{1} &
  \multicolumn{1}{r|}{0} &
  \multicolumn{1}{r|}{1} &
  \multicolumn{1}{r|}{0} &
  \multicolumn{1}{r|}{0} &
  \multicolumn{1}{r|}{0} \\ \hline
\multicolumn{1}{|l|}{\textbf{S11}} &
  \multicolumn{1}{l|}{\textbf{S12}} &
  \multicolumn{1}{l|}{\textbf{S13}} &
  \multicolumn{1}{l|}{\textbf{S14}} &
  \multicolumn{1}{l|}{\textbf{S15}} &
  \multicolumn{1}{l|}{\textbf{S16}} &
  \multicolumn{1}{l|}{\textbf{S17}} &
  \multicolumn{1}{l|}{\textbf{S18}} &
  \multicolumn{1}{l|}{\textbf{S19}} &
  \textbf{S20} \\ \hline
\multicolumn{1}{|r|}{0} &
  \multicolumn{1}{r|}{0} &
  \multicolumn{1}{r|}{0} &
  \multicolumn{1}{r|}{0} &
  \multicolumn{1}{r|}{0} &
  \multicolumn{1}{r|}{2} &
  \multicolumn{1}{r|}{0} &
  \multicolumn{1}{r|}{0} &
  \multicolumn{1}{r|}{0} &
  \multicolumn{1}{r|}{0} \\ \hline
\multicolumn{1}{|l|}{\textbf{S21}} &
  \multicolumn{1}{l|}{\textbf{S22}} &
  \multicolumn{1}{l|}{\textbf{S23}} &
  \multicolumn{1}{l|}{\textbf{S24}} &
  \multicolumn{1}{l|}{\textbf{S25}} &
  \multicolumn{1}{l|}{\textbf{S26}} &
  \multicolumn{1}{l|}{\textbf{S27}} &
  \multicolumn{3}{l|}{\multirow{2}{*}{}} \\ 
\multicolumn{1}{|r|}{0} &
  \multicolumn{1}{r|}{1} &
  \multicolumn{1}{r|}{0} &
  \multicolumn{1}{r|}{0} &
  \multicolumn{1}{r|}{0} &
  \multicolumn{1}{r|}{0} &
  \multicolumn{1}{r|}{0} &
  \multicolumn{3}{l|}{} \\ \hline
\multicolumn{3}{|l|}{\textbf{Penalty Total}} &
  \multicolumn{7}{r|}{11} \\ \hline
\end{tabular}
\end{table}

\subsubsection{Results}
Since vehicle kinematics, by definition, require no additional eHMI standardization, the standardization baseline value is calculated using Equation (\ref{BaselineCalculation:12}), where \({S_{count}} = 27\) and \({S_{P_{no eHMI}}} = 4\).
After adding up the penalties across all elements for each proposal using Equation (\ref{SPx:1}) and calculating the Standardization Score for each using Equation (\ref{S_S:2}), the results show Frontal Brake Lights and Bumper Smile Display as the easiest proposals to standardize after the vehicle kinematics, while the Knight Rider Display scores lowest due to its multiple non-identical eHMIs. 

\begin{table}[]
\centering
\caption{Standardization Questionnaire Results - Standardization Results}
\label{tab:result1_table_total}
\begin{tabular}{|lcrrrr|}
\hline
\multicolumn{6}{|c|}{\textbf{STANDARDIZATION RESULTS}}            \\ \hline
\multicolumn{1}{|l|}{\textbf{Baseline}} & \multicolumn{5}{c|}{31} \\ \hline
\multicolumn{1}{|l|}{} &
  \multicolumn{1}{c|}{\textbf{No eHMI}} &
  \multicolumn{1}{c|}{\textbf{FBL}} &
  \multicolumn{1}{c|}{\textbf{KRD}} &
  \multicolumn{1}{c|}{\textbf{BSD}} &
  \multicolumn{1}{c|}{\textbf{BTD}} \\ \hline
\multicolumn{1}{|l|}{\textbf{Penalty Total (All Elements)}} &
  \multicolumn{1}{r|}{4.00} &
  \multicolumn{1}{r|}{5.00} &
  \multicolumn{1}{r|}{15.00} &
  \multicolumn{1}{r|}{8.00} &
  11.00 \\ \hline
\multicolumn{1}{|l|}{\textbf{TOTAL}} &
  \multicolumn{1}{r|}{10.00} &
  \multicolumn{1}{r|}{9.63} &
  \multicolumn{1}{r|}{5.93} &
  \multicolumn{1}{r|}{8.52} &
  7.41 \\ \hline
\end{tabular}
\end{table}

\subsection{Cost Effectiveness Scores}
None of the proposals outline a manufacturing, installation, maintenance, or operation cost. To nonetheless present a frame of reference, this section uses information available on similar technology to produce an estimate. All adjustment for inflation were made using the CPI Inflation Calculator by the United Stated Bureau of Labor Statistics \cite{us_bureau_of_labor_statistics_cpi_nodate}, using December 2022 as the target date. 
As stated in the Appendix, whenever a range is given, the value used is the median of that range, and whenever the answer is unknown, the highest value in the questionnaire for that eHMI is used instead. 

\subsubsection{Technologies}

\paragraph{No eHMI} 
Since the kinematics of a vehicle are part of the normal behavior of modern vehicles, there is no additional cost to implement it, maintain it, or develop it.

\begin{table}[h]
\centering
\caption{Cost Effectiveness Questionnaire Results - No eHMI}
\label{tab:result2_table1}
\begin{tabular}{|lr|}
\hline
\multicolumn{2}{|c|}{\textbf{NO EHMI RESULTS}}    \\ \hline
\multicolumn{1}{|l|}{\textbf{CE1\_NoeHMI}} & 0.00 \\ \hline
\multicolumn{1}{|l|}{\textbf{CE2\_NoeHMI}} & 0.00 \\ \hline
\multicolumn{1}{|l|}{\textbf{CE3\_NoeHMI}} & 0.00 \\ \hline
\multicolumn{1}{|l|}{\textbf{CE4\_NoeHMI}} & 0.00 \\ \hline
\multicolumn{1}{|l|}{\textbf{CE5\_NoeHMI}} & 0.00 \\ \hline
\end{tabular}
\end{table}

\paragraph{Tail Lights} 
Frontal Brake Lights use the same technology as tail braking lights. The cheapest replacement for a single tail brake light with light bulb in 2024 is \$20.99 in carparts.com \cite{noauthor_carpartscom_nodate}. After accounting for both lights, and adjusting for inflation, it makes \({CE_{1_{Brake Light}}} = 39.68\). 
Repair Pal estimates the labor cost of replacing a tail light to be between \$40 and \$51 \cite{noauthor_brake_nodate}, which makes \(CE_{3_{Brake Light}} = 45.50\) after calculating the median and adjusting for inflation. The manufacturing cost is not disclosed, forcing us to use the same cost as an existing vehicle, accounting for the difference in prevalence \(CE_{2_{Brake Light}} = CE_{3_{Brake Light}} \times 0.75 = 34.13\).
A car light bulb has a lifetime of between 14 and 42 months, according to Michael \cite{michael_how_2019}. By using the following equation

\begin{equation} \label{yearly_cost_equation}
    CE_4 = \frac{CE_1 + \frac{(CE_2 + CE_3)}{2}}{{Median Lifetime in Years}}
\end{equation}
where \({Median Lifetime in Years} \approx 2.33\), we can conclude that \(CE_{4_{Brake Light}} \approx 53.94\).
Information on the operating costs of tail lights was not available, so the highest value in this category was chosen, making \(CE_{5_{Brake Light}} = 53.94\).

\begin{table}[h]
\centering
\caption{Cost Effectiveness Questionnaire Results - Tail Lights}
\label{tab:result2_table2}
\begin{tabular}{|lr|}
\hline
\multicolumn{2}{|c|}{\textbf{TRAIL LIGHTS RESULTS}}    \\ \hline
\multicolumn{1}{|l|}{\textbf{CE1\_BrakeLight}} & 39.68 \\ \hline
\multicolumn{1}{|l|}{\textbf{CE2\_BrakeLight}} & 34.13 \\ \hline
\multicolumn{1}{|l|}{\textbf{CE3\_BrakeLight}} & 45.50 \\ \hline
\multicolumn{1}{|l|}{\textbf{CE4\_BrakeLight}} & 53.94 \\ \hline
\multicolumn{1}{|l|}{\textbf{CE5\_BrakeLight}} & 53.94 \\ \hline
\end{tabular}
\end{table}

\paragraph{LED Displays}
The Knight Rider Display, Bumper Smile Display, and Bumper Text Display all make use of LED screens to show pictograms or text to the observer. These can be compared to LED advertisement signs placed on vehicles, as they both must endure the elements and be visible to observers. 

According to LED Sign City \cite{ledsigncity_car_nodate}, the cost of a single Taxi Top LED sign ranges from \$3900.00 and \$4875.00. After calculating the median and adjusting for inflation, this makes \(CE_{1_{LED Display}} = 4606.90\).

LED Sign City includes installation cost with the purchase, making \(CE_{2_{LED Display}} = 0.00\) and \(CE_{3_{LED Display}} = 0.00\).

These signs have a warranty of 2 years. Using Equation (\ref{yearly_cost_equation}),  \(CE_{4_{LED Display}} = 2303.45\).
The LED signs in \cite{ledsigncity_car_nodate} require up to 570 watts. According to Gross, the average person living in the United States spends at least 17600 minutes \(\approx293.33 hours\) driving each year \cite{gross_americans_2016}. Since all examples making use of this display have it on at all times during driving, we calculated the energy consumption of the display by using the equation 
\begin{equation} \label{W to kWh}
    {kWh} = \frac{W \times {hours}}{1000}
\end{equation}
where \(W = 570\) and \({hours} \approx 293.33\). The operation cost was then calculated using
    \begin{equation} \label{kWh to usd}
        {CE5_{x}} = {kWh} \times {rate_{2022}}
    \end{equation}
where \({rate_{2022}} = 0.165 \frac{USD}{kWh}\), which is the average price of a kWh in the United States in December of 2022 according to the United States Bureau of Labor Statistics \cite{us_bureau_of_labor_statistics_average_2022}, making \(CE_{5_{LED Display}} \approx 27.59\)

\begin{table}[h]
\centering
\caption{Cost Effectiveness Questionnaire Results - LED Display}
\label{tab:result2_table3}
\begin{tabular}{|lr|}
\hline
\multicolumn{2}{|c|}{\textbf{LED DISPLAYS RESULTS}}       \\ \hline
\multicolumn{1}{|l|}{\textbf{CE1\_LEDDisplay}} & 4,606.90 \\ \hline
\multicolumn{1}{|l|}{\textbf{CE2\_LEDDisplay}} & 0.00     \\ \hline
\multicolumn{1}{|l|}{\textbf{CE3\_LEDDisplay}} & 0.00     \\ \hline
\multicolumn{1}{|l|}{\textbf{CE4\_LEDDisplay}} & 2,303.45 \\ \hline
\multicolumn{1}{|l|}{\textbf{CE5\_LEDDisplay}} & 27.59    \\ \hline
\end{tabular}
\end{table}

\paragraph{LED Light Array}
The Knight Rider display makes use of seven additional LED lights that work in conjunction with its main display. These can be compared to 5mm color LED lamps, like the ones in LED Supply \cite{noauthor_5mm_nodate}.
Each lamp costs \$1.20 each, which can be adjusted for inflation to \(cost_{LED Light} = 1.13\). Using the following equation
    \begin{equation} \label{cost_of_led_lights}
        {CE_{1_{LED Light Array}}} = 7 \times {cost_{LED Light}}
    \end{equation}
results in \(CE_{1_{LED Light Array}} = 7.91\).

To calculate the yearly operating cost of a single LED light bulb we need to find its wattage first using the equation
    \begin{equation} \label{power_equation}
        W = {Voltage} \times {Current}
    \end{equation}
where \(Voltage = 3.5V\) and \(Current = 0.03A\), resulting in \(W = 0.105W\). The energy used by the light was calculated by applying Equation (\ref{W to kWh}). Unlike the LED display, the lights are not on when the vehicle is in motion, but since no data can be found on how much time a vehicle would spend telling a pedestrian to cross we used the same value for time as with the display, which is to say \({hours} \approx 293.33\), this results in \(kWh \approx 0.03kWh\). Using the same value for \(rate\) as before and Equation (\ref{kWh to usd}), we determined that the average yearly operating cost of a single LED lamp is \$0.00495, which means \(CE_{5_{LED Light Array}} = 0.03465 \approx 0.03\) for all seven lights in conjunction.

Since this proposal hasn't been implemented, there was not an appropriate comparison for installation costs of the light. In lieu of this, we used the value for \(CE_{1_{LED Light Array}}\) so \(CE_{2_{LED Light Array}} = 0.8475\) and \(CE_{3_{LED Light Array}} = 1.13\).

According to Inline Lighting Electrical, a LED bulb will last, on average, 25000 hours before it produces 70\% of its original light \cite{noauthor_life_nodate}. Considering that the average person living in the United States drives \(\approx293.33\) hours per year, it would take \(\approx85.23 years\) before the average LED light bulb on this eHMI needs to be replaced. Using Equation (\ref{yearly_cost_equation}), this means \(CE_{4_{LED Light Array}} \approx 0.01\).

\begin{table}[h]
\centering
\caption{Cost Effectiveness Questionnaire Results - LED Light Array}
\label{tab:result2_table4}
\begin{tabular}{|lr|}
\hline
\multicolumn{2}{|c|}{\textbf{LED LIGHT ARRAY RESULTS}} \\ \hline
\multicolumn{1}{|l|}{\textbf{CE1\_LEDLights}} & 1.13   \\ \hline
\multicolumn{1}{|l|}{\textbf{CE2\_LEDLights}} & 0.8475 \\ \hline
\multicolumn{1}{|l|}{\textbf{CE3\_LEDLights}} & 1.13   \\ \hline
\multicolumn{1}{|l|}{\textbf{CE4\_LEDLights}} & 0.01   \\ \hline
\multicolumn{1}{|l|}{\textbf{CE5\_LEDLights}} & 0.03   \\ \hline
\end{tabular}
\end{table}

\subsubsection{Vehicle Kinematics as Indication (No eHMI)}
The kinematic motion doesn't use any additional cost, so \(CE_{{1 to 5}_{No eHMI}} = 0\).

\subsubsection{Frontal Brake Lights (FBL)}
The Frontal Braking Light proposal uses solely the breaking light technology so its score is calculated using the equation 
    \begin{equation} \label{frontal_brake_light_cost_equation}
        CE_{n_{FBL}} = CE_{n_{Brake Light}}
    \end{equation}
    
\subsubsection{Knight Rider Display (KRD)}
The Knight Rider Display  uses the following equation to calculate its answers:
    \begin{equation} \label{knight_rider_cost_equation}
        CE_{n_{KRD}} = CE_{n_{LED Display}} + CE_{n_{LED Light Array}}
    \end{equation}

\subsubsection{Bumper Smile Display (BSD) and Bumper Text Display (BTD)}
Both the Bumper Smile Display and Bumper Text Display use the same technology so they use the equation
    \begin{equation} \label{BSD_and_BTD_cost_equation}
        CE_{n_{BSD}} = CE_{n_{BTD}} = CE_{n_{LED Display}}
    \end{equation}

\subsubsection{Results}
The results for this section can be seen in Table \ref{tab:result2_table}

\begin{table}[h]
\centering
\caption{Cost Effectiveness Questionnaire Results}
\label{tab:result2_table}
\begin{tabular}{|lrrrrr|}
\hline
\multicolumn{6}{|c|}{\textbf{COST EFFECTIVENESS RESULTS}} \\ \hline
\multicolumn{1}{|l|}{} &
  \multicolumn{1}{c|}{\textbf{No eHMI}} &
  \multicolumn{1}{c|}{\textbf{FBL}} &
  \multicolumn{1}{c|}{\textbf{KRD}} &
  \multicolumn{1}{c|}{\textbf{BSD}} &
  \multicolumn{1}{c|}{\textbf{BTD}} \\ \hline
\multicolumn{1}{|l|}{\textbf{CE1}} &
  \multicolumn{1}{r|}{0.00} &
  \multicolumn{1}{r|}{39.68} &
  \multicolumn{1}{r|}{4,608.03} &
  \multicolumn{1}{r|}{4,606.90} &
  4,606.90 \\ \hline
\multicolumn{1}{|l|}{\textbf{CE2}} &
  \multicolumn{1}{r|}{0.00} &
  \multicolumn{1}{r|}{34.13} &
  \multicolumn{1}{r|}{0.85} &
  \multicolumn{1}{r|}{0.00} &
  0.00 \\ \hline
\multicolumn{1}{|l|}{\textbf{CE3}} &
  \multicolumn{1}{r|}{0.00} &
  \multicolumn{1}{r|}{45.50} &
  \multicolumn{1}{r|}{1.13} &
  \multicolumn{1}{r|}{0.00} &
  0.00 \\ \hline
\multicolumn{1}{|l|}{\textbf{CE4}} &
  \multicolumn{1}{r|}{0.00} &
  \multicolumn{1}{r|}{53.94} &
  \multicolumn{1}{r|}{2,303.46} &
  \multicolumn{1}{r|}{2,303.45} &
  2,303.45 \\ \hline
\multicolumn{1}{|l|}{\textbf{CE5}} &
  \multicolumn{1}{r|}{0.00} &
  \multicolumn{1}{r|}{53.94} &
  \multicolumn{1}{r|}{27.62} &
  \multicolumn{1}{r|}{27.59} &
  27.59 \\ \hline
\multicolumn{1}{|l|}{} &
  \multicolumn{1}{c|}{\textbf{No eHMI}} &
  \multicolumn{1}{c|}{\textbf{FBL}} &
  \multicolumn{1}{c|}{\textbf{KRD}} &
  \multicolumn{1}{c|}{\textbf{BSD}} &
  \multicolumn{1}{c|}{\textbf{GTD}} \\ \hline
\multicolumn{1}{|l|}{\textbf{Baseline}} &
  \multicolumn{5}{c|}{48,301.00} \\ \hline
\multicolumn{1}{|l|}{\textbf{TOTAL}} &
  \multicolumn{1}{r|}{10.00} &
  \multicolumn{1}{r|}{9.95} &
  \multicolumn{1}{r|}{8.56} &
  \multicolumn{1}{r|}{8.56} &
  8.56 \\ \hline
\end{tabular}
\end{table}

\subsection{Accessibility Scores}
None of the proposals evaluated make use of sound components (\(A_{4} = 0, A_{{36 to 41}} = 0\)) or tactile components (\(A_{3} = 0, A_{{35}} = 0\)), nor do they employ imagery that could be distressing (\(A_{{64 to 68}} = 1\)) or psychologically offensive (\(A_{{69 to 73}} = 1\)). Based on the results of \cite{de_clercq_external_2019}, all of the presented proposals provide enough time to the observer to be understood (\(A_{{48}} = 1\)). The results of this evaluation can be seen in Table \ref{tab:result3}.

\subsubsection{Vehicle Kinematics as Indication (No eHMI)} 
The vehicle kinematics do not use pictograms (\(A_{1_{No eHMI}} = 0, A_{{9 to 17}_{No eHMI}} = 0\)), or text (\(A_{2_{No eHMI}} = 0, A_{{18 to 29}_{No eHMI}} = 0, A_{{47}_{No eHMI}} = 0, A_{{52 to 53}_{No eHMI}} = 1\)). While a vehicle may make a screeching sound as it stops, this is not mentioned in the literature as a relevant factor in how pedestrians interact with a yielding vehicle.

Using the sides of the vehicle as the 'display' means that while the placement of information is consistent (\(A_{7_{No eHMI}} = 1\)), it is also affected by the size of said vehicle (\(A_{5_{No eHMI}} = 0\)). No measures are taken to prevent the creation of shadows on this part of the vehicle (\(A_{8_{No eHMI}} = 0\)), and since the kinematics are observed through the chassis of the vehicle, some types of vehicles such a emergency ones may have additional visual clutter that difficult the reading of the motion (\(A_{6_{No eHMI}} = 0\)). The vehicle kinematics do not rely on color to communicate information (\(A_{30_{No eHMI}} = 1, A_{32_{No eHMI}} = 1, A_{33_{No eHMI}} = 1\)), but the color of the vehicle may be one commonly confused by color blind people (\(A_{31_{No eHMI}} = 0\)) such as red (\(A_{44_{No eHMI}} = 0\)). This proposal does not use flashing lights (\(A_{42_{No eHMI}} = 1, A_{43_{No eHMI}} = 1\)) or animations (\(A_{45_{No eHMI}} = 1, A_{46_{No eHMI}} = 1\)). Since the color of the eHMI is the same as its paint job, questions regarding that aspect are non-applicable (\(A_{34_{No eHMI}} = 0\)).

According to the results observed by Moore et al \cite{moore_case_2019}, pedestrians mainly rely on vehicle kinematics to determine when the vehicle is yielding to them. However, the motion is only and solely communicated implicitly (\(A_{{49 to 51}_{No eHMI}} = 0 \)). This method of communication is universal (\(A_{{54 to 58}_{No eHMI}} = 1\)), but it may hold a different meaning or be read differently depending on the situational context (\(A_{{59 to 63}_{No eHMI}} = 0\)).

\subsubsection{Front Braking Lights (FBL)} 
These do not use pictograms (\(A_{1_{FBL}} = 0, A_{{9 to 17}_{FBL}} = 0\)), or text (\(A_{2_{FBL}} = 0, A_{{18 to 29}_{FBL}} = 0, A_{{47}_{FBL}} = 0, A_{{52 to 53}_{FBL}} = 1\)). Since all vehicles have headlights, frontal brake lights can be consistently placed regardless of vehicle type or size (\(A_{5_{FBL}} = 1\)), the information is always displayed in the same eHMI (\(A_{7_{FBL}} = 1\)). The placement of frontal brake lights is close to the vehicle headlights, which creates visual clutter that may confuse an observer (\(A_{6_{FBL}} = 0\)). Headlights take measures to prevent the creation of shadows that may obscure them (\(A_{8_{FBL}} = 1\)).

Not the color, but the brightness state of the lights determine the intent of the vehicle (\(A_{30_{FBL}} = 1\)), and the sole color cyan in the proposal (\(A_{32_{FBL}} = 1, A_{44_{FBL}} = 1\)) does not fall within the most common color blind colors (\(A_{31_{FBL}} = 1\)). It may also be adjusted to fit local regulations (\(A_{33_{FBL}} = 0\)), although it does not have a way to ensure it doesn't blend with the rest of the vehicle (\(A_{34_{FBL}} = 0\)). The lights do not flash (\(A_{42_{FBL}} = 1, A_{43_{FBL}} = 1\)) nor do they employ any animations (\(A_{45_{FBL}} = 1, A_{46_{FBL}} = 1\)). 

The information communicated relates to the speed of the vehicle instead of its yielding decision, meaning it is presented implicitly (\(A_{50_{FBL}} = 0\)) and in only one way (\(A_{49_{FBL}} = 0\)) that is exocentric to the observer (\(A_{51_{FBL}} = 0\)). Considering the proliferation of tail lights in modern vehicles, the meaning of frontal brake lights, while new, can be considered universal (\(A_{{54 to 58}_{FBL}} = 1\)), although not independent from context since it may signal yielding or a simple slowing down of the vehicle (\(A_{{59 to 63}_{FBL}} = 0\)).

\subsubsection{Knight Rider Display (KRD)}
This display uses an animated, one dimensional pictogram (\(A_{1_{KRD}} = 1\)), but no text (\(A_{2_{KRD}} = 0, A_{{18 to 29}_{KRD}} = 0, A_{{47}_{KRD}} = 0, A_{{52 to 53}_{KRD}} = 1\)). Part of the placement is on the bumper which is not a cluttered space (\(A_{6_{KRD}} = 1\)), but it is one that can be affected or changed by vehicle type (\(A_{5_{KRD}} = 0\)). This proposal consistently places the display on the bumper (\(A_{7_{KRD}} = 1\)). The sample for this proposal is based on the sign display in \cite{ledsigncity_car_nodate}, which adjusts its brightness based on the environment to prevent shadows (\(A_{8_{KRD}} = 1\)) and glare (\(A_{14_{KRD}} = 1\)).

The pictogram is made up of identical upright straight thick bold cyan lights (\(A_{{10 to 12}_{KRD}} = 1\)) over a black background (\(A_{{15 to 17}_{KRD}} = 1\)), inside of a silver frame. This combination of light and dark should provide enough contrast from the vehicle paint job (\(A_{34_{KRD}} = 1\)) and surrounding elements (\(A_{9_{KRD}} = 1\)). The sweeping motion is also distinguishable from the static motion (\(A_{13_{KRD}} = 1\)). It does not rely on color to communicate information (\(A_{30_{KRD}} = 1\)) or use a color that is commonly problematic to color blind people (\(A_{31_{KRD}} = 1\)) such as red (\(A_{32_{KRD}} = 1, A_{44_{KRD}} = 1\)). The color can be changed during manufacturing (\(A_{33_{KRD}} = 1\)).

Each segment of the Knight Rider Display may flash (\(A_{42_{KRD}} = 0\)) but no more than three times per second (\(A_{43_{KRD}} = 1\)). While this proposal uses animations, these are all simple (\(A_{45_{KRD}} = 1\)) and not featuring multiple elements scrolling at different speeds (\(A_{46_{KRD}} = 1\)). 

While this proposal features two eHMIs, they both present information in the same exocentric way (\(A_{49_{KRD}} = 0, A_{51_{KRD}} = 0\)). The eHMIs only play their respective animation when the vehicle intends to yield, but while the communication is explicit, it is not directed at the viewer (\(A_{{50}_{KRD}} = 0\)). This method of communication is potentially not universal, and based on an American television show from 1982 \cite{larson_knight_1982} (\(A_{{54 to 58}_{KRD}} = 0\)). There isn't enough data to determine if the animation of this proposal may hold a different meaning based on the surroundings of the vehicle (\(A_{{59 to 63}_{KRD}} = 0\)).

\subsubsection{Bumper Smile Display (BSD)}
This proposal has a single eHMI: A LED display that uses the pictogram of a smile (\(A_{{1}_{BSD}} = 1\)) to communicate yielding intent. No text is employed (\(A_{2_{BSD}} = 0, A_{{18 to 29}_{BSD}} = 0,  A_{{47}_{BSD}} = 0, A_{{50}_{BSD}} = 1, A_{{51 to 53}_{BSD}} = 1\)). The eHMI is consistently placed on the bumper (\(A_{7_{BSD}} = 1\)), which is not a cluttered space (\(A_{6_{BSD}} = 1\)) but one that can change based on vehicle type (\(A_{5_{BSD}} = 0\)). The sample for this proposal is based on the sign display in \cite{ledsigncity_car_nodate}, which adjusts its brightness based on the environment to prevent shadows (\(A_{8_{BSD}} = 1\)) and glare (\(A_{14_{BSD}} = 1\)).

The pictogram of the smile uses a curved (\(A_{12_{BSD}} = 0\)) upright (\(A_{10_{BSD}} = 1\)), thick, uniform, and bold cyan line (\(A_{11_{BSD}} = 1\)) over a black background (\(A_{{15 to 17}_{BSD}} = 1\)), inside of a silver frame that distinguishes it from the rest of the vehicle (\(A_{{34}_{BSD}} = 1\)). This combination of light and dark should provide enough contrast with the vehicle paintjob and other surrounding elements (\(A_{9_{BSD}} = 1\)). The straight and the curved pictogram are distinguishable from each other (\(A_{13_{BSD}} = 1\)). It does not rely on color to communicate information (\(A_{30_{BSD}} = 1\)) or use a color that is commonly problematic to color blind people (\(A_{31_{BSD}} = 1\)) such as red (\(A_{44_{BSD}} = 1\)). The color used in this proposal is cyan (\(A_{{32}_{BSD}} = 1\)) and can be adjusted during manufacturing (\(A_{{33}_{BSD}} = 1\)).

While this proposal features a display, it does not flash (\(A_{{42 to 43}_{BSD}} = 1\)), as it just switches between two states instead of employing any animations (\(A_{{45 to 46}_{BSD}} = 1\)). Information is always presented in one exocentric way (\(A_{{49}_{BSD}} = 0, A_{{51}_{BSD}} = 0\)) but not in a way that is explicitly stated to the observer (\(A_{{50}_{BSD}} = 0\)).

Smiles are not a universal symbol of friendliness (\(A_{{54 to 58}_{BSD}} = 0\)) that may take on a different meaning depending on context \cite{krys_be_2015} (\(A_{{59 to 63}_{BSD}} = 0\)).

\subsubsection{Bumper Text Display (BTD)}
This proposal employs text (\(A_{2_{BTD}} = 1\)) instead of pictograms  (\(A_{1_{BTD}} = 0, A_{{9 to 14}_{BTD}} = 0\)). While some variants of the proposal place the text on the windshield, this specific example places it on the bumper (\(A_{7_{BTD}} = 1\)), which is a cluttered space (\(A_{6_{BTD}} = 0\)) that can change based on vehicle type (\(A_{5_{BTD}} = 0\)). This display adjusts its brightness based on the environment to prevent shadows (\(A_{8_{BTD}} = 1\)) and glare (\(A_{25_{BTD}} = 1\)) \cite{ledsigncity_car_nodate}.

The eHMI displays letters over a black background (\(A_{{15 to 18}_{BTD}} = 1\)), which contrasts with the adjustable cyan letters (\(A_{{20}_{BTD}} = 1, A_{{31 to 33}_{BTD}} = 1\)) and the rest of the vehicle (\(A_{{34}_{BTD}} = 1\)). The text is displayed upright (\(A_{{19}_{BTD}} = 1\)), using the bold Helvetica font which uses straight, uniform thick lines (\(A_{{20 to 22}_{BTD}} = 1\)) with wide horizontal proportions (\(A_{{27}_{BTD}} = 1\)) to display distinct characters (\(A_{{23}_{BTD}} = 1\)) with prominent ascenders, descenders, and open counterforms (\(A_{{24}_{BTD}} = 1\)).

The text displayed is ``Walk' and ``Don't walk', making use of initial uppercase letters (\(A_{{26}_{BTD}} = 1\)) and simple, idiom-free language (\(A_{{52 to 53}_{BTD}} = 1\)). It does not rotate (\(A_{{28}_{BTD}} = 1\)), include hyphens or em dashes (\(A_{{29}_{BTD}} = 1\)), or rely on color to communicate information (\(A_{{30}_{BTD}} = 1\)). The display only uses the color black and cyan (\(A_{{31}_{BTD}} = 1, A_{44_{BTD}} = 1\)), and it does not feature flashing lights (\(A_{{42 to 43}_{BTD}} = 1\)) or animations (\(A_{{45 to 46}_{BTD}} = 1\)). 

The ``Walk' message is explicitly egocentric to the pedestrian (\(A_{{50 to 51}_{BTD}} = 1\)) when the vehicle starts its yielding behavior. Based on the results in \cite{de_clercq_external_2019}, a pedestrian should be able to read and interpret the instructions in the time given by the vehicle (\(A_{{47}_{BTD}} = 1\)).

The instructions are not presented in another way (\(A_{{49}_{BTD}} = 0\)). Text relies on language to be understood, meaning that it is affected by the cultural and contextual sections of the questionnaire (\(A_{{54 to 58}_{BTD}} = 0, A_{{59 to 63}_{BTD}} = 0\)).
\pagebreak
\begin{longtable}[c]{|lccccc|}
\caption{Accessibility Questionnaire Results}
\label{tab:result3}\\
\hline
\multicolumn{6}{|c|}{\textbf{ACCESSIBILITY RESULTS}} \\ \hline
\endfirsthead
\multicolumn{6}{c}%
{{\bfseries Table \thetable\ continued from previous page}} \\
\hline
\multicolumn{6}{|c|}{\textbf{ACCESSIBILITY RESULTS}} \\ \hline
\endhead
\multicolumn{1}{|l|}{} &
  \multicolumn{1}{c|}{\textbf{No eHMI}} &
  \multicolumn{1}{c|}{\textbf{FBL}} &
  \multicolumn{1}{c|}{\textbf{KRD}} &
  \multicolumn{1}{c|}{\textbf{BSD}} &
  \textbf{BTD} \\ \hline
\multicolumn{1}{|l|}{\textbf{A1}} &
  \multicolumn{1}{c|}{0} &
  \multicolumn{1}{c|}{0} &
  \multicolumn{1}{c|}{1} &
  \multicolumn{1}{c|}{1} &
  0 \\ \hline
\multicolumn{1}{|l|}{\textbf{A2}} &
  \multicolumn{1}{c|}{0} &
  \multicolumn{1}{c|}{0} &
  \multicolumn{1}{c|}{0} &
  \multicolumn{1}{c|}{0} &
  1 \\ \hline
\multicolumn{1}{|l|}{\textbf{A3}} &
  \multicolumn{1}{c|}{0} &
  \multicolumn{1}{c|}{0} &
  \multicolumn{1}{c|}{0} &
  \multicolumn{1}{c|}{0} &
  0 \\ \hline
\multicolumn{1}{|l|}{\textbf{A4}} &
  \multicolumn{1}{c|}{0} &
  \multicolumn{1}{c|}{0} &
  \multicolumn{1}{c|}{0} &
  \multicolumn{1}{c|}{0} &
  0 \\ \hline
\multicolumn{1}{|l|}{\textbf{A5}} &
  \multicolumn{1}{c|}{0} &
  \multicolumn{1}{c|}{1} &
  \multicolumn{1}{c|}{0} &
  \multicolumn{1}{c|}{0} &
  0 \\ \hline
\multicolumn{1}{|l|}{\textbf{A6}} &
  \multicolumn{1}{c|}{0} &
  \multicolumn{1}{c|}{1} &
  \multicolumn{1}{c|}{1} &
  \multicolumn{1}{c|}{1} &
  1 \\ \hline
\multicolumn{1}{|l|}{\textbf{A7}} &
  \multicolumn{1}{c|}{1} &
  \multicolumn{1}{c|}{1} &
  \multicolumn{1}{c|}{1} &
  \multicolumn{1}{c|}{1} &
  1 \\ \hline
\multicolumn{1}{|l|}{\textbf{A8}} &
  \multicolumn{1}{c|}{0} &
  \multicolumn{1}{c|}{1} &
  \multicolumn{1}{c|}{1} &
  \multicolumn{1}{c|}{1} &
  1 \\ \hline
\multicolumn{1}{|l|}{\textbf{A9}} &
  \multicolumn{1}{c|}{0} &
  \multicolumn{1}{c|}{0} &
  \multicolumn{1}{c|}{1} &
  \multicolumn{1}{c|}{1} &
  1 \\ \hline
\multicolumn{1}{|l|}{\textbf{A10}} &
  \multicolumn{1}{c|}{0} &
  \multicolumn{1}{c|}{0} &
  \multicolumn{1}{c|}{1} &
  \multicolumn{1}{c|}{1} &
  0 \\ \hline
\multicolumn{1}{|l|}{\textbf{A11}} &
  \multicolumn{1}{c|}{0} &
  \multicolumn{1}{c|}{0} &
  \multicolumn{1}{c|}{1} &
  \multicolumn{1}{c|}{1} &
  0 \\ \hline
\multicolumn{1}{|l|}{\textbf{A12}} &
  \multicolumn{1}{c|}{0} &
  \multicolumn{1}{c|}{0} &
  \multicolumn{1}{c|}{1} &
  \multicolumn{1}{c|}{0} &
  0 \\ \hline
\multicolumn{1}{|l|}{\textbf{A13}} &
  \multicolumn{1}{c|}{0} &
  \multicolumn{1}{c|}{0} &
  \multicolumn{1}{c|}{1} &
  \multicolumn{1}{c|}{1} &
  0 \\ \hline
\multicolumn{1}{|l|}{\textbf{A14}} &
  \multicolumn{1}{c|}{0} &
  \multicolumn{1}{c|}{0} &
  \multicolumn{1}{c|}{1} &
  \multicolumn{1}{c|}{1} &
  0 \\ \hline
\multicolumn{1}{|l|}{\textbf{A15}} &
  \multicolumn{1}{c|}{0} &
  \multicolumn{1}{c|}{0} &
  \multicolumn{1}{c|}{1} &
  \multicolumn{1}{c|}{1} &
  1 \\ \hline
\multicolumn{1}{|l|}{\textbf{A16}} &
  \multicolumn{1}{c|}{0} &
  \multicolumn{1}{c|}{0} &
  \multicolumn{1}{c|}{1} &
  \multicolumn{1}{c|}{1} &
  1 \\ \hline
\multicolumn{1}{|l|}{\textbf{A17}} &
  \multicolumn{1}{c|}{0} &
  \multicolumn{1}{c|}{0} &
  \multicolumn{1}{c|}{1} &
  \multicolumn{1}{c|}{1} &
  1 \\ \hline
\multicolumn{1}{|l|}{\textbf{A18}} &
  \multicolumn{1}{c|}{0} &
  \multicolumn{1}{c|}{0} &
  \multicolumn{1}{c|}{0} &
  \multicolumn{1}{c|}{0} &
  1 \\ \hline
\multicolumn{1}{|l|}{\textbf{A19}} &
  \multicolumn{1}{c|}{0} &
  \multicolumn{1}{c|}{0} &
  \multicolumn{1}{c|}{0} &
  \multicolumn{1}{c|}{0} &
  1 \\ \hline
\multicolumn{1}{|l|}{\textbf{A20}} &
  \multicolumn{1}{c|}{0} &
  \multicolumn{1}{c|}{0} &
  \multicolumn{1}{c|}{0} &
  \multicolumn{1}{c|}{0} &
  1 \\ \hline
\multicolumn{1}{|l|}{\textbf{A21}} &
  \multicolumn{1}{c|}{0} &
  \multicolumn{1}{c|}{0} &
  \multicolumn{1}{c|}{0} &
  \multicolumn{1}{c|}{0} &
  1 \\ \hline
\multicolumn{1}{|l|}{\textbf{A22}} &
  \multicolumn{1}{c|}{0} &
  \multicolumn{1}{c|}{0} &
  \multicolumn{1}{c|}{0} &
  \multicolumn{1}{c|}{0} &
  1 \\ \hline
\multicolumn{1}{|l|}{\textbf{A23}} &
  \multicolumn{1}{c|}{0} &
  \multicolumn{1}{c|}{0} &
  \multicolumn{1}{c|}{0} &
  \multicolumn{1}{c|}{0} &
  1 \\ \hline
\multicolumn{1}{|l|}{\textbf{A24}} &
  \multicolumn{1}{c|}{0} &
  \multicolumn{1}{c|}{0} &
  \multicolumn{1}{c|}{0} &
  \multicolumn{1}{c|}{0} &
  1 \\ \hline
\multicolumn{1}{|l|}{\textbf{A25}} &
  \multicolumn{1}{c|}{0} &
  \multicolumn{1}{c|}{0} &
  \multicolumn{1}{c|}{0} &
  \multicolumn{1}{c|}{0} &
  1 \\ \hline
\multicolumn{1}{|l|}{\textbf{A26}} &
  \multicolumn{1}{c|}{0} &
  \multicolumn{1}{c|}{0} &
  \multicolumn{1}{c|}{0} &
  \multicolumn{1}{c|}{0} &
  1 \\ \hline
\multicolumn{1}{|l|}{\textbf{A27}} &
  \multicolumn{1}{c|}{0} &
  \multicolumn{1}{c|}{0} &
  \multicolumn{1}{c|}{0} &
  \multicolumn{1}{c|}{0} &
  1 \\ \hline
\multicolumn{1}{|l|}{\textbf{A28}} &
  \multicolumn{1}{c|}{0} &
  \multicolumn{1}{c|}{0} &
  \multicolumn{1}{c|}{0} &
  \multicolumn{1}{c|}{0} &
  1 \\ \hline
\multicolumn{1}{|l|}{\textbf{A29}} &
  \multicolumn{1}{c|}{0} &
  \multicolumn{1}{c|}{0} &
  \multicolumn{1}{c|}{0} &
  \multicolumn{1}{c|}{0} &
  1 \\ \hline
\multicolumn{1}{|l|}{\textbf{A30}} &
  \multicolumn{1}{c|}{1} &
  \multicolumn{1}{c|}{1} &
  \multicolumn{1}{c|}{1} &
  \multicolumn{1}{c|}{1} &
  1 \\ \hline
\multicolumn{1}{|l|}{\textbf{A31}} &
  \multicolumn{1}{c|}{0} &
  \multicolumn{1}{c|}{1} &
  \multicolumn{1}{c|}{1} &
  \multicolumn{1}{c|}{1} &
  1 \\ \hline
\multicolumn{1}{|l|}{\textbf{A32}} &
  \multicolumn{1}{c|}{1} &
  \multicolumn{1}{c|}{1} &
  \multicolumn{1}{c|}{1} &
  \multicolumn{1}{c|}{1} &
  1 \\ \hline
\multicolumn{1}{|l|}{\textbf{A33}} &
  \multicolumn{1}{c|}{1} &
  \multicolumn{1}{c|}{1} &
  \multicolumn{1}{c|}{1} &
  \multicolumn{1}{c|}{1} &
  1 \\ \hline
\multicolumn{1}{|l|}{\textbf{A34}} &
  \multicolumn{1}{c|}{0} &
  \multicolumn{1}{c|}{0} &
  \multicolumn{1}{c|}{1} &
  \multicolumn{1}{c|}{1} &
  1 \\ \hline
\multicolumn{1}{|l|}{\textbf{A35}} &
  \multicolumn{1}{c|}{0} &
  \multicolumn{1}{c|}{0} &
  \multicolumn{1}{c|}{0} &
  \multicolumn{1}{c|}{0} &
  0 \\ \hline
\multicolumn{1}{|l|}{\textbf{A36}} &
  \multicolumn{1}{c|}{0} &
  \multicolumn{1}{c|}{0} &
  \multicolumn{1}{c|}{0} &
  \multicolumn{1}{c|}{0} &
  0 \\ \hline
\multicolumn{1}{|l|}{\textbf{A37}} &
  \multicolumn{1}{c|}{0} &
  \multicolumn{1}{c|}{0} &
  \multicolumn{1}{c|}{0} &
  \multicolumn{1}{c|}{0} &
  0 \\ \hline
\multicolumn{1}{|l|}{\textbf{A38}} &
  \multicolumn{1}{c|}{0} &
  \multicolumn{1}{c|}{0} &
  \multicolumn{1}{c|}{0} &
  \multicolumn{1}{c|}{0} &
  0 \\ \hline
\multicolumn{1}{|l|}{\textbf{A39}} &
  \multicolumn{1}{c|}{0} &
  \multicolumn{1}{c|}{0} &
  \multicolumn{1}{c|}{0} &
  \multicolumn{1}{c|}{0} &
  0 \\ \hline
\multicolumn{1}{|l|}{\textbf{A40}} &
  \multicolumn{1}{c|}{0} &
  \multicolumn{1}{c|}{0} &
  \multicolumn{1}{c|}{0} &
  \multicolumn{1}{c|}{0} &
  0 \\ \hline
\multicolumn{1}{|l|}{\textbf{A41}} &
  \multicolumn{1}{c|}{0} &
  \multicolumn{1}{c|}{0} &
  \multicolumn{1}{c|}{0} &
  \multicolumn{1}{c|}{0} &
  0 \\ \hline
\multicolumn{1}{|l|}{\textbf{A42}} &
  \multicolumn{1}{c|}{1} &
  \multicolumn{1}{c|}{1} &
  \multicolumn{1}{c|}{0} &
  \multicolumn{1}{c|}{1} &
  1 \\ \hline
\multicolumn{1}{|l|}{\textbf{A43}} &
  \multicolumn{1}{c|}{1} &
  \multicolumn{1}{c|}{1} &
  \multicolumn{1}{c|}{1} &
  \multicolumn{1}{c|}{1} &
  1 \\ \hline
\multicolumn{1}{|l|}{\textbf{A44}} &
  \multicolumn{1}{c|}{0} &
  \multicolumn{1}{c|}{1} &
  \multicolumn{1}{c|}{1} &
  \multicolumn{1}{c|}{1} &
  1 \\ \hline
\multicolumn{1}{|l|}{\textbf{A45}} &
  \multicolumn{1}{c|}{1} &
  \multicolumn{1}{c|}{1} &
  \multicolumn{1}{c|}{1} &
  \multicolumn{1}{c|}{1} &
  1 \\ \hline
\multicolumn{1}{|l|}{\textbf{A46}} &
  \multicolumn{1}{c|}{1} &
  \multicolumn{1}{c|}{1} &
  \multicolumn{1}{c|}{1} &
  \multicolumn{1}{c|}{1} &
  1 \\ \hline
\multicolumn{1}{|l|}{\textbf{A47}} &
  \multicolumn{1}{c|}{0} &
  \multicolumn{1}{c|}{0} &
  \multicolumn{1}{c|}{0} &
  \multicolumn{1}{c|}{0} &
  1 \\ \hline
\multicolumn{1}{|l|}{\textbf{A48}} &
  \multicolumn{1}{c|}{1} &
  \multicolumn{1}{c|}{1} &
  \multicolumn{1}{c|}{1} &
  \multicolumn{1}{c|}{1} &
  1 \\ \hline
\multicolumn{1}{|l|}{\textbf{A49}} &
  \multicolumn{1}{c|}{0} &
  \multicolumn{1}{c|}{0} &
  \multicolumn{1}{c|}{0} &
  \multicolumn{1}{c|}{0} &
  0 \\ \hline
\multicolumn{1}{|l|}{\textbf{A50}} &
  \multicolumn{1}{c|}{0} &
  \multicolumn{1}{c|}{0} &
  \multicolumn{1}{c|}{0} &
  \multicolumn{1}{c|}{0} &
  1 \\ \hline
\multicolumn{1}{|l|}{\textbf{A51}} &
  \multicolumn{1}{c|}{0} &
  \multicolumn{1}{c|}{0} &
  \multicolumn{1}{c|}{0} &
  \multicolumn{1}{c|}{0} &
  1 \\ \hline
\multicolumn{1}{|l|}{\textbf{A52}} &
  \multicolumn{1}{c|}{1} &
  \multicolumn{1}{c|}{1} &
  \multicolumn{1}{c|}{1} &
  \multicolumn{1}{c|}{1} &
  1 \\ \hline
\multicolumn{1}{|l|}{\textbf{A53}} &
  \multicolumn{1}{c|}{1} &
  \multicolumn{1}{c|}{1} &
  \multicolumn{1}{c|}{1} &
  \multicolumn{1}{c|}{1} &
  1 \\ \hline
\multicolumn{1}{|l|}{\textbf{A54}} &
  \multicolumn{1}{c|}{1} &
  \multicolumn{1}{c|}{1} &
  \multicolumn{1}{c|}{0} &
  \multicolumn{1}{c|}{0} &
  0 \\ \hline
\multicolumn{1}{|l|}{\textbf{A55}} &
  \multicolumn{1}{c|}{1} &
  \multicolumn{1}{c|}{1} &
  \multicolumn{1}{c|}{0} &
  \multicolumn{1}{c|}{0} &
  0 \\ \hline
\multicolumn{1}{|l|}{\textbf{A56}} &
  \multicolumn{1}{c|}{1} &
  \multicolumn{1}{c|}{1} &
  \multicolumn{1}{c|}{0} &
  \multicolumn{1}{c|}{0} &
  0 \\ \hline
\multicolumn{1}{|l|}{\textbf{A57}} &
  \multicolumn{1}{c|}{1} &
  \multicolumn{1}{c|}{1} &
  \multicolumn{1}{c|}{0} &
  \multicolumn{1}{c|}{0} &
  0 \\ \hline
\multicolumn{1}{|l|}{\textbf{A58}} &
  \multicolumn{1}{c|}{1} &
  \multicolumn{1}{c|}{1} &
  \multicolumn{1}{c|}{0} &
  \multicolumn{1}{c|}{0} &
  0 \\ \hline
\multicolumn{1}{|l|}{\textbf{A59}} &
  \multicolumn{1}{c|}{0} &
  \multicolumn{1}{c|}{0} &
  \multicolumn{1}{c|}{0} &
  \multicolumn{1}{c|}{0} &
  0 \\ \hline
\multicolumn{1}{|l|}{\textbf{A60}} &
  \multicolumn{1}{c|}{0} &
  \multicolumn{1}{c|}{0} &
  \multicolumn{1}{c|}{0} &
  \multicolumn{1}{c|}{0} &
  0 \\ \hline
\multicolumn{1}{|l|}{\textbf{A61}} &
  \multicolumn{1}{c|}{0} &
  \multicolumn{1}{c|}{0} &
  \multicolumn{1}{c|}{0} &
  \multicolumn{1}{c|}{0} &
  0 \\ \hline
\multicolumn{1}{|l|}{\textbf{A62}} &
  \multicolumn{1}{c|}{0} &
  \multicolumn{1}{c|}{0} &
  \multicolumn{1}{c|}{0} &
  \multicolumn{1}{c|}{0} &
  0 \\ \hline
\multicolumn{1}{|l|}{\textbf{A63}} &
  \multicolumn{1}{c|}{0} &
  \multicolumn{1}{c|}{0} &
  \multicolumn{1}{c|}{0} &
  \multicolumn{1}{c|}{0} &
  0 \\ \hline
\multicolumn{1}{|l|}{\textbf{A64}} &
  \multicolumn{1}{c|}{1} &
  \multicolumn{1}{c|}{1} &
  \multicolumn{1}{c|}{1} &
  \multicolumn{1}{c|}{1} &
  1 \\ \hline
\multicolumn{1}{|l|}{\textbf{A65}} &
  \multicolumn{1}{c|}{1} &
  \multicolumn{1}{c|}{1} &
  \multicolumn{1}{c|}{1} &
  \multicolumn{1}{c|}{1} &
  1 \\ \hline
\multicolumn{1}{|l|}{\textbf{A66}} &
  \multicolumn{1}{c|}{1} &
  \multicolumn{1}{c|}{1} &
  \multicolumn{1}{c|}{1} &
  \multicolumn{1}{c|}{1} &
  1 \\ \hline
\multicolumn{1}{|l|}{\textbf{A67}} &
  \multicolumn{1}{c|}{1} &
  \multicolumn{1}{c|}{1} &
  \multicolumn{1}{c|}{1} &
  \multicolumn{1}{c|}{1} &
  1 \\ \hline
\multicolumn{1}{|l|}{\textbf{A68}} &
  \multicolumn{1}{c|}{1} &
  \multicolumn{1}{c|}{1} &
  \multicolumn{1}{c|}{1} &
  \multicolumn{1}{c|}{1} &
  1 \\ \hline
\multicolumn{1}{|l|}{\textbf{A69}} &
  \multicolumn{1}{c|}{1} &
  \multicolumn{1}{c|}{1} &
  \multicolumn{1}{c|}{1} &
  \multicolumn{1}{c|}{1} &
  1 \\ \hline
\multicolumn{1}{|l|}{\textbf{A70}} &
  \multicolumn{1}{c|}{1} &
  \multicolumn{1}{c|}{1} &
  \multicolumn{1}{c|}{1} &
  \multicolumn{1}{c|}{1} &
  1 \\ \hline
\multicolumn{1}{|l|}{\textbf{A71}} &
  \multicolumn{1}{c|}{1} &
  \multicolumn{1}{c|}{1} &
  \multicolumn{1}{c|}{1} &
  \multicolumn{1}{c|}{1} &
  1 \\ \hline
\multicolumn{1}{|l|}{\textbf{A72}} &
  \multicolumn{1}{c|}{1} &
  \multicolumn{1}{c|}{1} &
  \multicolumn{1}{c|}{1} &
  \multicolumn{1}{c|}{1} &
  1 \\ \hline
\multicolumn{1}{|l|}{\textbf{A73}} &
  \multicolumn{1}{c|}{1} &
  \multicolumn{1}{c|}{1} &
  \multicolumn{1}{c|}{1} &
  \multicolumn{1}{c|}{1} &
  1 \\ \hline
\multicolumn{1}{|l|}{} &
  \multicolumn{1}{c|}{\textbf{No eHMI}} &
  \multicolumn{1}{c|}{\textbf{FBL}} &
  \multicolumn{1}{c|}{\textbf{KRD}} &
  \multicolumn{1}{c|}{\textbf{BSD}} &
  \textbf{GTD} \\ \hline
\multicolumn{1}{|l|}{\textbf{TOTAL}} &
  \multicolumn{1}{c|}{3.56} &
  \multicolumn{1}{c|}{4.25} &
  \multicolumn{1}{c|}{4.79} &
  \multicolumn{1}{c|}{4.79} &
  6.30 \\ \hline
\end{longtable}

\subsection{Ease of Understanding Scores}
This evaluation uses the \textit{feel-safe percentages} \((FSP)\) given in \cite[Fig.~9]{de_clercq_external_2019}, rounded to the nearest whole number. Since the participants in that experiment had a mean age of 24.57 years, these values are used for the question corresponding to adult pedestrians.

Since the study does not produce other demographic information, no other questions were answered in this scenario.

\begin{longtable}[c]{|lccccc|}
\caption{Ease of Understanding Questionnaire Results}
\label{tab:result4_table}\\
\hline
\multicolumn{6}{|c|}{\textbf{EASE OF UNDERSTANDING RESULTS}} \\ \hline
\endfirsthead
\endhead
\multicolumn{1}{|l|}{} &
  \multicolumn{1}{c|}{\textbf{No eHMI}} &
  \multicolumn{1}{c|}{\textbf{FBL}} &
  \multicolumn{1}{c|}{\textbf{KRD}} &
  \multicolumn{1}{c|}{\textbf{BSD}} &
  \textbf{BTD} \\ \hline
\multicolumn{1}{|l|}{\textbf{EU1}} &
  \multicolumn{1}{c|}{0} &
  \multicolumn{1}{c|}{0} &
  \multicolumn{1}{c|}{0} &
  \multicolumn{1}{c|}{0} &
  0 \\ \hline
\multicolumn{1}{|l|}{\textbf{EU2}} &
  \multicolumn{1}{c|}{65} &
  \multicolumn{1}{c|}{74} &
  \multicolumn{1}{c|}{79} &
  \multicolumn{1}{c|}{74} &
  76 \\ \hline
\multicolumn{1}{|l|}{\textbf{EU3}} &
  \multicolumn{1}{c|}{0} &
  \multicolumn{1}{c|}{0} &
  \multicolumn{1}{c|}{0} &
  \multicolumn{1}{c|}{0} &
  0 \\ \hline
\multicolumn{1}{|l|}{\textbf{EU4}} &
  \multicolumn{1}{c|}{0} &
  \multicolumn{1}{c|}{0} &
  \multicolumn{1}{c|}{0} &
  \multicolumn{1}{c|}{0} &
  0 \\ \hline
\multicolumn{1}{|l|}{\textbf{EU5}} &
  \multicolumn{1}{c|}{0} &
  \multicolumn{1}{c|}{0} &
  \multicolumn{1}{c|}{0} &
  \multicolumn{1}{c|}{0} &
  0 \\ \hline
\multicolumn{1}{|l|}{\textbf{EU6}} &
  \multicolumn{1}{c|}{0} &
  \multicolumn{1}{c|}{0} &
  \multicolumn{1}{c|}{0} &
  \multicolumn{1}{c|}{0} &
  0 \\ \hline
\multicolumn{1}{|l|}{\textbf{EU7}} &
  \multicolumn{1}{c|}{0} &
  \multicolumn{1}{c|}{0} &
  \multicolumn{1}{c|}{0} &
  \multicolumn{1}{c|}{0} &
  0 \\ \hline
\multicolumn{1}{|l|}{\textbf{EU8}} &
  \multicolumn{1}{c|}{0} &
  \multicolumn{1}{c|}{0} &
  \multicolumn{1}{c|}{0} &
  \multicolumn{1}{c|}{0} &
  0 \\ \hline
\multicolumn{1}{|l|}{} &
  \multicolumn{1}{c|}{\textbf{No eHMI}} &
  \multicolumn{1}{c|}{\textbf{FBL}} &
  \multicolumn{1}{c|}{\textbf{KRD}} &
  \multicolumn{1}{c|}{\textbf{BSD}} &
  \textbf{GTD} \\ \hline
\multicolumn{1}{|l|}{\textbf{TOTAL}} &
  \multicolumn{1}{c|}{0.81} &
  \multicolumn{1}{c|}{0.93} &
  \multicolumn{1}{c|}{0.99} &
  \multicolumn{1}{c|}{0.93} &
  0.95 \\ \hline
\end{longtable}

\subsection{Constant Communication Scores}
Due to the reduced scope of this work, most questions in this sections can be answered with a ``No`` (0) or ``Not Applicable'' (0), so we will only mention the exceptions.

\subsubsection{Vehicle Kinematics as Indication (No eHMI)} 
By themselves, the vehicle kinematics can communicate motion (\(CC_{1_{No eHMI}} = 1\)) and acceleration (\(CC_{3_{No eHMI}} = 1, CC4_{4_{No eHMI}} = 1\)). Vehicle kinematics do not explicitly communicate to observers the many intricacies and possible errors of the navigation system. It is also limited to situations where a vehicle may or may not yield for a pedestrian. 

\subsubsection{Front Braking Lights (FBL)} 
The Frontal Braking Lights can communicate when the vehicle is slowing down (\(CC_{3_{FBL}} = 1\)), however it doesn't have a state exclusive to when the vehicle is in motion or for when the vehicle is accelerating.

\subsubsection{Knight Rider Display (KRD)}
Since this display does not turn on when a person is driving the vehicle, it communicates when the vehicle is in autonomous mode (\(CC_{5_{KRD}} = 1\)), but it does not explicitly communicate otherwise. This display also remains solid when the vehicle is yielding to a pedestrian (\(CC_{15_{KRD}} = 1\)) and plays a swiping animation when it's not yielding to them (\(CC_{15_{KRD}} = 1\)). While the display communicates when the vehicle is slowing down as part of its yielding behavior, it does not explicitly communicate when it's slowing down for another reason.  

\subsubsection{Bumper Smile Display (BSD)}
Similarly to the Knight Rider Display, the Bumper Smile Display can communicate when the vehicle is in autonomous mode (\(CC_{5_{BSD}} = 1\)), when it is yielding to a pedestrian (\(CC_{14_{BSD}} = 1\)), and when it is not (\(CC_{15_{BSD}} = 1\)), but it does not communicate additional information.

\subsubsection{Bumper Text Display (BTD)}
The technology in the proposal can be used to display a wide variety of information, but within the scope of this evaluation and the description given, it only communicates the vehicle is in autonomous mode (\(CC_{5_{BTD}} = 1\)), when it is yielding to a pedestrian (\(CC_{14_{BTD}} = 1\)), and when it is not yielding to a pedestrian (\(CC_{15_{BTD}} = 1\)). 

\begin{table}[h]
\centering
\caption{Constant Communication Questionnaire Results}
\label{tab:result5_table}
\begin{tabular}{|lccccc|}
\hline
\multicolumn{6}{|c|}{\textbf{CONSTANT COMMUNICATION RESULTS}} \\ \hline
\multicolumn{1}{|l|}{} &
  \multicolumn{1}{c|}{\textbf{No eHMI}} &
  \multicolumn{1}{c|}{\textbf{FBL}} &
  \multicolumn{1}{c|}{\textbf{KRD}} &
  \multicolumn{1}{c|}{\textbf{BSD}} &
  \textbf{BTD} \\ \hline
\multicolumn{1}{|l|}{\textbf{CC1}} &
  \multicolumn{1}{c|}{1} &
  \multicolumn{1}{c|}{0} &
  \multicolumn{1}{c|}{0} &
  \multicolumn{1}{c|}{0} &
  1 \\ \hline
\multicolumn{1}{|l|}{\textbf{CC2}} &
  \multicolumn{1}{c|}{0} &
  \multicolumn{1}{c|}{0} &
  \multicolumn{1}{c|}{0} &
  \multicolumn{1}{c|}{0} &
  0 \\ \hline
\multicolumn{1}{|l|}{\textbf{CC3}} &
  \multicolumn{1}{c|}{1} &
  \multicolumn{1}{c|}{1} &
  \multicolumn{1}{c|}{0} &
  \multicolumn{1}{c|}{0} &
  0 \\ \hline
\multicolumn{1}{|l|}{\textbf{CC4}} &
  \multicolumn{1}{c|}{0} &
  \multicolumn{1}{c|}{0} &
  \multicolumn{1}{c|}{0} &
  \multicolumn{1}{c|}{0} &
  0 \\ \hline
\multicolumn{1}{|l|}{\textbf{CC5}} &
  \multicolumn{1}{c|}{0} &
  \multicolumn{1}{c|}{0} &
  \multicolumn{1}{c|}{1} &
  \multicolumn{1}{c|}{1} &
  1 \\ \hline
\multicolumn{1}{|l|}{\textbf{CC6}} &
  \multicolumn{1}{c|}{0} &
  \multicolumn{1}{c|}{0} &
  \multicolumn{1}{c|}{0} &
  \multicolumn{1}{c|}{0} &
  0 \\ \hline
\multicolumn{1}{|l|}{\textbf{CC7}} &
  \multicolumn{1}{c|}{0} &
  \multicolumn{1}{c|}{0} &
  \multicolumn{1}{c|}{0} &
  \multicolumn{1}{c|}{0} &
  0 \\ \hline
\multicolumn{1}{|l|}{\textbf{CC8}} &
  \multicolumn{1}{c|}{0} &
  \multicolumn{1}{c|}{0} &
  \multicolumn{1}{c|}{0} &
  \multicolumn{1}{c|}{0} &
  0 \\ \hline
\multicolumn{1}{|l|}{\textbf{CC9}} &
  \multicolumn{1}{c|}{0} &
  \multicolumn{1}{c|}{0} &
  \multicolumn{1}{c|}{0} &
  \multicolumn{1}{c|}{0} &
  0 \\ \hline
\multicolumn{1}{|l|}{\textbf{CC10}} &
  \multicolumn{1}{c|}{0} &
  \multicolumn{1}{c|}{0} &
  \multicolumn{1}{c|}{0} &
  \multicolumn{1}{c|}{0} &
  0 \\ \hline
\multicolumn{1}{|l|}{\textbf{CC11}} &
  \multicolumn{1}{c|}{0} &
  \multicolumn{1}{c|}{0} &
  \multicolumn{1}{c|}{0} &
  \multicolumn{1}{c|}{0} &
  0 \\ \hline
\multicolumn{1}{|l|}{\textbf{CC12}} &
  \multicolumn{1}{c|}{0} &
  \multicolumn{1}{c|}{0} &
  \multicolumn{1}{c|}{0} &
  \multicolumn{1}{c|}{0} &
  0 \\ \hline
\multicolumn{1}{|l|}{\textbf{CC13}} &
  \multicolumn{1}{c|}{0} &
  \multicolumn{1}{c|}{0} &
  \multicolumn{1}{c|}{0} &
  \multicolumn{1}{c|}{0} &
  0 \\ \hline
\multicolumn{1}{|l|}{\textbf{CC14}} &
  \multicolumn{1}{c|}{0} &
  \multicolumn{1}{c|}{0} &
  \multicolumn{1}{c|}{1} &
  \multicolumn{1}{c|}{1} &
  1 \\ \hline
\multicolumn{1}{|l|}{\textbf{CC15}} &
  \multicolumn{1}{c|}{0} &
  \multicolumn{1}{c|}{0} &
  \multicolumn{1}{c|}{1} &
  \multicolumn{1}{c|}{1} &
  1 \\ \hline
\multicolumn{1}{|l|}{\textbf{CC16}} &
  \multicolumn{1}{c|}{0} &
  \multicolumn{1}{c|}{0} &
  \multicolumn{1}{c|}{0} &
  \multicolumn{1}{c|}{0} &
  0 \\ \hline
\multicolumn{1}{|l|}{\textbf{CC17}} &
  \multicolumn{1}{c|}{0} &
  \multicolumn{1}{c|}{0} &
  \multicolumn{1}{c|}{0} &
  \multicolumn{1}{c|}{0} &
  0 \\ \hline
\multicolumn{1}{|l|}{\textbf{CC18}} &
  \multicolumn{1}{c|}{0} &
  \multicolumn{1}{c|}{0} &
  \multicolumn{1}{c|}{0} &
  \multicolumn{1}{c|}{0} &
  0 \\ \hline
\multicolumn{1}{|l|}{\textbf{CC19}} &
  \multicolumn{1}{c|}{0} &
  \multicolumn{1}{c|}{0} &
  \multicolumn{1}{c|}{0} &
  \multicolumn{1}{c|}{0} &
  0 \\ \hline
\multicolumn{1}{|l|}{\textbf{CC20}} &
  \multicolumn{1}{c|}{0} &
  \multicolumn{1}{c|}{0} &
  \multicolumn{1}{c|}{0} &
  \multicolumn{1}{c|}{0} &
  0 \\ \hline
\multicolumn{1}{|l|}{\textbf{CC21}} &
  \multicolumn{1}{c|}{0} &
  \multicolumn{1}{c|}{0} &
  \multicolumn{1}{c|}{0} &
  \multicolumn{1}{c|}{0} &
  0 \\ \hline
\multicolumn{1}{|l|}{\textbf{CC22}} &
  \multicolumn{1}{c|}{0} &
  \multicolumn{1}{c|}{0} &
  \multicolumn{1}{c|}{0} &
  \multicolumn{1}{c|}{0} &
  0 \\ \hline
\multicolumn{1}{|l|}{\textbf{CC22}} &
  \multicolumn{1}{c|}{0} &
  \multicolumn{1}{c|}{0} &
  \multicolumn{1}{c|}{0} &
  \multicolumn{1}{c|}{0} &
  0 \\ \hline
\multicolumn{1}{|l|}{\textbf{CC24}} &
  \multicolumn{1}{c|}{0} &
  \multicolumn{1}{c|}{0} &
  \multicolumn{1}{c|}{0} &
  \multicolumn{1}{c|}{0} &
  0 \\ \hline
\multicolumn{1}{|l|}{} &
  \multicolumn{1}{c|}{\textbf{No eHMI}} &
  \multicolumn{1}{c|}{\textbf{FBL}} &
  \multicolumn{1}{c|}{\textbf{KRD}} &
  \multicolumn{1}{c|}{\textbf{BSD}} &
  \textbf{GTD} \\ \hline
\multicolumn{1}{|l|}{\textbf{TOTAL}} &
  \multicolumn{1}{c|}{0.83} &
  \multicolumn{1}{c|}{0.42} &
  \multicolumn{1}{c|}{1.25} &
  \multicolumn{1}{c|}{1.25} &
  1.67 \\ \hline
\end{tabular}
\end{table}

\subsection{Positioning Scores}
For this analysis, the purpose of all of these eHMI is solely to communicate to pedestrians and drivers looking at the vehicle's front of the it's intent to yield to a pedestrian (\(P_{35} = 0, P_{40} = 0\)). All of these proposals are symmetrical (\(P_{1} = 1\)), and none are placed on the roof of the vehicle (\(P_{2} = 0\)). Results can be seen on Table \ref{tab:result6_table}.

\subsubsection{Vehicle Kinematics as Indication (No eHMI)} 
Both sides of the vehicle serve as the display of this proposal, this includes both pairs of fenders (\(P_{5_{No eHMI}} = 2, P_{15_{No eHMI}} = 2\)), pairs of doors (\(P_{14_{No eHMI}} = 2, P_{23_{No eHMI}} = 2\)), pairs of wheels (\(P_{18_{No eHMI}} = 2, P_{24_{No eHMI}} = 2\)), windshield rails (\(P_{17_{No eHMI}} = 2\)), back window rails (\(P_{8_{No eHMI}} = 2\)), headlights (\(P_{21_{No eHMI}} = 2\)), rocker panels (\(P_{26_{No eHMI}} = 2\)), side mirrors (\(P_{28_{No eHMI}} = 2, P_{29_{No eHMI}} = 2\)), tail lights (\(P_{31_{No eHMI}} = 2\)), and quartet back windows (\(P_{33_{No eHMI}} = 2\)). 

This makes \(P_{34_{No eHMI}} = 1\), \(P_{36_{No eHMI}} = 1\), \(P_{37_{No eHMI}} = 1\), \(P_{38_{No eHMI}} = 1\), \(P_{39_{No eHMI}} = 1\), and \(P_{41_{No eHMI}} = 1\).

\subsubsection{Front Braking Lights (FBL)} 
These are placed under the headlights, on the front bumper (\(P_{12_{FBL}} = 2\)). This makes \(P_{34_{FBL}} = 1\), \(P_{36_{FBL}} = 1\), \(P_{37_{FBL}} = 0\), \(P_{38_{FBL}} = 1\), \(P_{39_{FBL}} = 1\), and \(P_{41_{FBL}} = 1\).

\subsubsection{Knight Rider Display (KRD)}
While this proposal has two separate elements, both of these serve the same purpose, so they are treated as the same for this section of the evaluation. One of the displays is on the front bumper (\(P_{12_{KRD}} = 2\)) and front plate (\(P_{16_{KRD}} = 2\)), while the other is on the hood of the vehicle (\(P_{22_{KRD}} = 2\)). This makes \(P_{34_{KRD}} = 1\), \(P_{36_{KRD}} = 1\), \(P_{37_{KRD}} = 0\), \(P_{38_{KRD}} = 1\), \(P_{39_{KRD}} = 1\), and \(P_{41_{KRD}} = 1\).

\subsubsection{Bumper Smile Display (BSD)}
This display is on the front bumper (\(P_{12_{BSD}} = 2\)) and front plate (\(P_{16_{BSD}} = 2\)). This makes \(P_{34_{BSD}} = 1\), \(P_{36_{BSD}} = 1\), \(P_{37_{BSD}} = 0\), \(P_{38_{BSD}} = 1\), \(P_{39_{BSD}} = 1\), and \(P_{41_{BSD}} = 1\).

\subsubsection{Bumper Text Display (BTD)}
This display is on the front bumper (\(P_{12_{BTD}} = 2\)) and front plate (\(P_{16_{BTD}} = 2\)). This makes \(P_{34_{BTD}} = 1\), \(P_{36_{BTD}} = 1\), \(P_{37_{BTD}} = 0\), \(P_{38_{BTD}} = 1\), \(P_{39_{BTD}} = 1\), and \(P_{41_{BTD}} = 1\).

\begin{longtable}[hc]{|lccccc|}
\caption{Positioning Questionnaire Results}
\label{tab:result6_table}\\
\hline
\multicolumn{6}{|c|}{\textbf{POSITIONING RESULTS}} \\ \hline
\endfirsthead
\multicolumn{6}{c}%
{{\bfseries Table \thetable\ continued from previous page}} \\
\hline
\multicolumn{6}{|c|}{\textbf{POSITIONING RESULTS}} \\ \hline
\endhead
\multicolumn{1}{|l|}{} &
  \multicolumn{1}{c|}{\textbf{No eHMI}} &
  \multicolumn{1}{c|}{\textbf{FBL}} &
  \multicolumn{1}{c|}{\textbf{KRD}} &
  \multicolumn{1}{c|}{\textbf{BSD}} &
  \textbf{BTD} \\ \hline
\multicolumn{1}{|l|}{\textbf{P1}} &
  \multicolumn{1}{c|}{1} &
  \multicolumn{1}{c|}{1} &
  \multicolumn{1}{c|}{1} &
  \multicolumn{1}{c|}{1} &
  1 \\ \hline
\multicolumn{1}{|l|}{\textbf{P2}} &
  \multicolumn{1}{c|}{0} &
  \multicolumn{1}{c|}{0} &
  \multicolumn{1}{c|}{0} &
  \multicolumn{1}{c|}{0} &
  0 \\ \hline
\multicolumn{1}{|l|}{\textbf{P3}} &
  \multicolumn{1}{c|}{0} &
  \multicolumn{1}{c|}{0} &
  \multicolumn{1}{c|}{0} &
  \multicolumn{1}{c|}{0} &
  0 \\ \hline
\multicolumn{1}{|l|}{\textbf{P4}} &
  \multicolumn{1}{c|}{0} &
  \multicolumn{1}{c|}{0} &
  \multicolumn{1}{c|}{0} &
  \multicolumn{1}{c|}{0} &
  0 \\ \hline
\multicolumn{1}{|l|}{\textbf{P5}} &
  \multicolumn{1}{c|}{2} &
  \multicolumn{1}{c|}{0} &
  \multicolumn{1}{c|}{0} &
  \multicolumn{1}{c|}{0} &
  0 \\ \hline
\multicolumn{1}{|l|}{\textbf{P6}} &
  \multicolumn{1}{c|}{0} &
  \multicolumn{1}{c|}{0} &
  \multicolumn{1}{c|}{0} &
  \multicolumn{1}{c|}{0} &
  0 \\ \hline
\multicolumn{1}{|l|}{\textbf{P7}} &
  \multicolumn{1}{c|}{0} &
  \multicolumn{1}{c|}{0} &
  \multicolumn{1}{c|}{0} &
  \multicolumn{1}{c|}{0} &
  0 \\ \hline
\multicolumn{1}{|l|}{\textbf{P8}} &
  \multicolumn{1}{c|}{2} &
  \multicolumn{1}{c|}{0} &
  \multicolumn{1}{c|}{0} &
  \multicolumn{1}{c|}{0} &
  0 \\ \hline
\multicolumn{1}{|l|}{\textbf{P9}} &
  \multicolumn{1}{c|}{0} &
  \multicolumn{1}{c|}{0} &
  \multicolumn{1}{c|}{0} &
  \multicolumn{1}{c|}{0} &
  0 \\ \hline
\multicolumn{1}{|l|}{\textbf{P10}} &
  \multicolumn{1}{c|}{0} &
  \multicolumn{1}{c|}{0} &
  \multicolumn{1}{c|}{0} &
  \multicolumn{1}{c|}{0} &
  0 \\ \hline
\multicolumn{1}{|l|}{\textbf{P11}} &
  \multicolumn{1}{c|}{0} &
  \multicolumn{1}{c|}{0} &
  \multicolumn{1}{c|}{0} &
  \multicolumn{1}{c|}{0} &
  0 \\ \hline
\multicolumn{1}{|l|}{\textbf{P12}} &
  \multicolumn{1}{c|}{0} &
  \multicolumn{1}{c|}{2} &
  \multicolumn{1}{c|}{2} &
  \multicolumn{1}{c|}{2} &
  2 \\ \hline
\multicolumn{1}{|l|}{\textbf{P13}} &
  \multicolumn{1}{c|}{0} &
  \multicolumn{1}{c|}{0} &
  \multicolumn{1}{c|}{0} &
  \multicolumn{1}{c|}{0} &
  0 \\ \hline
\multicolumn{1}{|l|}{\textbf{P14}} &
  \multicolumn{1}{c|}{2} &
  \multicolumn{1}{c|}{0} &
  \multicolumn{1}{c|}{0} &
  \multicolumn{1}{c|}{0} &
  0 \\ \hline
\multicolumn{1}{|l|}{\textbf{P15}} &
  \multicolumn{1}{c|}{2} &
  \multicolumn{1}{c|}{0} &
  \multicolumn{1}{c|}{0} &
  \multicolumn{1}{c|}{0} &
  0 \\ \hline
\multicolumn{1}{|l|}{\textbf{P16}} &
  \multicolumn{1}{c|}{0} &
  \multicolumn{1}{c|}{0} &
  \multicolumn{1}{c|}{2} &
  \multicolumn{1}{c|}{2} &
  2 \\ \hline
\multicolumn{1}{|l|}{\textbf{P17}} &
  \multicolumn{1}{c|}{2} &
  \multicolumn{1}{c|}{0} &
  \multicolumn{1}{c|}{0} &
  \multicolumn{1}{c|}{0} &
  0 \\ \hline
\multicolumn{1}{|l|}{\textbf{P18}} &
  \multicolumn{1}{c|}{2} &
  \multicolumn{1}{c|}{0} &
  \multicolumn{1}{c|}{0} &
  \multicolumn{1}{c|}{0} &
  0 \\ \hline
\multicolumn{1}{|l|}{\textbf{P19}} &
  \multicolumn{1}{c|}{0} &
  \multicolumn{1}{c|}{0} &
  \multicolumn{1}{c|}{0} &
  \multicolumn{1}{c|}{0} &
  0 \\ \hline
\multicolumn{1}{|l|}{\textbf{P20}} &
  \multicolumn{1}{c|}{0} &
  \multicolumn{1}{c|}{0} &
  \multicolumn{1}{c|}{0} &
  \multicolumn{1}{c|}{0} &
  0 \\ \hline
\multicolumn{1}{|l|}{\textbf{P21}} &
  \multicolumn{1}{c|}{2} &
  \multicolumn{1}{c|}{0} &
  \multicolumn{1}{c|}{0} &
  \multicolumn{1}{c|}{0} &
  0 \\ \hline
\multicolumn{1}{|l|}{\textbf{P22}} &
  \multicolumn{1}{c|}{0} &
  \multicolumn{1}{c|}{0} &
  \multicolumn{1}{c|}{2} &
  \multicolumn{1}{c|}{0} &
  0 \\ \hline
\multicolumn{1}{|l|}{\textbf{P23}} &
  \multicolumn{1}{c|}{2} &
  \multicolumn{1}{c|}{0} &
  \multicolumn{1}{c|}{0} &
  \multicolumn{1}{c|}{0} &
  0 \\ \hline
\multicolumn{1}{|l|}{\textbf{P24}} &
  \multicolumn{1}{c|}{2} &
  \multicolumn{1}{c|}{0} &
  \multicolumn{1}{c|}{0} &
  \multicolumn{1}{c|}{0} &
  0 \\ \hline
\multicolumn{1}{|l|}{\textbf{P25}} &
  \multicolumn{1}{c|}{0} &
  \multicolumn{1}{c|}{0} &
  \multicolumn{1}{c|}{0} &
  \multicolumn{1}{c|}{0} &
  0 \\ \hline
\multicolumn{1}{|l|}{\textbf{P26}} &
  \multicolumn{1}{c|}{2} &
  \multicolumn{1}{c|}{0} &
  \multicolumn{1}{c|}{0} &
  \multicolumn{1}{c|}{0} &
  0 \\ \hline
\multicolumn{1}{|l|}{\textbf{P27}} &
  \multicolumn{1}{c|}{0} &
  \multicolumn{1}{c|}{0} &
  \multicolumn{1}{c|}{0} &
  \multicolumn{1}{c|}{0} &
  0 \\ \hline
\multicolumn{1}{|l|}{\textbf{P28}} &
  \multicolumn{1}{c|}{2} &
  \multicolumn{1}{c|}{0} &
  \multicolumn{1}{c|}{0} &
  \multicolumn{1}{c|}{0} &
  0 \\ \hline
\multicolumn{1}{|l|}{\textbf{P29}} &
  \multicolumn{1}{c|}{2} &
  \multicolumn{1}{c|}{0} &
  \multicolumn{1}{c|}{0} &
  \multicolumn{1}{c|}{0} &
  0 \\ \hline
\multicolumn{1}{|l|}{\textbf{P30}} &
  \multicolumn{1}{c|}{0} &
  \multicolumn{1}{c|}{0} &
  \multicolumn{1}{c|}{0} &
  \multicolumn{1}{c|}{0} &
  0 \\ \hline
\multicolumn{1}{|l|}{\textbf{P31}} &
  \multicolumn{1}{c|}{2} &
  \multicolumn{1}{c|}{0} &
  \multicolumn{1}{c|}{0} &
  \multicolumn{1}{c|}{0} &
  0 \\ \hline
\multicolumn{1}{|l|}{\textbf{P32}} &
  \multicolumn{1}{c|}{0} &
  \multicolumn{1}{c|}{0} &
  \multicolumn{1}{c|}{0} &
  \multicolumn{1}{c|}{0} &
  0 \\ \hline
\multicolumn{1}{|l|}{\textbf{P33}} &
  \multicolumn{1}{c|}{1} &
  \multicolumn{1}{c|}{0} &
  \multicolumn{1}{c|}{0} &
  \multicolumn{1}{c|}{0} &
  0 \\ \hline
\multicolumn{1}{|l|}{\textbf{P34}} &
  \multicolumn{1}{c|}{1} &
  \multicolumn{1}{c|}{1} &
  \multicolumn{1}{c|}{1} &
  \multicolumn{1}{c|}{1} &
  1 \\ \hline
\multicolumn{1}{|l|}{\textbf{P35}} &
  \multicolumn{1}{c|}{0} &
  \multicolumn{1}{c|}{0} &
  \multicolumn{1}{c|}{0} &
  \multicolumn{1}{c|}{0} &
  0 \\ \hline
\multicolumn{1}{|l|}{\textbf{P36}} &
  \multicolumn{1}{c|}{1} &
  \multicolumn{1}{c|}{1} &
  \multicolumn{1}{c|}{1} &
  \multicolumn{1}{c|}{1} &
  1 \\ \hline
\multicolumn{1}{|l|}{\textbf{P37}} &
  \multicolumn{1}{c|}{1} &
  \multicolumn{1}{c|}{0} &
  \multicolumn{1}{c|}{0} &
  \multicolumn{1}{c|}{0} &
  0 \\ \hline
\multicolumn{1}{|l|}{\textbf{P38}} &
  \multicolumn{1}{c|}{1} &
  \multicolumn{1}{c|}{0} &
  \multicolumn{1}{c|}{0} &
  \multicolumn{1}{c|}{0} &
  0 \\ \hline
\multicolumn{1}{|l|}{\textbf{P39}} &
  \multicolumn{1}{c|}{1} &
  \multicolumn{1}{c|}{1} &
  \multicolumn{1}{c|}{1} &
  \multicolumn{1}{c|}{1} &
  1 \\ \hline
\multicolumn{1}{|l|}{\textbf{P40}} &
  \multicolumn{1}{c|}{0} &
  \multicolumn{1}{c|}{0} &
  \multicolumn{1}{c|}{0} &
  \multicolumn{1}{c|}{0} &
  0 \\ \hline
\multicolumn{1}{|l|}{\textbf{P41}} &
  \multicolumn{1}{c|}{1} &
  \multicolumn{1}{c|}{1} &
  \multicolumn{1}{c|}{1} &
  \multicolumn{1}{c|}{1} &
  1 \\ \hline
\multicolumn{1}{|l|}{} &
  \multicolumn{1}{c|}{\textbf{No eHMI}} &
  \multicolumn{1}{c|}{\textbf{FBL}} &
  \multicolumn{1}{c|}{\textbf{KRD}} &
  \multicolumn{1}{c|}{\textbf{BSD}} &
  \textbf{GTD} \\ \hline
\multicolumn{1}{|l|}{\textbf{TOTAL}} &
  \multicolumn{1}{c|}{10.00} &
  \multicolumn{1}{c|}{6.67} &
  \multicolumn{1}{c|}{6.67} &
  \multicolumn{1}{c|}{6.67} &
  6.67 \\ \hline
\end{longtable}

\subsection{Readability Scores}
While plenty of research has been conducted on the readability of screens under different lighting conditions, these studies mainly focus on mobile devices with lower luminance than the vehicle display used in \cite{ledsigncity_car_nodate}. These results are reflected in Table \ref{tab:result7_table}. All proposals with a display can adjust their brightness based on weather and lighting conditions (\(R_{{1 to 3}_{KRD}} = 1, R_{{1 to 3}_{BSD}} = 1, R_{{1 to 3}_{BTD}} = 1\)).

Otherwise, there is no sufficiently empirical data to evaluate the visibility of the selected proposals in the specified conditions (\(R_{{1 to 40}_{No eHMI}} = 0, R_{{1 to 40}_{FBL}} = 0, R_{{4 to 40}_{KRD}} = 0, R_{{4 to 40}_{BSD}} = 0, R_{{4 to 40}_{BTD}} = 0\)).
\begin{longtable}[c]{|lccccc|}
\caption{Readability Questionnaire Results}
\label{tab:result7_table}\\
\hline
\multicolumn{6}{|c|}{\textbf{READABILITY RESULTS}} \\ \hline
\endfirsthead
\multicolumn{6}{c}%
{{\bfseries Table \thetable\ continued from previous page}} \\
\hline
\multicolumn{6}{|c|}{\textbf{READABILITY RESULTS}} \\ \hline
\endhead
\multicolumn{1}{|l|}{} &
  \multicolumn{1}{c|}{\textbf{No eHMI}} &
  \multicolumn{1}{c|}{\textbf{FBL}} &
  \multicolumn{1}{c|}{\textbf{KRD}} &
  \multicolumn{1}{c|}{\textbf{BSD}} &
  \textbf{BTD} \\ \hline
\multicolumn{1}{|l|}{\textbf{R1}} &
  \multicolumn{1}{c|}{0} &
  \multicolumn{1}{c|}{0} &
  \multicolumn{1}{c|}{1} &
  \multicolumn{1}{c|}{1} &
  1 \\ \hline
\multicolumn{1}{|l|}{\textbf{R2}} &
  \multicolumn{1}{c|}{0} &
  \multicolumn{1}{c|}{0} &
  \multicolumn{1}{c|}{1} &
  \multicolumn{1}{c|}{1} &
  1 \\ \hline
\multicolumn{1}{|l|}{\textbf{R3}} &
  \multicolumn{1}{c|}{0} &
  \multicolumn{1}{c|}{0} &
  \multicolumn{1}{c|}{1} &
  \multicolumn{1}{c|}{1} &
  1 \\ \hline
\multicolumn{1}{|l|}{\textbf{R4}} &
  \multicolumn{1}{c|}{0} &
  \multicolumn{1}{c|}{0} &
  \multicolumn{1}{c|}{0} &
  \multicolumn{1}{c|}{0} &
  0 \\ \hline
\multicolumn{1}{|l|}{\textbf{R5}} &
  \multicolumn{1}{c|}{0} &
  \multicolumn{1}{c|}{0} &
  \multicolumn{1}{c|}{0} &
  \multicolumn{1}{c|}{0} &
  0 \\ \hline
\multicolumn{1}{|l|}{\textbf{R6}} &
  \multicolumn{1}{c|}{0} &
  \multicolumn{1}{c|}{0} &
  \multicolumn{1}{c|}{0} &
  \multicolumn{1}{c|}{0} &
  0 \\ \hline
\multicolumn{1}{|l|}{\textbf{R7}} &
  \multicolumn{1}{c|}{0} &
  \multicolumn{1}{c|}{0} &
  \multicolumn{1}{c|}{0} &
  \multicolumn{1}{c|}{0} &
  0 \\ \hline
\multicolumn{1}{|l|}{\textbf{R8}} &
  \multicolumn{1}{c|}{0} &
  \multicolumn{1}{c|}{0} &
  \multicolumn{1}{c|}{0} &
  \multicolumn{1}{c|}{0} &
  0 \\ \hline
\multicolumn{1}{|l|}{\textbf{R9}} &
  \multicolumn{1}{c|}{0} &
  \multicolumn{1}{c|}{0} &
  \multicolumn{1}{c|}{0} &
  \multicolumn{1}{c|}{0} &
  0 \\ \hline
\multicolumn{1}{|l|}{\textbf{R10}} &
  \multicolumn{1}{c|}{0} &
  \multicolumn{1}{c|}{0} &
  \multicolumn{1}{c|}{0} &
  \multicolumn{1}{c|}{0} &
  0 \\ \hline
\multicolumn{1}{|l|}{\textbf{R11}} &
  \multicolumn{1}{c|}{0} &
  \multicolumn{1}{c|}{0} &
  \multicolumn{1}{c|}{0} &
  \multicolumn{1}{c|}{0} &
  0 \\ \hline
\multicolumn{1}{|l|}{\textbf{R12}} &
  \multicolumn{1}{c|}{0} &
  \multicolumn{1}{c|}{0} &
  \multicolumn{1}{c|}{0} &
  \multicolumn{1}{c|}{0} &
  0 \\ \hline
\multicolumn{1}{|l|}{\textbf{R13}} &
  \multicolumn{1}{c|}{0} &
  \multicolumn{1}{c|}{0} &
  \multicolumn{1}{c|}{0} &
  \multicolumn{1}{c|}{0} &
  0 \\ \hline
\multicolumn{1}{|l|}{\textbf{R14}} &
  \multicolumn{1}{c|}{0} &
  \multicolumn{1}{c|}{0} &
  \multicolumn{1}{c|}{0} &
  \multicolumn{1}{c|}{0} &
  0 \\ \hline
\multicolumn{1}{|l|}{\textbf{R15}} &
  \multicolumn{1}{c|}{0} &
  \multicolumn{1}{c|}{0} &
  \multicolumn{1}{c|}{0} &
  \multicolumn{1}{c|}{0} &
  0 \\ \hline
\multicolumn{1}{|l|}{\textbf{R16}} &
  \multicolumn{1}{c|}{0} &
  \multicolumn{1}{c|}{0} &
  \multicolumn{1}{c|}{0} &
  \multicolumn{1}{c|}{0} &
  0 \\ \hline
\multicolumn{1}{|l|}{\textbf{R17}} &
  \multicolumn{1}{c|}{0} &
  \multicolumn{1}{c|}{0} &
  \multicolumn{1}{c|}{0} &
  \multicolumn{1}{c|}{0} &
  0 \\ \hline
\multicolumn{1}{|l|}{\textbf{R18}} &
  \multicolumn{1}{c|}{0} &
  \multicolumn{1}{c|}{0} &
  \multicolumn{1}{c|}{0} &
  \multicolumn{1}{c|}{0} &
  0 \\ \hline
\multicolumn{1}{|l|}{\textbf{R19}} &
  \multicolumn{1}{c|}{0} &
  \multicolumn{1}{c|}{0} &
  \multicolumn{1}{c|}{0} &
  \multicolumn{1}{c|}{0} &
  0 \\ \hline
\multicolumn{1}{|l|}{\textbf{R20}} &
  \multicolumn{1}{c|}{0} &
  \multicolumn{1}{c|}{0} &
  \multicolumn{1}{c|}{0} &
  \multicolumn{1}{c|}{0} &
  0 \\ \hline
\multicolumn{1}{|l|}{\textbf{R21}} &
  \multicolumn{1}{c|}{0} &
  \multicolumn{1}{c|}{0} &
  \multicolumn{1}{c|}{0} &
  \multicolumn{1}{c|}{0} &
  0 \\ \hline
\multicolumn{1}{|l|}{\textbf{R22}} &
  \multicolumn{1}{c|}{0} &
  \multicolumn{1}{c|}{0} &
  \multicolumn{1}{c|}{0} &
  \multicolumn{1}{c|}{0} &
  0 \\ \hline
\multicolumn{1}{|l|}{\textbf{R23}} &
  \multicolumn{1}{c|}{0} &
  \multicolumn{1}{c|}{0} &
  \multicolumn{1}{c|}{0} &
  \multicolumn{1}{c|}{0} &
  0 \\ \hline
\multicolumn{1}{|l|}{\textbf{R24}} &
  \multicolumn{1}{c|}{0} &
  \multicolumn{1}{c|}{0} &
  \multicolumn{1}{c|}{0} &
  \multicolumn{1}{c|}{0} &
  0 \\ \hline
\multicolumn{1}{|l|}{\textbf{R25}} &
  \multicolumn{1}{c|}{0} &
  \multicolumn{1}{c|}{0} &
  \multicolumn{1}{c|}{0} &
  \multicolumn{1}{c|}{0} &
  0 \\ \hline
\multicolumn{1}{|l|}{\textbf{R26}} &
  \multicolumn{1}{c|}{0} &
  \multicolumn{1}{c|}{0} &
  \multicolumn{1}{c|}{0} &
  \multicolumn{1}{c|}{0} &
  0 \\ \hline
\multicolumn{1}{|l|}{\textbf{R27}} &
  \multicolumn{1}{c|}{0} &
  \multicolumn{1}{c|}{0} &
  \multicolumn{1}{c|}{0} &
  \multicolumn{1}{c|}{0} &
  0 \\ \hline
\multicolumn{1}{|l|}{\textbf{R28}} &
  \multicolumn{1}{c|}{0} &
  \multicolumn{1}{c|}{0} &
  \multicolumn{1}{c|}{0} &
  \multicolumn{1}{c|}{0} &
  0 \\ \hline
\multicolumn{1}{|l|}{\textbf{R29}} &
  \multicolumn{1}{c|}{0} &
  \multicolumn{1}{c|}{0} &
  \multicolumn{1}{c|}{0} &
  \multicolumn{1}{c|}{0} &
  0 \\ \hline
\multicolumn{1}{|l|}{\textbf{R30}} &
  \multicolumn{1}{c|}{0} &
  \multicolumn{1}{c|}{0} &
  \multicolumn{1}{c|}{0} &
  \multicolumn{1}{c|}{0} &
  0 \\ \hline
\multicolumn{1}{|l|}{\textbf{R31}} &
  \multicolumn{1}{c|}{0} &
  \multicolumn{1}{c|}{0} &
  \multicolumn{1}{c|}{0} &
  \multicolumn{1}{c|}{0} &
  0 \\ \hline
\multicolumn{1}{|l|}{\textbf{R32}} &
  \multicolumn{1}{c|}{0} &
  \multicolumn{1}{c|}{0} &
  \multicolumn{1}{c|}{0} &
  \multicolumn{1}{c|}{0} &
  0 \\ \hline
\multicolumn{1}{|l|}{\textbf{R33}} &
  \multicolumn{1}{c|}{0} &
  \multicolumn{1}{c|}{0} &
  \multicolumn{1}{c|}{0} &
  \multicolumn{1}{c|}{0} &
  0 \\ \hline
\multicolumn{1}{|l|}{\textbf{R34}} &
  \multicolumn{1}{c|}{0} &
  \multicolumn{1}{c|}{0} &
  \multicolumn{1}{c|}{0} &
  \multicolumn{1}{c|}{0} &
  0 \\ \hline
\multicolumn{1}{|l|}{\textbf{R35}} &
  \multicolumn{1}{c|}{0} &
  \multicolumn{1}{c|}{0} &
  \multicolumn{1}{c|}{0} &
  \multicolumn{1}{c|}{0} &
  0 \\ \hline
\multicolumn{1}{|l|}{\textbf{R36}} &
  \multicolumn{1}{c|}{0} &
  \multicolumn{1}{c|}{0} &
  \multicolumn{1}{c|}{0} &
  \multicolumn{1}{c|}{0} &
  0 \\ \hline
\multicolumn{1}{|l|}{\textbf{R37}} &
  \multicolumn{1}{c|}{0} &
  \multicolumn{1}{c|}{0} &
  \multicolumn{1}{c|}{0} &
  \multicolumn{1}{c|}{0} &
  0 \\ \hline
\multicolumn{1}{|l|}{} &
  \multicolumn{1}{c|}{\textbf{No eHMI}} &
  \multicolumn{1}{c|}{\textbf{FBL}} &
  \multicolumn{1}{c|}{\textbf{KRD}} &
  \multicolumn{1}{c|}{\textbf{BSD}} &
  \textbf{GTD} \\ \hline
\multicolumn{1}{|l|}{\textbf{TOTAL}} &
  \multicolumn{1}{c|}{0} &
  \multicolumn{1}{c|}{0} &
  \multicolumn{1}{c|}{0.81} &
  \multicolumn{1}{c|}{0.81} &
  0.81 \\ \hline
\end{longtable}

\subsection{Final Results}
Table \ref{tab:finalresults_table} displays the final score for each category, once they're added together for each eHMI proposal.

\begin{table}[h]
\centering
\caption{Evaluation Results}
\label{tab:finalresults_table}
\begin{tabular}{|lccccc|}
\hline
\multicolumn{6}{|c|}{\textbf{FINAL RESULTS}} \\ \hline
\multicolumn{1}{|l|}{} &
  \multicolumn{1}{c|}{\textbf{No eHMI}} &
  \multicolumn{1}{c|}{\textbf{FBL}} &
  \multicolumn{1}{c|}{\textbf{KRD}} &
  \multicolumn{1}{c|}{\textbf{BSD}} &
  \textbf{BTD} \\ \hline
\multicolumn{1}{|l|}{\textbf{Standarization}} &
  \multicolumn{1}{c|}{10.00} &
  \multicolumn{1}{c|}{9.63} &
  \multicolumn{1}{c|}{5.93} &
  \multicolumn{1}{c|}{8.52} &
  7.41 \\ \hline
\multicolumn{1}{|l|}{\textbf{Cost Effectiveness}} &
  \multicolumn{1}{c|}{10.00} &
  \multicolumn{1}{c|}{9.95} &
  \multicolumn{1}{c|}{8.56} &
  \multicolumn{1}{c|}{8.56} &
  8.56 \\ \hline
\multicolumn{1}{|l|}{\textbf{Accessibility}} &
  \multicolumn{1}{c|}{3.56} &
  \multicolumn{1}{c|}{4.25} &
  \multicolumn{1}{c|}{4.79} &
  \multicolumn{1}{c|}{4.79} &
  6.30 \\ \hline
\multicolumn{1}{|l|}{\textbf{Ease of Understanding}} &
  \multicolumn{1}{c|}{0.81} &
  \multicolumn{1}{c|}{0.93} &
  \multicolumn{1}{c|}{0.99} &
  \multicolumn{1}{c|}{0.93} &
  0.95 \\ \hline
\multicolumn{1}{|l|}{\textbf{Constant Communication}} &
  \multicolumn{1}{c|}{0.83} &
  \multicolumn{1}{c|}{0.42} &
  \multicolumn{1}{c|}{1.25} &
  \multicolumn{1}{c|}{1.25} &
  1.67 \\ \hline
\multicolumn{1}{|l|}{\textbf{Positioning}} &
  \multicolumn{1}{c|}{10.00} &
  \multicolumn{1}{c|}{6.67} &
  \multicolumn{1}{c|}{6.67} &
  \multicolumn{1}{c|}{6.67} &
  6.67 \\ \hline
\multicolumn{1}{|l|}{\textbf{Readability}} &
  \multicolumn{1}{c|}{0.00} &
  \multicolumn{1}{c|}{0.00} &
  \multicolumn{1}{c|}{0.81} &
  \multicolumn{1}{c|}{0.81} &
  0.81 \\ \hline
\multicolumn{1}{|l|}{\textbf{TOTAL}} &
  \multicolumn{1}{c|}{\textbf{35.21}} &
  \multicolumn{1}{c|}{\textbf{31.84}} &
  \multicolumn{1}{c|}{\textbf{29.00}} &
  \multicolumn{1}{c|}{\textbf{31.53}} &
  \textbf{32.37} \\ \hline
\multicolumn{1}{|l|}{\textbf{\%}} &
  \multicolumn{1}{c|}{\textbf{50.30\%}} &
  \multicolumn{1}{c|}{\textbf{45.48\%}} &
  \multicolumn{1}{c|}{\textbf{41.43\%}} &
  \multicolumn{1}{c|}{\textbf{45.04\%}} &
  \textbf{46.24\%} \\ \hline
\end{tabular}
\end{table}

\subsubsection{Highest Scores}
The highest scoring eHMI proposals were No eHMI (50.30\%), the Bumper Text Display (46.24\%) and Frontal Brake Lights (45.48\%) . 

The results show the vehicle kinematics are extremely cost effective, easy to standardize, and positioned where the target audience can see it, since it occupies so much of the vehicle exterior. However, this comes at the expense of a lower ease of understanding, accessibility, and readability than the other eHMI proposals.

The Bumper Text Display is the most accessible and communicative proposal, but it's less cost effective and requires more work to standardize when compared to the other two highest-scoring proposals. The positioning is also not ideal by itself, and should be complemented with other elements.

Finally, the Frontal Brake Lights are highly cost effective and easy to standardize, but communicate the lowest amount of information across all proposals tested. 

\subsubsection{Lowest Scores}
The lowest-scoring proposal was the Knight Rider Display. This is mainly due to its difficulty to standardize, and otherwise average results in the other categories when compared to the other proposals. 

The second lowest-scoring proposal belongs to the Bumper Smile Display, which scored higher than the Knight Rider Display in standardization, slightly lower (0.5 points) in ease of understanding, and the same in the other categories.

\subsection{Recommendations and Future Work}
Based on these initial results, obtained by using the questionnaire postulated in this work, the ideal solution to communicate yielding intent to pedestrians includes a combination of intentionally-designed vehicle kinematics and a text-based display placed on the vehicle, perhaps coupled with another display somewhere else on the vehicle that complements the bumper. The results of \cite{gonzalez-belmonte_analytical_2026} recommend a distributive approach, making use of the front fenders, windshield, or side mirrors, as they're as highly visible spots for observers focused on the front of the vehicle, although the applicability of this design with a text-based display has yet to be tested.

This said, the highest score obtained in this testing was that of a 50.30\%, revealing many gaps in the knowledge regarding eHMI design:

\begin{itemize}
    \item In the case of vehicle kinematics, there is no research showing how many times it takes a pedestrian to learn how to interpret the vehicle's motions as a sign of yielding. Since this response is unconscious and often learned at a young age, it will be difficult to obtain more data for this specific case.

    \item There is limited recorded data on the amount of times it takes a variety of demographics to learn the meaning behind the different eHMIs. Present data exists for adult pedestrians, but little research has been done in the reaction of minors and the elderly, as well a cyclists and other drivers.
    
    \item Readability also suffers from a lack of recorded data. Previous simulations for eHMIs have been conducted either on clear days \cite{loew_go_2022}, in virtual reality simulations with controlled static conditions \cite{de_clercq_external_2019}, or using videos and images \cite{guo_video-based_2022, bazilinskyy_survey_2019}. Research has yet to be done on readability during different weather conditions, time of day, distance, and angles. 

    \item All evaluations of cost effectiveness were done using publicly available data and prices. This may not reflect the actual manufacturer cost for some of the proposed devices.

    \item This are only the results from evaluating five eHMI proposals focused on one specific task (communicating yielding behavior to the pedestrian). Future research may apply this study to other types of methods of communication.

    \item The data in this work was based on information from 2022, and as advances are made in the realm of autonomous vehicles we are likely to see much of it in need of an update soon.
\end{itemize}

Finally, while in this example all guidelines of General eHMI Evaluation Method were weighted equally, this may not be the best course of action in the future.  Future work should be done to determine levels of priority and determine the best weight for each guideline.

\section{Conclusion} \label{Conclusion}

This paper puts forward a standardized methodology to evaluate the quality of an external Human-Machine Interface, and proves its applicability and flexibility by evaluating five popular eHMI proposals. Initial testing of this methodology suggest a combination of vehicle kinematics and well-positioned text displays as the best methods of communication for incoming AVs and other agents on the road, but additional research must be done in readability and learning in order to reach a full conclusion. Future work may also include evaluating other methods of external communication, or testing different weights for the final score depending on what aspect of the technology is considered more important. Either way, this should prove a valuable tool for future research aiming to standardize communication between AVs and road users such as pedestrians and other drivers, as it provides a summary of a variety of specific guidelines for eHMI design and a method of quantifying their contributions to the user experience for comparisons.

\section{Declarations}
\subsection{Funding}
Jaerock Kwon is an associate professor for University of Michigan Dearborn. Jose Gonzalez-Belmonte is an assistant professor for Lawrence Technological University, and his ongoing studies are being partially funded by it. The authors did not receive support from any other organization for the submitted work. 
\subsection{Conflict of Interest}
The author of this paper declares that they have no known competing financial interests or personal relationships that could have appeared to influence the work reported in this paper.
\subsection{Ethics approval and consent to participate}
Not applicable.
\subsection{Consent for publication}
Not applicable.
\subsection{Data availability }
All data used for this paper can be found at https://jgonzalez-uom.github.io/review-method
\subsection{Materials availability}
Not applicable.
\subsection{Code availability }
Not applicable.
\subsection{Author contribution}
All authors contributed to the study conception and design. Material preparation, software development, data collection, and analysis were performed by Jose Gonzalez-Belmonte. The first draft of the manuscript was written by Jose Gonzalez-Belmonte, and Jaerock Kwon commented and suggested edits on previous versions of the manuscript. All authors have read and approved the final manuscript.  
\subsection{Acknowledgements}
We would like to thank Dr. Joost de Winter for providing the Unity project used in \cite{de_clercq_external_2019}, which provided an accurate comparison for the evaluation of the eHMI proposals used in that study.






\begin{appendices}

\section{Evaluation Questionnaire for external Human-Machine Interfaces}\label{secA1}

This questionnaire is a way to evaluate an external Human-Machine Interface (eHMI) proposal. The evaluation is divided on seven categories: Standardization, Cost Effectiveness, Accessibility, Ease of Understanding, Constant Communication, Positioning, and Readability. Each category is evaluated by its own separate questionnaire and formula, in order to produce a score between 0 and 10. Each score is multiplied by a weighted value, then added together to produce a final score between 0 and 70.

\subsection{How to Use the General eHMI Evaluation Method}
Most questions in these questionnaires can be answered as “Yes” (1), “No” (0), “Unknown” (0), or with a number prompted by the question. Unless otherwise stated, “Not Applicable” should be answered as 0.

This method is meant to provide an objective way to compare multiple eHMI proposals, as well as to establish guidelines for their implementation; as such some of the variables are designed to be modified according to the needs of the user. Note that different eHMIs may have different priorities, and as such neither the final score or any of the individual category score should be taken out of context for comparison purposes.

\subsection{Filling the Questionnaires and Calculating the Scores}
\subsubsection{Standardization Score (\(S_S\))}
\begin{enumerate}
    \item Identify all non-identical elements of the eHMI proposal. For example, if a display was placed on the front of a vehicle showing both text and pictograms, this would be considered as two different elements to be evaluated independently; however, two separate identical displays showing the same information count as one display.
    \item Assign each element a unique natural number, starting with 1. This will be the identifier of that element.
    \item For each element, answer each question of the Standardization Questionnaire in Table \ref{tab:QS_table}
    \item For each element, calculate \(S_{P_x}\) using \ref{SPx:1} where \(S_{count}\) corresponds to the total number of questions in Table \ref{tab:QS_table}, and \(S_{x_n}\) is the value obtained by answering question \(S_n\) in respect for the element with identifier \(x\).
    \begin{equation} \label{SPx:1}
        S_{P_x} = \sum_{n=1}^{S_{count}} S_{x_n} 
    \end{equation}
    
    \item Obtain the Standardization Score (\(S_S\)) making use of \ref{S_S:2}, where \(S_B=31\), and \(x_{count}\) is the number of elements identified.
        \begin{equation} \label{S_S:2}
        S_S = \frac{S_B - \sum_{n=1}^{x_{count}} S_{P_{n}}}{S_{count}} \times 10  
        \end{equation}
        
        \[
            S_S =
        \begin{cases}
            0, & \text{if } S_S < 0\\
            10, & \text{if } S_S > 10\\
            S_S, & \text{otherwise}
        \end{cases}
        \]
\end{enumerate}

\begin{longtblr}[
  caption = {Standardization Questionnaire},
  label = {tab:QS_table},
]{
  width = \linewidth,
  colspec = {Q[40]Q[79]Q[821]},
  row{1} = {c,font=\bfseries},
  cell{1}{1} = {c=3}{0.94\linewidth},
  cell{3}{3} = {font=\bfseries},
  cell{36}{3} = {font=\bfseries},
  hlines,
  vlines,
}
STANDARDIZATION   &  & \\
ID & PTS & QUESTION\\
 &  & GENERAL\\
S1 & 1 & PLACEMENT: The proposal does not specify the eHMI's specific placement on the vehicle (e.g. grill, windshield, door, etc)\\
S2 & 1 & ON STATE: The proposed eHMI has a display that can turn on and off.\\
S3 & 4 & USES TEXT: The proposed eHMI makes use of text, requiring standardization of font, letter spacing, color, and size.\\
 &  & {\textbf{FRAME}\\\textbf{If this eHMI does not use a frame, answer S4 through S6 with “0”.}\\\textbf{If this eHMI doesn't specify whether it uses a frame, answer S4 through S6 with "1"}}\\
S4 & 1 & FRAME COLOR: The eHMI has a frame around it that can be made in more than one color.\\
S5 & 1 & FRAME THICKNESS: The eHMI has a frame around it that can be made in more than one thickness.\\
S6 & 1 & FRAME BRIGHTNESS: The eHMI has a frame around it that lights up at more than one possible brightness.\\
 &  & {\textbf{BACKGROUND}\\\textbf{If this eHMI does not use a background, answer S7 through S9 with “0”.}\\\textbf{If this eHMI doesn't specify whether it uses a frame, answer S7 through S9 with "1".}}\\
S7 & 1 & BACKGROUND COLOR: The eHMI has a uniform background that can be made in more than one possible color, but does not change colors dynamically.\\
S8 & 1 & BACKGROUND CONTENT: The eHMI uses a pattern or image as a background, but cannot change it dynamically\\
S9 & Bcc & BACKGROUND CHANGE CONDITIONS (Bcc: the number of conditions that could trigger a change in the background.)\\
 &  & {\textbf{PICTOGRAM}\\\textbf{If this eHMI does not use pictograms, answer S10 through S15 with “0”.}}\\
S10 & Pdc & PICTOGRAM DESIGN CHANGE CONDITIONS (Pdc: the number of conditions that could trigger a change in the pictogram design.)\\
S11 & 1 & PICTOGRAM COLOR: The pictogram can be made in multiple colors, and as such requires color standardization.\\
S12 & Pcc & PICTOGRAM COLOR CHANGE CONDITIONS (Pcc: the number of conditions that could trigger a change in the pictogram color.)\\
S13 & Psc & PICTOGRAM SIZE CHANGE CONDITIONS (Psc: the number of conditions that could trigger a change in the pictogram size.)\\
S14 & Pmc & PICTOGRAM MARGIN CHANGE CONDITIONS (Pmc: the number of conditions that could trigger a change in the pictogram margins.)\\
S15 & Pv - 1 & PICTOGRAM VARIANTS (Pv: all the possible states of this display, made from the possible combinations of pictogram design, color, size, and margins.)\\
 &  & {\textbf{TEXT}\\\textbf{If this eHMI does not use text, answer S16 through S22 with “0”.}}\\
S16 & Ttc & TEXT CHANGE CONDITIONS (Ttc: the number of conditions that could trigger a change in the text content.)\\
S17 & 1 & TEXT COLOR: The text can be made to display multiple colors, and as such requires color standardization.\\
S18 & Tcc & TEXT COLOR CHANGE CONDITIONS (Tcc: the number of conditions that could trigger a change in the text color.)\\
S19 & Tf & TEXT FONT CHANGE CONDITIONS (Tf: the number of conditions that could trigger a change in the text typeface.)\\
S20 & Ts & TEXT SIZE CHANGE CONDITIONS (Ts: the number of conditions that could trigger a change in the text size.)\\
S21 & Tm & TEXT MARGIN CHANGE CONDITIONS (Tm: the number of conditions that could trigger a change in the text margins.)\\
S22 & Tv - 1 & TEXT VARIANT CHANGE CONDITIONS (Tv: all the possible states of this display, made from the possible combinations of: Text content, text color, text font, font size, font spacing, and text margins. )\\
 &  & {\textbf{ANIMATION}\\\textbf{If this eHMI does not use animations or motions, answer S26 with “0”}}\\
S23 & {(Ams / 1000 \\+ Aks \\+ Arc)} & {TOTAL MOTION PLAYTIME: \\(Ams: The sum of all milliseconds taken by each individual animation that this eHMI plays (e.g. swiping, growing, fade in, etc), rounding up.\\Aks: The number of kinematic actions that this eHMI employs.\\Arc: The amount of animations that repeat intermittently.).}\\
 &  & {\textbf{SOUND}\\\textbf{If this eHMI does not use sound, answer S27 through S29 with “0”}}\\
S24 & Sc & SOUND COUNT (Sc: the amount of sounds used by this eHMI. )\\
S25 & Ssc & SPEAKERS COUNT (Ssc: The amount of speakers used by this eHMI.)\\
S26 & Sc & PITCH AND VOLUME (Sa: the amount of sounds used by this eHMI.)\\
 &  & OTHERS\\
S27 & Ot & OTHER TRIGGERS (Ot: the amount of conditions that trigger any change in the eHMI otherwise not mentioned in this questionnaire. )
\end{longtblr}

\subsubsection{Cost Effectiveness Score (\(S_{CE}\))}
Answer the Cost Effectiveness Questionnaire found in Table \ref{tab:QCE_table}, in United States dollars, and adjusting all monetary for inflation for 2022. Then calculate the Cost Effectiveness Score (\(S_{CE}\)) using \ref{S_CE:3}, where \(CE_{count}\) is the number of questions in Table \ref{tab:QCE_table}, \(CE_{n}\) is the value obtained after answering the corresponding question, and  \(B_{CE}=48 301\).
    \begin{equation} \label{S_CE:3}
        S_{CE} = \frac{B_{CE} - \sum_{n=1}^{CE_{count}} CE_n}{B_{CE}} \times 10  
    \end{equation}
        
        \[
            S_{CE} =
        \begin{cases}
            0, & \text{if } S_{CE} < 0\\
            10, & \text{if } S_{CE} > 10\\
            S_{CE}, & \text{otherwise}
        \end{cases}
        \]

\begin{longtblr}[
  caption = {Cost Effectiveness Questionnaire},
  label = {tab:QCE_table},
]{
  width = \linewidth,
  colspec = {Q[40]Q[75]Q[825]},
  row{1} = {font=\bfseries},
  row{2} = {font=\bfseries},
  cell{1}{1} = {c=3}{0.94\linewidth},
  cell{3}{3} = {font=\bfseries},
  cell{4}{3} = {font=\bfseries},
  cell{5}{3} = {font=\bfseries},
  hlines,
  vlines,
}
COST EFFECTIVENESS   &  & \\
ID & PTS & QUESTION\\
 &  & All costs should be adjusted for inflation for 2022\\
 &  & Use media value when a range is provided.\\
 &  & If a value is unknown, use the highest value from CE1 through CE5.\\
CE1 & Buy & AFTER MARKET COST (Man: The average cost of buying a single set of this technology, without accounting for product development or infrastructure.)\\
CE2 & InsN * 0.75 & COST OF INSTALLATION ON NEW VEHICLE (ImpN: The average cost of implementing this technology in a new vehicle.)\\
CE3 & InsE & COST OF INSTALLATION ON EXISTING VEHICLE (ImpE: The average cost of implementing this technology in an existing, operating vehicle.)\\
CE4 & Mai & MAINTENANCE COST (Mai: The average yearly cost of maintaining this technology once installed.)\\
CE5 & OC & OPERATION COST (OC: Without accounting for maintenance and replacements, but accounting for any resource consumption such as energy and gasoline, the average yearly cost of operating this technology as intended in a single vehicle.)
\end{longtblr}

\subsubsection{Accessibility Score (\(S_{A}\))}
Answer the Accessibility Questionnaire found in Table \ref{tab:QA_table}. Then calculate the Accessibility Score (\(S_{A}\)) using \ref{S_A:4}, where \(A_{count}\) is the number of questions in Table \ref{tab:QA_table}, and \(A_{n}\) is the value obtained after answering the corresponding question. 
    \begin{equation} \label{S_A:4}
        S_{A} = \frac{\sum_{n=1}^{A_{count}} A_n}{A_{count}} \times 10  
    \end{equation}

\begin{longtblr}[
  caption = {Accessibility Questionnaire},
  label = {tab:QA_table},
]{
  width = \linewidth,
  colspec = {Q[37]Q[58]Q[844]},
  row{1} = {c,font=\bfseries},
  row{2} = {font=\bfseries},
  cell{1}{1} = {c=3}{0.939\linewidth},
  cell{3}{3} = {font=\bfseries},
  cell{8}{3} = {font=\bfseries},
  cell{52}{3} = {font=\bfseries},
  cell{58}{3} = {font=\bfseries},
  cell{64}{3} = {font=\bfseries},
  hlines,
  vlines,
}
ACCESSIBILITY   &  & \\
ID & PTS & QUESTION\\
 &  & GENERAL\\
A1 & 1 & All visible eHMI in this proposal use pictograms.\\
A2 & 1 & All visible eHMI in this proposal use text.\\
A3 & 1 & This proposal includes a tactile component.\\
A4 & 1 & This proposal includes sound cues.\\
 &  & VISUAL - GENERAL\\
A5 & 1 & The placement of information is not affected by vehicle type or size.\\
A6 & 1 & All information is placed in an isolated space with little visual clutter.\\
A7 & 1 & All information is consistently displayed in the same location.\\
A8 & 1 & Measures are taken to prevent the creation of shadows on or caused by the information.\\
 &  & {\textbf{VISUAL - PICTOGRAMS}\\\textbf{If answered "No" to A1, answer A9 through A14 with “0”}}\\
A9 & 1 & All pictograms contrast with their surroundings.\\
A10 & 1 & All pictograms are always upright.\\
A11 & 1 & All pictograms use bold, uniform, thick lines.\\
A12 & 1 & All pictograms use straight lines.\\
A13 & 1 & All pictograms are fully distinct from each other with no shared elements.\\
A14 & 1 & All pictograms and their backgrounds are non-reflective and anti-glare.\\
 &  & {\textbf{VISUAL - BACKGROUND}\\\textbf{If answered "No" to A1 and A2, answer A15 through A17 with “0”}}\\
A15 & 1 & All eHMI(s) in this proposal use a single color background.\\
A16 & 1 & All eHMI(s) in this proposal use dark text in a light background. If not applicable answer with "1".\\
A17 & 1 & All eHMI(s) in this proposal avoid pure-white bright white background.\\
 &  & {\textbf{VISUAL - TEXT}\\\textbf{If answered "No" to A2, answer A18 through A29 with “0”}}\\
A18 & 1 & All text contrasts with its surroundings.\\
A19 & 1 & All text is straight and readable.\\
A20 & 1 & None of the text is in italics.\\
A21 & 1 & All text uses bold, uniform, thick lines.\\
A22 & 1 & All text uses straight lines.\\
A23 & 1 & All text characters are distinct from each other.\\
A24 & 1 & All text characters include prominent ascenders, descenders, and open counterforms.\\
A25 & 1 & All text and its background are non-reflective (No glare).\\
A26 & 1 & All text uses initial uppercase letters and is not in all caps.\\
A27 & 1 & All text fonts and typefaces have wide horizontal proportions.\\
A28 & 1 & The text does not rotate.\\
A29 & 1 & The text does not include hyphens (-) and em dashes (—).\\
 &  & {\textbf{VISUAL - COLOR}\\\textbf{If non applicable, answer A30 and A34 with "1"}}\\
A30 & 1 & The eHMI(s) in this proposal do not depend exclusively on color to communicate any information.\\
A31 & 1 & The proposal avoids combinations of colors most commonly confused by color blind people, such as red, green, brown, and orange.\\
A32 & (Cu / Cuu) & {COLOR ASSOCIATION \\(Cuu: Number of colors used by this eHMI - such as cyan, blue-green, and turquoise - that do not have a previous association in the road.\\Cu: Total number of colors used by this eHMI)\\If this proposal does not use color, answer with a 1.}\\
A33 & 1 & The color(s) used by the visual eHMIs in this proposal can be adjusted during manufacturing or shipping to adjust to local regulation.\\
A34 & 1 & The proposal includes in it a color or design for the eHMI(s) so that it will not blend in with the rest of the vehicle, regardless of its paint job.\\
 &  & {\textbf{TACTILE}\\\textbf{If answered "No" to A3, answer A35 with “0”}}\\
A35 & 1 & This proposal has a tactile component without employing braille, or it uses braille with signage standardization (shape, size, distance, etc).\\
 &  & {\textbf{AUDITORY}\\\textbf{If answered "No" to A4, answer A36 through A41 with “0”}}\\
A36 & 1 & The speakers in this proposal use speakers that minimize echo.\\
A37 & 1 & All sound cues in this proposal are distinguishable from other road sounds.\\
A38 & 1 & All sound cues in this proposal are clearly distinguishable from each other.\\
A39 & 1 & All non-verbal signals have a maximum frequency of 1500 Hz\\
A40 & 1 & All audible verbal signals have a frequency between 300 Hz and 3000 Hz\\
A41 & 1 & All sounds have a volume of between 10 dB and 80 dB, above ambient.\\
 &  & NEUROLOGICAL\\
A42 & 1 & None of the visual eHMI(s) in this proposal use flashing lights.\\
A43 & 1 & None of the visual eHMI(s) in this proposal contain any elements that flash more than three times in any one second.\\
A44 & 1 & None of the visual eHMI(s) in this proposal use the color red.\\
A45 & 1 & None of the visual eHMI(s) in this proposal use animations, or all animations employed are simple moving elements.\\
A46 & 1 & None of the visual eHMI(s) in this proposal use animations with parallax.\\
 &  & COGNITIVE - GENERAL\\
A47 & 1 & This proposal uses text, all of which is short and concise.\\
A48 & 1 & All information provides enough time to be read and understood comfortably by the average pedestrian or driver.\\
A49 & 1 & All information is presented in more than one way at a time.\\
A50 & 1 & The eHMI(s) in this proposal presents information explicitly and directly to the observer (e.g. symbol aimed at pedestrians shows a person walking, instead of an arrow).\\
A51 & 1 & The eHMI(s) in this proposal present information in an egocentric approach to communicate information (i.e. “Walk” instead of “Stopping” when yielding). Score with 0 if not applicable.\\
 &  & COGNITIVE - LANGUAGE\\
A52 & 1 & The eHMI(s) in this proposal don't use text, or use text with plain language.\\
A53 & 1 & The eHMI(s) in this proposal don't use text or use text without making use of idioms (phrases whose meaning cannot be deduced from the meaning of the individual words).\\
 &  & {\textbf{COGNITIVE - CULTURAL}\\\textbf{Rounding up, what percentage of the non-textual components (pictograms, lights, sounds, tactile components, etc) of the total eHMI(s) in this proposal are not universal, and hence may hold different meanings depending on local factors (e.g. language, culture)? If not applicable, score A54 through A58 with 1}}\\
A54 & 1 & 80\% or less\\
A55 & 1 & 60\% or less\\
A56 & 1 & 40\% or less\\
A57 & 1 & 20\% or less\\
A58 & 1 & 0\%.\\
 &  & {\textbf{COGNITIVE - CONTEXTUAL}\\\textbf{Rounding up, what percentage of the non-textual components (pictograms, lights, sounds, tactile components, etc) of the total eHMI(s) in this proposal may hold different meanings depending on environmental factors (e.g. weather, time of day)? If not applicable, score A59 through A63 with "1"}}\\
A59 & 1 & 80\% or less\\
A60 & 1 & 60\% or less\\
A61 & 1 & 40\% or less\\
A62 & 1 & 20\% or less\\
A63 & 1 & 0\%.\\
 &  & {\textbf{PSYCHOLOGY - DISTRESSING IMAGERY}\\\textbf{Rounding up, what percentage of all symbols, text, sounds, and tactile components used by the eHMI(s) in this proposal could trigger adverse psychological reactions, such as depicting those in distress or feelings of hopelessness? If not applicable, score A64 through A68 with "1"}}\\
A64 & 1 & 80\% or less\\
A65 & 1 & 60\% or less\\
A66 & 1 & 40\% or less\\
A67 & 1 & 20\% or less\\
A68 & 1 & 0\%.\\
 &  & {\textbf{PSYCHOLOGY - LANGUAGE}\\\textbf{Rounding up, what percentage of all symbols, text, sounds, and tactile components used by the eHMI(s) in this proposal depict or use language that can be deemed offensive to those within the mental health community, such as “insane”, “crazy”, “maniac”, “normal”, etc. If not applicable, score A69 through A73 with "1"}}\\
A69 & 1 & 80\% or less\\
A70 & 1 & 60\% or less\\
A71 & 1 & 40\% or less\\
A72 & 1 & 20\% or less\\
A73 & 1 & 0\%.
\end{longtblr}
    
\subsubsection{Ease of Understanding Score (\(S_{EU}\))}
Answer the Ease of Understanding Questionnaire found in Table \ref{tab:QEU_table}, which are related to the average \textit{feel-safe percentage} (\(FSP\)) of pedestrians after encountering a vehicle making use of the proposal.

The \(FSP\) can be found by using the following formula:
    \begin{equation} \label{fspf:5}
        FSP = \frac{{Total Time}_{Correct Response}}{{Total Time}_{Interaction}} \times 100  
    \end{equation}

Where \({Total Time}_{Interaction}\) is the total amount of time that the interaction between the observer and the vehicle took place, and \({Total Time}_{Interaction}\) is the additive amount of time where the observer was responding to the eHMI(s) of the proposal in the way intended by the proposal.

If this proposal includes multiple non-identical eHMIs or communicates a variety of information, use the lowest \(FSP\) between all of them.

Then calculate the Ease of Understanding Score (\(S_{EU}\)) using \ref{S_EU:5}, where \(EU_{count}\) is the number of questions in Table \ref{tab:QEU_table}, and \(EU_{n}\) is the value obtained after answering the corresponding question. 
    \begin{equation} \label{S_EU:5}
        S_{EU} = \frac{\sum_{n=1}^{EU_{count}} EU_n}{800} \times 10  
    \end{equation}
    
\begin{longtblr}[
  caption = {Ease of Understanding Questionnaire},
  label = {tab:QEU_table},
]{
  width = \linewidth,
  colspec = {Q[42]Q[50]Q[854]},
  row{1} = {c,font=\bfseries},
  row{2} = {font=\bfseries},
  cell{1}{1} = {c=3}{0.94\linewidth},
  cell{3}{3} = {font=\bfseries},
  cell{7}{3} = {font=\bfseries},
  cell{11}{3} = {font=\bfseries},
  hlines,
  vlines,
}
EASE OF UNDERSTANDING   &  & \\
ID & PTS & QUESTION\\
 &  & PEDESTRIANS\\
EU1 & Mpp & CLARITY FOR MINOR PEDESTRIANS (Mpp: The "feel-safe percentage" for pedestrians under the age of 18 interacting with this proposal, after nine (9) encounters with a vehicle making use of it.)\\
EU2 & App & CLARITY FOR ADULT PEDESTRIANS (App:  The "feel-safe percentage" for pedestrians between the age of 18 and 64 interacting with this proposal, after nine (9) encounters with a vehicle making use of it.)\\
EU3 & Epp & CLARITY FOR ELDERLY PEDESTRIANS (Epp:  The "feel-safe percentage" for pedestrians over the age of 64 interacting with this proposal, after nine (9) encounters with a vehicle making use of it.)\\
 &  & CICLYSTS\\
EU4 & Mpc & CLARITY FOR MINOR CYCLISTS (Mpc: The "feel-safe percentage" for cyclists under the age of 18 interacting with this proposal, after nine (9) encounters with a vehicle making use of it.)\\
EU5 & Apc & CLARITY FOR ADULT CYCLISTS (Apc:  The "feel-safe percentage" for cyclists between the age of 18 and 64 interacting with this proposal, after nine (9) encounters with a vehicle making use of it.)\\
EU6 & Epc & CLARITY FOR ELDERLY CYCLISTS (Epc:  The "feel-safe percentage" for cyclists over the age of 64 interacting with this proposal, after nine (9) encounters with a vehicle making use of it.)\\
 &  & DRIVERS\\
EU7 & Ap & CLARITY FOR ADULT DRIVERS (Ap:  The "feel-safe percentage" for pedestrians between the age of 18 and 64 interacting with this proposal, after nine (9) encounters with a vehicle making use of it.)\\
EU8 & Ep & CLARITY FOR ELDERLY DRIVERS (Ep:  The "feel-safe percentage" for pedestrians over the age of 64 interacting with this proposal, after nine (9) encounters with a vehicle making use of it.)
\end{longtblr}

\subsubsection{Constant Communication Score (\(S_{CC}\))}
Answer the Constant Communication Questionnaire found in Table \ref{tab:QCC_table}. Then calculate the Constant Communication Score (\(S_{CC}\)) using \ref{S_CC:6}, where \(CC_{count}\) is the number of questions in Table \ref{tab:QCC_table}, and \(CC_{n}\) is the value obtained after answering the corresponding question. 
    \begin{equation} \label{S_CC:6}
        S_{CC} = \frac{\sum_{n=1}^{CC_{count}} CC_n}{CC_{count}} \times 10  
    \end{equation}

\begin{longtblr}[
  caption = {Constant Communication Questionnaire},
  label = {tab:QCC_table},
]{
  width = \linewidth,
  colspec = {Q[58]Q[55]Q[833]},
  row{1} = {c,font=\bfseries},
  row{2} = {font=\bfseries},
  cell{1}{1} = {c=3}{0.939\linewidth},
  hlines,
  vlines,
}
CONSTANT COMMUNICATION   &  & \\
ID & PTS & QUESTION\\
 &  & {\textbf{ONE-WAY GENERAL COMMUNICATION}\\\textbf{The eHMI(s) in this proposal can explicitly communicate to observers...}}\\
CC1 & 1 & ...when the vehicle is moving forward.\\
CC2 & 1 & ...when the vehicle is backing up.\\
CC3 & 1 & ...when the vehicle is slowing down.\\
CC4 & 1 & ...when the vehicle is speeding up.\\
CC5 & 1 & ...when autonomous mode is engaged.\\
CC6 & 1 & ...when autonomous mode is not engaged.\\
CC7 & 1 & ...when the vehicle is unable to move.\\
CC8 & 1 & ...when a door is opening.\\
CC9 & 1 & ...when a door is opening, and it specifies which door(s) are opening.\\
CC10 & 1 & ...when this vehicle is obscuring another moving vehicle.\\
CC11 & 1 & ...when the vehicle is turning, and what direction it's turning to (does not replace blinkers).\\
CC12 & 1 & ...when the vehicle is changing lanes, and what lane it's changing into (does not replace blinkers).\\
 &  & {\textbf{ONE-WAY PEDESTRIAN COMMUNICATION}\\\textbf{The eHMI(s) in this proposal can explicitly communicate to pedestrians...}}\\
CC13 & 1 & ...when they’ve been detected.\\
CC14 & 1 & ...when the vehicle is yielding to them.\\
CC15 & 1 & ...when the vehicle is not yielding to them.\\
 &  & {\textbf{TWO-WAY PEDESTRIAN COMMUNICATION}\\\textbf{The eHMI(s) in this proposal can explicitly communicate to pedestrians...}}\\
CC16 & 1 & ...where the vehicle estimates they are.\\
CC17 & 1 & ...how to request the vehicle to be stopped, if applicable. Otherwise, score with 1.\\
CC18 & 1 & ...whether the pedestrian's reaction to the vehicle have been detected and interpreted, if applicable. Otherwise, score with 1.\\
 &  & {\textbf{MALFUNCTIONS}\\\textbf{The eHMI(s) in this proposal can explicitly communicate externally...}}\\
CC19 & 1 & ...when the vehicle estimates it will not be able to prevent an impact.\\
CC20 & 1 & ...when the vehicle’s sensors are malfunctioning.\\
CC21 & 1 & ...when the vehicle’s driving systems are malfunctioning.\\
CC22 & 1 & ...when weather conditions are interfering with the vehicle's driving systems\\
CC23 & 1 & ...when weather conditions are interfering with the sensors.\\
 &  & {\textbf{OTHER}\\\textbf{The eHMI(s) in this proposal can explicitly communicate externally...}}\\
CC24 & 1 & ...when the vehicle is turned on and when it is turned off.
\end{longtblr}

\subsubsection{Positioning Score (\(S_{P}\))}

\begin{enumerate}
    \item Identify all individual eHMI elements of the proposal that serve different purposes to any degree. For example, if the proposal outlines a screen on the windshield and another on the grill, and both only tell when the vehicle is in autonomous mode, then they can be counted as a single element for the purpose of this section of the questionnaire; but if the screen on the grill also tells pedestrians of the vehicle deceleration, then it should be counted separately.
    \item Assign each identified element a unique natural number, starting with 1. This will be the identifier of that element.
    \item For each element, answer each question of the Positioning Questionnaire in Table \ref{tab:QP_table}.
    \item For each element, calculate \(P_{P_x}\) using \ref{P_P_x:1} where \(S_{x_n}\) is the value obtained by answering question \(P_n\) in respect for the element with identifier \(x\), and \(y_{count}\) is the number of questions, between \(P_{34}\) and \(P_{41}\) that were answered with a ``Yes".
    
    \begin{equation} \label{P_P_x:1}
        P_{P_x} = \frac{\sum_{n=33}^{40} P_{x_n}}{y_{count}}  
    \end{equation}

    \item Obtain the Positioning Score (\(P_S\)) making use of \ref{P_S:2}, where  \(x_{count}\) is the number of elements identified.
        \begin{equation} \label{P_S:2}
        S_{P} = \frac{\sum_{n=1}^{x_{count}} P_{P_{n}}}{x_{count}} \times 10  
        \end{equation}
\end{enumerate}

\begin{longtblr}[
  caption = {Positioning Questionnaire},
  label = {tab:QP_table},
]{
  width = \linewidth,
  colspec = {Q[38]Q[200]Q[427]},
  row{1} = {c,font=\bfseries},
  row{2} = {font=\bfseries},
  cell{1}{1} = {c=3}{0.94\linewidth},
  cell{3}{3} = {font=\bfseries},
  hlines,
  vlines,
}
POSITIONING   &  & \\
ID & PTS & QUESTION\\
 &  & MIRRORING AND SYMMETRY\\
P1 & 1 & This eHMI is symmetrical accross the length of this vehicle (i.e. the left and right side of the vehicle are symmetrical).\\
P2 & 2 & This eHMI is standing on the roof of the vehicle.\\
 &  & {\textbf{VISIBILITY}\\\textbf{In a given vehicle, the eHMI can be found anywhere on...}}\\
P3 & P1 + 1 & ...the back bumpers.\\
P4 & 2 & ...the back central light.\\
P5 & P1 + 1 & ...the back fenders.\\
P6 & P1 + 1 & ...the back low reflector.\\
P7 & 2 & ...the location of the back plate.\\
P8 & P1 + 1 & ...the back window rails.\\
P9 & P1 + 1 & ...the back window.\\
P10 & P1 + 1 & ...the back window lower frame.\\
P11 & 0 & ...the cowl cover.\\
P12 & P1 + 1 & ...the front bumper.\\
P13 & P1 + 1 & ...the front low reflector.\\
P14 & P1 + 1 & ...the front doors.\\
P15 & P1 + 1 & ...the front fenders.\\
P16 & 2 & ...the location of the front plate.\\
P17 & P1 + 1 & ...the windshield side rails.\\
P18 & P1 + 1 & ...the front wheels.\\
P19 & P1 + 1 & ...the front windows.\\
P20 & P1 + 1 & ...the grill.\\
P21 & P1 + 1 & ...the headlights or their immediate surroundings.\\
P22 & P1 + 1 & ...the hood.\\
P23 & 0 & ...the back doors.\\
P24 & P1 + 1 & ...the rear wheels.\\
P25 & P1 + 1 & ...the back windows.\\
P26 & P1 + 1 & ...the rocker panels.\\
P27 & P2 & ...the roof.\\
P28 & P1 + 1 & ...the side mirrors' back.\\
P29 & P1 + 1 & ...the side mirrors' front.\\
P30 & P1 + 1 & ...the windshield.\\
P31 & P1 + 1 & ...the tail lights.\\
P32 & P1 + 1 & ...the trunk.\\
P33 & P1 + 1 & ...the quarter back window.\\
 &  & {\textbf{EHMI(S) PURPOSE}\\\textbf{The purpose of this eHMI in this proposal is to...}}\\
P34 & {[}MAX(P12, P13, P15, P17, P18, P20, P21, P22, P24, P25, P29, P30) + MAX(P5, P8, P12, P14, P15, P17, P18, P19, P20, P21, P22, P24, P25, P29, P33)] / 4 & ...inform pedestrians on the sidewalk of the intent of the AV as an approaching vehicle on the road.\\
P35 & {[}MAX(P15, P21, P22) + MAX(P3, P4, P5, P7, P8, P9, P10, P18, P28, P29, P31, P32)] / 4 & ...inform pedestrians on the sidewalk of the intent of the vehicle as it is moving away from them on the road.\\
P36 & MAX(P12, P13, P15, P16, P17, P20, P21, P22, P29, P30) / 2 & ...inform pedestrians on the sidewalk as the vehicle comes out of an alleyway or exit that is on the opposite side of the road\\
P37 & MAX(P14, P15, P17, P18, P26) / 2 & ...inform pedestrians on the sidewalk of the vehicle's intent as it exits an alleyway or exit that goes over the sidewalk the pedestrian is on.\\
P38 & MAX(P5, P8, P14, P17, P33) / 2 & ...inform drivers driving in the opposite direction of the vehicle of its intent.\\
P39 & MAX(P12, P13, P15, P16, P20, P21, P22) / 2 & ...inform drivers in front of the vehicle that are driving in the same direction.\\
P40 & MAX(P3, P6, P7, P9, P10, P32) / 2 & ...inform drivers behind the vehicle that are driving in the same direction.\\
P41 & MAX(P12, P13, P15, P20, P21, P22) / 2 & ...inform drivers on the road as the vehicle is about to turn into it from an alleyway or exit
\end{longtblr}

\subsubsection{Readability Score (\(S_{R}\))}
Answer the Readability Questionnaire found in Table \ref{tab:QR_table}. Then calculate the Readability Score (\(S_{R}\)) using \ref{S_R:8}, where \(R_{count}\) is the number of questions in Table \ref{tab:QR_table}, and \(R_{n}\) is the value obtained after answering the corresponding question. 
    \begin{equation} \label{S_R:8}
        S_{R} = \frac{\sum_{n=1}^{R_{count}} R_n}{R_{count}} \times 10  
    \end{equation}

\begin{longtblr}[
  caption = {Readability Questionnaire},
  label = {tab:QR_table},
]{
  width = \linewidth,
  colspec = {Q[50]Q[50]Q[860]},
  row{1} = {c,font=\bfseries},
  row{2} = {font=\bfseries},
  cell{1}{1} = {c=3}{0.941\linewidth},
  hlines,
  vlines,
}
READABILITY   &  & \\
ID & PTS & QUESTION\\
 &  & DYNAMIC LIGHT ADAPTATION\\
R1 & 1 & At least 1/3 of the eHMIs described in the proposal can adjust their brightness based on weather and lighting conditions.\\
R2 & 1 & At least 2/3 of the eHMIs described in the proposal can adjust their brightness based on weather and lighting conditions.\\
R3 & 1 & All of the eHMIs described in the proposal can adjust their brightness based on weather and lighting conditions.\\
 &  & {\textbf{READABILITY CONDITIONS - LIGHT}\\\textbf{Accounting for dynamic adjustments if applicable, in clear weather, all relevant eHMI(s) in this proposal 33 feet away from the observer are...}}\\
R4 & 1 & ...fully readable when placed under an average illuminance of between 0.0001 lux (e.g. moonless night with overcasted sky) and 0.002 lux (e.g. moonless night with airglow).\\
R5 & 1 & ...somewhat distinguishable when placed under an average illuminance of between 0.0001 lux (e.g. moonless night with overcasted sky) and 0.002 lux (e.g. moonless night with airglow). Answer 1 if R1 is 1.\\
R6 & 1 & ...fully distinguishable when placed under an average illuminance of between 3.5 lux (e.g. dark limit of civil twilight) and 20 lux (e.g. public area with dark surroundings).\\
R7 & 1 & ...somewhat distinguishable when placed under an average illuminance of between 3.5 lux (e.g. dark limit of civil twilight) and 20 lux (e.g. public area with dark surroundings). Answer 1 if R3 is 1.\\
R8 & 1 & ...fully distinguishable when placed under an average illuminance of between 100 lux (e.g. very dark overcast day) and 1000 lux (e.g overcast day).\\
R9 & 1 & ...somewhat distinguishable when placed under an average illuminance of between 100 lux (e.g. very dark overcast day) and 1000 lux (e.g overcast day). Answer 1 if R5 is 1.\\
R10 & 1 & ...fully distinguishable when placed under an average illuminance of between 10000 lux and 25000 lux (e.g full daylight, not direct sun).\\
R11 & 1 & ...somewhat distinguishable when placed under an average illuminance of between 10000 lux and 25000 lux (e.g full daylight, not direct sun). Answer 1 if R7 is 1.\\
R12 & 1 & ...fully distinguishable when placed under an average illuminance of between 50000 and 130000 lux (e.g full daylight, sun overhead).\\
R13 & 1 & ...somewhat distinguishable when placed under an average illuminance of between 50000 and 130000 lux (e.g full daylight, sun overhead). Answer 1 if R9 is 1.\\
 &  & {\textbf{READABILITY CONDITIONS - WEATHER}\\\textbf{Accounting for dynamic adjustments if applicable, and under the best lighting conditions, all relevant eHMI(s) in this proposal 33 feet away from the observer are fully readable...}}\\
R14 & 1 & ...under a light rain (the average diameter of liquid water drops is no more than 0.10 inches ).\\
R15 & 1 & ...under moderate rain (the average diameter of liquid water drops is between 0.11 and 0.30 inches).\\
R16 & 1 & ...under a heavy rain (the average diameter of liquid water drops is more than 0.30 inches).\\
R17 & 1 & ...when lightning is present.\\
R18 & 1 & ...in dense fog (horizontal visibility reduced to under 1320 ft).\\
R19 & 1 & ...in medium fog (horizontal visibility is reduced to between 1321 ft and 5280 ft).\\
R20 & 1 & ...in light fog (horizontal visibility is not reduced to under 5280 ft ).\\
 &  & {\textbf{READABILITY CONDITIONS - DISTANCE}\\\textbf{Under the best weather and light conditions for the relevant eHMI(s) in this proposal, they can be fully read and understood by a person...}}\\
R21 & 1 & 33 feet away.\\
R22 & 1 & 87 feet away.\\
R23 & 1 & 142 feet away.\\
R24 & 1 & 197 feet away.\\
R25 & 1 & 252 feet away.\\
R26 & 1 & 306 feet away.\\
R27 & 1 & 361 feet away.\\
R28 & 1 & 416 feet away.\\
R29 & 1 & 470 feet away.\\
R30 & 1 & 525 feet away.\\
 &  & {\textbf{READABILITY CONDITIONS - ANGLE}\\\textbf{If at any given time at least one of the eHMIs in this proposal is fully readable and understandable from all horizontal angles, answer R31 through R40 with a 1. Otherwise, under the best possible conditions (weather, light, distance, etc), can the average pedestrian or driver fully read and understand any and all given information displayed by any of the relevant eHMIs in this proposal, when the horizontal angle formed by the direction of their sight and the element directly displaying the information is...}}\\
R31 & 1 & 90 degrees (direction of view perpendicular to the eHMI, observer in full-front view of it)?\\
R32 & 1 & Between 90 and 100 (90°, 100°]?\\
R33 & 1 & Between 100 and 110 (100°, 110°]?\\
R34 & 1 & Between 110 and 120 (110°, 120°]?\\
R35 & 1 & Between 120 and 130 (120°, 130°]?\\
R36 & 1 & Between 130 and 140 (130°, 140°]?\\
R37 & 1 & Between 140 and 150 (140°, 150°]?\\
R38 & 1 & Between 150 and 160 (150°, 160°]?\\
R39 & 1 & Between 160 and 170 (160°, 170°]?\\
R40 & 1 & More than 170 degrees from it?
\end{longtblr}

\subsubsection{Weighted eHMI Score}
After the score for all seven categories has been calculated, one must set a weight for each category following the equality in \ref{W:13}. 
    \begin{equation} \label{W:12}
        \chi \rightarrow \{S, CE, A, EU, CC, P, R\}
    \end{equation}
    \begin{equation} \label{W:13}
        \sum_{i \varepsilon \chi}^{} W_i = 7
    \end{equation}

Finally, the Final Weighted Score can be calculated using \ref{S:14}
    \begin{equation} \label{S:14}
        S({W_{S}},{W_{CE}},{W_{A}},{W_{EU}},{W_{CC}},{W_{P}},{W_{R}}) = \sum_{i \varepsilon \chi}^{} (W_i \times S_i)
    \end{equation}







\end{appendices}


\bibliography{sn-bibliography}

\end{document}